\documentclass[jou]{apa7}

\usepackage[american]{babel}
\usepackage[style=apa,sortcites=true,sorting=nyt,backend=biber]{biblatex}
\usepackage{amsfonts}

\usepackage[utf8]{inputenc} % allow utf-8 input
\usepackage{csquotes}
\usepackage[T1]{fontenc}    % use 8-bit T1 fonts
\usepackage{hyperref}       % hyperlinks
\usepackage{url}            % simple URL typesetting
\usepackage{booktabs}       % professional-quality tables
\usepackage{amsfonts}       % blackboard math symbols
\usepackage{nicefrac}       % compact symbols for 1/2, etc.
\usepackage{microtype}      % microtypography
\usepackage{graphicx}
\usepackage{doi}
\usepackage{booktabs}       % professional-quality tables
\usepackage{adjustbox}
\usepackage{amsmath}
\usepackage{amssymb}
\usepackage{float}
\usepackage{graphicx}
\usepackage{hhline}
\usepackage{mathtools}
\usepackage{multicol}
\usepackage{multirow}
\usepackage{enumitem}
\usepackage[normalem]{ulem}
\usepackage{wasysym}
\usepackage[toc,page]{appendix}
\graphicspath{{media/}}     % organize your images and other figures under media/ folder

\usepackage{pgf}
\usepackage[absolute, overlay]{textpos}
\textblockorigin{\paperwidth}{25.0 pt}
\usepackage{tcolorbox}

\usepackage[labelfont=bf,textfont=normalfont]{caption}

\DeclareLanguageMapping{american}{american-apa}
\title{A Theory of Relation Learning and Cross-domain Generalization}
\shorttitle{Cross-domain Generalization}

\authorsnames[1,1,{2,3},4]{
    \addORCIDlink{Leonidas A. A. Doumas}{0000-0002-4048-6282},
    \addORCIDlink{Guillermo Puebla}{0000-0001-7002-7776}, \addORCIDlink{Andrea E. Martin}{0000-0002-3395-7234}, \addORCIDlink{John E. Hummel}{0000-0002-1585-9155}
    }\authorsaffiliations{{\footnotesize School of Philosophy, Psychology, and Language Sciences, The University of Edinburgh},{\footnotesize Language and Computation in Neural Systems Group, Max Planck Institute for Psycholinguistics},{\footnotesize Donders Centre for Cognitive Neuroimaging, Radboud University},{\footnotesize Department of Psychology, University of Illinois}}

\leftheader{Doumas, Puebla, Martin, \& Hummel}

\abstract{
People readily generalize knowledge to novel domains and stimuli. We present a theory, instantiated in a computational model, based on the idea that cross-domain generalization in humans is a case of analogical inference over structured (i.e., symbolic) relational representations. The model is an extension of the LISA \parencite{hummel1997distributed} and DORA \parencite{doumas2008theory} models of relational inference and learning. The resulting model learns both the content and format (i.e., structure) of relational representations from non-relational inputs without supervision, when augmented with the capacity for reinforcement learning, leverages these representations to learn individual domains, and then generalizes to new domains on the first exposure (i.e., zero-shot learning) via analogical inference. We demonstrate the capacity of the model to learn structured relational representations from a variety of simple visual stimuli, and to perform cross-domain generalization between video games (Breakout and Pong) and between several psychological tasks. We demonstrate that the model's trajectory closely mirrors the trajectory of children as they learn about relations, accounting for phenomena from the literature on the development of children's reasoning and analogy making. The model's ability to generalize between domains demonstrates the flexibility afforded by representing domains in terms of their underlying relational structure, rather than simply in terms of the statistical relations between their inputs and outputs.
}

\keywords{relation learning, generalization, learning relational content, learning structured representations, neural oscillations}

\authornote{
  This research was supported by the University of Edinburgh PPLS pilot grant scheme to LAAD, grant ES/K009095/1 from Economic and Social Research Council of the United Kingdom and grant 016.Vidi.188.029 from the Netherlands Organization for Scientific Research to AEM, and AFOSR Grant AF-FA9550-12-1-003 to JEH. AEM was further supported by the Max Planck Research Group and Lise Meitner Research Group ‘Language and Computation in Neural Systems' from the Max Planck Society. We thank Mante S. Nieuwland and Hugh Rabagliati and three anonymous reviewers helpful for comments on earlier versions of this manuscript, and Dylan Opdam and Zina Al-Jibouri for assistance with the initial research. Ethics and consent for human participants from Ethics Committee of the faculty of social science of Radboud University. Parts of this work were presented at the 2019 Cognitive Computational Neuroscience meeting and the 2020 Cognitive Science Society meeting. The authors declare no conflicts of interest. The data generated and presented in the current study, and source code are available from \url{https://github.com/AlexDoumas/BrPong_1}.
  $^*$Correspondence should be addressed to Leonidas A. A. Doumas, e-mail: alex.doumas@ed.ac.uk.
  }

\begin{document}

% Annotation at top of first page
\pgfmathwidth{100 pt} % the width of text is stored in '\pgfmathresult'
\begin{textblock}{\pgfmathresult}[1.015, 0](0, 0)
%% value of '\pgfmathresult' is used to set the width of text block
%% '[1, 0]' sets the anchor point of text block to be its top right corner
%% '(0, 0)' sets the anchor point right at the origin which is set by '\textblockorigin' in the preamble
\noindent
\begin{tcolorbox}[width=0.94\linewidth, boxsep=0.5mm, colback=white, colframe=red!75]
Author copy of: Doumas, L. A. A., Puebla, G., Martin, A. E., \& Hummel, J. E. (in press). A theory of relation learning and cross-domain generalization. \textit{Psychological Review}. This article may not exactly replicate the authoritative document published in \textit{Psychological Review}. It is not the copy of record.
\end{tcolorbox}
\end{textblock}

\maketitle

\section{Introduction}

Many children learn the apocryphal story of Newton discovering his laws of mechanics when an apple fell from a tree and hit him on the head. We were told that this incident gave him the insight that ultimately led to a theory of physics that, hundreds of years later, would make it possible for person to set foot on the moon. The point of this story is not about the role of fruit in humankind's quest to understand the universe; the point is that a single incident—a child's proxy for a small number of observations—led Newton to discover laws that generalize to an unbounded number of new cases involving not only apples, but also cannonballs and celestial bodies. Without being told, we all understand that \emph{generalization} —the realization that a single set of principles governs falling apples, cannonballs, and planets alike—is the real point of the story. 

Although the story of Newton and the apple is held up as a particularly striking example of insight leading to broad generalization, it resonates with us because it illustrates a fundamental property of human thinking: People are remarkably good—irresponsibly good, from a purely statistical perspective—at applying principles discovered in one domain to new domains that share little or no superficial similarity with the original. A person who learns how to count using stones can readily apply that knowledge to apples. A graduate student who learns to analyze data in the context psychological experiments can readily generalize that knowledge to data on consumer preferences or online search behavior. And a person who learns to play a simple videogame like Breakout (where the player moves a paddle at the bottom of the screen horizontally to hit a ball towards bricks at the top of the screen) can readily apply that knowledge to similar games, such as Pong (where the player moves a paddle on the side of the screen vertically and tries to hit a ball past the opponent paddle on the other side of the screen). 

This kind of "cross-domain" generalization is so commonplace that it is tempting to take it for granted, to assume it is a trivial consequence of the same kind of statistical associative learning taught in introductory psychology. But the truth is more complicated. First, there is no clear evidence that any species other than humans is capable of the kind of flexible cross-domain generalization we find so natural \parencite[see][]{penn2008darwin}. And second, while "cross-domain" generalization has also been the subject of substantial research in the field of machine learning \parencite[see, e.g.,][]{gamrian2019transfer,kansky2017schema,zhang2018study}, robust, human-like cross-domain generalization continues to frustrate even the most powerful deep neural networks \parencite[DNNs; see, e.g.,][]{bowers2017parallel, geirhos2020shortcut}. 

In the following we present a theory of human cross-domain generalization. Our primary claim is that human cross-domain generalization is a product of analogical inference performed over structured relational representations of multiple domains. We instantiate our theory in a unification and extension of two existing computational models of relational reasoning, LISA \parencite{hummel1997distributed, hummel2003symbolic} and DORA  \parencite{doumas2008theory}. The resulting model accounts for how we acquire structured relational representations from simple visual inputs, integrates with current methods for reinforcement learning to learn how to apply these representations, and accounts for how we leverage these representations to generalize knowledge across different domains. 

In what follows, we first review evidence for the role of structured relational representations in generalization, broadly defined, but especially in cross-domain generalization. Along the way, we discuss what it means for a representation to be structured and relational, contrasting the strengths and limitations of implicit and explicit representations of relations. Next, we outline our theory of human cross-domain generalization and describe the computational model that instantiates it. We then present a series of simulations demonstrating that the model learns structured representations of relations and uses these representations to perform zero-shot (i.e., first trial) cross-domain generalization between different video games and between completely different tasks (video games and analogy-making). We also show that the model's learning trajectory closely mirrors the developmental trajectory of human children. Finally, we discuss the implications for our account, contrasting it with purely statistical learning accounts such as DNNs, and we consider future extensions of the theory. 

In addition to providing an account of cross-domain transfer, the model also represents theoretical advance in at least two other domains, (1) the discovery of the kind of invariant semantic properties that constitute the meaning of a relation, and (2) the integration of explicitly relational representations with reinforcement learning to produce representations with the expressive power of explicitly relational representations and the productive capacity of implicit representations of relations (e.g., as weight matrices). 

\subsection{Relational Representations and Generalization}

Psychologists have long observed that people's concepts, even of "ordinary" domains like cooking and visual occlusion, are like theories or models of the domains in question \parencite[e.g.,][]{carey2000origin, susan2009origin, murphy1985role}, in that they specify the relations among the critical variables in the domain. For example, our understanding of visual occlusion specifies that a larger surface can occlude a smaller surface more than vice-versa \parencite[one can hide behind a hundred-year-old oak better than a sapling; e.g.,][]{hespos2001infants}; that an object continues to exist, even when it's hidden behind an occluder \parencite[e.g.,][]{piaget1954construction}; and that the ability of an occluder to hide an object depends on the relative distances between the occluder, the observer, and the hidden object. Our model of biology tells us that the offspring are the same species as the parents \parencite[e.g.,][]{gelman2003essential}. Our understanding of cooking specifies that the amount of salt one adds to a dish should be proportional to the size of the dish. And our model of a game like tennis, baseball, or Pong tells us that the ability of the racket, bat, or paddle to hit the ball depends on the locations and trajectories of these objects relative to one another. 

The critical property of all these models is that they specify—and thus depend on the capacity to learn and represent—an open-ended vocabulary of relations among variables: Whether \textit{x} can occlude \textit{y} from viewer \textit{v} depends on the relative sizes of \textit{x} and \textit{y}, and the relative distances and angles between \textit{x}, \textit{y}, and \textit{v}. Accordingly, learning a model of a domain entails learning a representation of the relations characterizing that domain. We take this claim—that an "understanding" of a domain consists of a representation of the relations characterizing that domain—to be uncontroversial. However, our claim is stronger than that. We claim that an understanding of a domain consists of \emph{structured relational} representations of the domain. As elaborated shortly, by \emph{structured relational representation}, we mean a representational format that explicitly captures both the semantic content (i.e., meaning) and the compositional structure (i.e., bindings of arguments to relational roles) of a relationship. 

One of the most important manifestations of a capacity to reason about relations is \textit{analogical inference}: inferences based on the relations in which objects are engaged, rather than just the literal features of the objects themselves. Analogical inference is evident in almost every circumstance in which a person demonstrates knowledge on which she was never explicitly trained. The physics student learning Newton's law, $f = ma$, does not need to learn multiplication \emph{de novo} in the context of this new equation, she knows it already. So instead of teaching her an enormous lookup table with all possible $m$ and $a$ as input and the corresponding $f$ as output, the physics teacher simply gives her the equation, knowing her knowledge of multiplication will generalize to the domain of physics. The student who knows she needs at least a grade of 70$\%$ to pass a course also knows, without additional training, that 69$\%$, 68$\%$, 67$\%$, etc. are all failing grades, an inference based on the relation between 70 and all the numbers smaller than it. And if you have a meeting at 2:00 pm, and the current time is 1:00 pm, then you know you are not yet late. Importantly, we know these things without explicit training on each domain individually. An understanding of a relation such as \textit{less-than} or \textit{multiplied-by} simultaneously confers understanding to \textit{all} domains in which it applies (e.g., grades, appointments, recipes, automobile manufacture, etc.). In brief, cross-domain generalization based on structured relational representations is not the exception in human thinking, it is the default. 

\subsection{Learning Relational Representations}

These facts have not escaped the notice of cognitive modelers \parencite[e.g.,][]{doumas2008theory, falkenhainer1989structure, halford1998processing, hummel1997distributed, hummel2003symbolic, paccanaro2001learning}, and recent years have seen increased interest in getting neural networks trained by back propagation to learn useful representations of relations in domains such as relation extraction from pictures \parencite{haldekar2017identifying}, visual question answering \parencite[e.g.,][]{cadene2019murel, ma2018visual, santoro2017simple,xu2016ask}, same-different discrimination \parencite{funke2021five,MESSINA202175}, and even visual  \parencite{hill2018learning,hoshen2017iq} and verbal \parencite{mikolov2013distributed} analogy-making. The core assumptions underlying these efforts are that (1) the relevant relational properties can be discovered as a natural consequence of the statistical properties of the input-output mappings, and that (2) the relevant relations will be represented in the learned weight matrices and will permit relational generalization. 

This statistical approach to relation learning has met with some substantial successes. One strength of this approach is that because the learned relations are represented implicitly in the networks' weight matrices, they are \textit{functional} in the sense that they directly impact the model's behavior: Given one term of a relation, for instance, along with a weight matrix representing a relation, a network of this kind can produce the other term of the relation as an output (see e.g., \cite{lu2012bayesian, leech2008analogy}, but cf. \cite{lu2021probabilistic}). By contrast, models based on more explicit representations of relations \parencite[e.g.,][]{anderson2007can, doumas2008theory, falkenhainer1989structure, hummel1997distributed, hummel2003symbolic}, including the model presented here, must explicitly decide how to apply the relations it knows to the task at hand (e.g., by adding an inferred proposition to a database of known facts; see  \cite{anderson2007can}). As elaborated in the Simulations section, one advance presented in this paper is a technique for using reinforcement learning to choose which relations to use in what circumstances in the context of video game play.  

Although statistical learning of implicit relations achieves impressive results when tested on examples that lie within the training distribution, their performance deteriorates sharply on out-of-distribution samples. For example, the relational network of  \citeauthor{santoro2017simple} (\citeyear{santoro2017simple}) was trained to answer same-different questions about pairs of objects in an image. When the model was tested on shape-color combinations withheld from the training data (e.g., a test image with two identical cyan squares where the model had seen squares and the color cyan individually but not on combination), its performance dropped to chance \parencite[][]{kim2018not}. The limited applicability of the relations learned by these models holds across application domains \parencite[for recent reviews see,][]{ricci2021same, peterson2020parallelograms, sengupta2018robust, stabinger2020evaluating}. 

Why are useful relational representations so hard to learn using traditional statistical learning? The short answer is that although these approaches might capture the content of (some of) the relevant relations in their respective domains, they do not represent those relations in a form that supports broad relational generalization. We argue that flexible, cross-domain generalization relies, not on implicit representations (like weight matrices), but instead on explicitly relational representations, that simultaneously (a) represent the semantic content of a relation (e.g., the meaning of the relation \textit{left-of }()) and (b) represent that content in a format that makes it possible to dynamically bind the relation to arguments without altering the representation of either. The following expands on the distinction between form and content with the goal of clarifying our claims about the nature and utility of structured relational representations for the purposes of generalization. We return to this issue in much more detail in the Discussion, where we relate our simulation results to the differences between structured relational representations, as endorsed in the current theory, and implicit representations of relations, as represented in the weight matrices of some neural networks. 

\subsubsection{Relational Content} 

What does it mean to represent a relation such as \textit{left-of }()? To a first approximation, it means having a unit (or pattern of activation over multiple units) that become active if and only if some part of the network's input is to the \textit{left-of }some other part. For example, imagine a neural network that learns to activate a node in its output layer or in some hidden layer if and only if the network's input contains at least two objects, \textit{i} and \textit{j}, whose locations in the horizontal dimension of the display are unequal. Note that in order to represent \textit{left-of }()\textit{,} \textit{per se}, as opposed to instances of \textit{left-of }() at a specific location in the visual field, the unit or pattern must become active whenever \textit{any} object \textit{i} is left of \textit{any other} object \textit{j}, regardless of the specific objects in question, and regardless of their specific locations in the visual field. That is, a representation of a relation, such as \textit{left-of} ($i$, $j$), is useful precisely to the extent that it is \textit{invariant} with the specific conditions (e.g., the particular retinal coordinates, \textit{x\textsubscript{i}} and \textit{x\textsubscript{j}}) giving rise to it and with the objects (arguments, $i$ and $j$) bound to it  \parencite{hummel1992dynamic}: Such a unit or pattern would, by virtue of its 1:1 correspondence with the presence of things that are left-of other things in the input, represent the semantic content of the relation \textit{left-of }(). 

The invariance of a relational representation is partly responsible for the flexible generalization afforded by such representations. A system that can represent \textit{left-of} (paddle, ball) in a form that is invariant with the specific locations of the paddle and ball is well-prepared to learn a rule such as "if the paddle is left of the ball, then move the paddle to the right" and then apply that rule regardless of the location of the paddle and ball in the game display. Such a representation would even permit generalization to a screen wider than the one used during training (i.e., with previously unseen values of \textit{x}). Conversely, representing \textit{left-of} (paddle, ball) with different units depending on where the paddle and ball are in the display will not permit generalization across locations: Having learned what to do when the paddle is at $x=10$ and the ball is at $x=11$ (represented by a unit we'll call \textit{left-of\textsubscript{[10,11]}}), such a network would not know what to do when the paddle is at $x=12$ and the ball at $x=15$ (represented by a different unit, \textit{left-of\textsubscript{[12,15]}}).

\subsubsection{The Form of Structured Relational Representations}

Being able to represent relational invariants such as \textit{left-of} () and \textit{above} () is extremely useful, if not necessary, for broad, cross-domain generalization, but it is not sufficient. Simply activating a unit or pattern representing \textit{left-of} () does not specify what is to the left of what: Is the paddle left of the ball, or the ball left of the paddle? Or is one of them, or some other object, to the left of some third or fourth object? Knowing only that something is left of something else provides no basis for deciding whether, for example, to move the paddle to the left or right. 

Representing a relation such as \textit{left-of} (\textit{i},\textit{ j}) in a way that can effectively guide reasoning or behavior entails representing both the relational content of the relation (e.g., that something is \textit{left-of} something else, as opposed to, say, \textit{larger-than} something else), and specifying that content in a \textit{format} that makes the bindings of arguments to relational roles explicit. Following the literature on analogy and relational reasoning, we will use the term \textit{predicate} to refer to a representation in this format. For our current purposes, a predicate is a representation (a symbol) that can be bound dynamically to its argument(s) in a way that preserves the invariance of both \parencite[see, e.g.][]{halford1998processing, hummel1992dynamic, hummel1997distributed, hummel2003symbolic}. By "dynamic binding" we mean a binding that can be created and destroyed as needed: To represent \textit{left-of} (paddle, ball), the units representing the paddle must be bound to the units representing the first role of \textit{left-of }() while the units representing the ball are bound to the units representing the second role; and to represent \textit{left-of} (ball, paddle), the very same units must be rebound so that ball is bound to the first role and paddle to the second role. In propositional notation, these bindings are represented by the order of the arguments inside the parentheses. Neural networks need a different way to signal these bindings, and the work reported here follows \citeauthor{doumas2008theory} (\citeyear{doumas2008theory}, see also \cite{hummel1992dynamic, hummel1997distributed, hummel2003symbolic}) and others in using systematic synchrony and asynchrony of firing for this purpose.

What matters about dynamic binding is not that it is based on synchrony or asynchrony of firing; one might imagine other ways to signal bindings dynamically in a neural network. What matters is only that the dynamic binding tag, whatever it is, is independent of the units it binds together. That is, the binding tag must be a second degree of freedom, independent of the units’ activations, with which to represent how those units (representing roles and objects) are bound together. Synchrony of firing happens to be convenient for this purpose, as well as neurally plausible \parencite[e.g.,][]{hummel1992dynamic, hummel1997distributed, reichert2013neuronal, rao2010objective, rao2011effects, shastri1993simple}. For example, to represent \textit{left-of} (paddle, ball), neurons representing \textit{left-of} would fire in synchrony with neurons representing the paddle, while neurons representing \textit{right-of} fire in synchrony with neurons representing the ball (and out of synchrony with the paddle and \textit{left-of} neurons). The very same neurons would also represent \textit{left-of} (ball, paddle), but the synchrony relations would be reversed. In this way, the representation captures the form of the representation (distinguishing \textit{left-of} (ball, paddle) from \textit{left-of} (paddle, ball)), without sacrificing the content of \textit{left-of} (), ball, or paddle. This ability to bind ball and paddle dynamically to the roles of \textit{left-of} () without changing the representation of either derives from the fact that \textit{when} a unit fires, is independent of \textit{how strongly} it fires: The representation is explicitly structured and relational because timing (which carries binding) and activation (which carries content) are independent. 

We posit that representing a relation as a structure that is invariant with its arguments and can be dynamically bound to arguments permits immediate (i.e., zero-shot) generalization to completely novel arguments—including arguments never seen during training. For example, a Breakout player who represents \textit{left-of }() in a way that remains unaffected by the arguments of the relation could adapt rapidly if the paddle and ball were suddenly replaced by, say, a triangle and a square, or a net and a bunny. Armed with the capacity to map a given relation such as \textit{left-of} () in one domain onto a different relation such as \textit{above} () in another—that is, armed with a capacity for analogical mapping—a player could also rapidly adapt "keep the paddle aligned with the ball in the horizontal direction", as in Breakout, to "keep the paddle aligned with the ball in the vertical direction", as in Pong. Such a player would exhibit very rapid cross domain generalization from Breakout to Pong. 

In summary, representing relations explicitly, with a pattern or unit that remains invariant over different instantiations of the relation (e.g., \textit{left-of }() in one location vs. another) and different role bindings (e.g., \textit{left-of} (paddle, ball) vs. \textit{left-of} (triangle, square)) and that can be dynamically bound to arguments, affords enormous flexibility in generalization. We argue that it is precisely this kind of relational generalization that gives rise to cross-domain transfer. 

\subsubsection{Learning Relations} 

An account of how people learn explicitly relational representations must explain how we learn both their content and their form  \parencite{doumas2008theory}.To account for the discovery of relational  \textit{content}, it must specify how we come to detect the basic relational invariants that remain constant across instances of the relation. For example, how can we discover an invariant that holds true across all instances of \textit{left-of} (), given that we only ever observe specific instances of \textit{left-of} () at specific locations? To account for the learning of the \textit{form} of a structured relational representation—that is, the capacity to bind relational roles to their arguments dynamically without destroying the invariant representations of either—the system must be able to solve two additional problems. First, it must be able to isolate the relevant invariants from the other properties of the objects engaged in the relation to be learned. Part of what makes relation learning difficult is that although a goal is to discover an invariant representation of the relation, during acquisition relations never occur in isolation, but always in the context of specific objects engaged in the relation (e.g., it is impossible to observe a disembodied example of \textit{left-of} () as an abstract invariant). Second, having discovered and isolated the relevant invariants, the system must learn a structured \textit{predicate} representation of the relation that can be stored in long-term memory, and can be dynamically bound to arbitrary arguments while remaining independent of those arguments.

\subsection{A Theory of Cross-domain Generalization}

We propose that human cross-domain generalization is a special case of analogical inference over explicitly relational representations. Accordingly, we propose that cross-domain generalization is subject to the constraints on relation learning summarized previously, plus the familiar constraints on analogical reasoning \parencite[see][]{holyoak1995mental, hummel1997distributed, hummel2003symbolic}. Specifically, we propose that cross-domain generalization is a consequence of four fundamental operations: (1) detecting (or learning to detect) relational invariants; (2) learning structured (i.e., predicate) representations of those invariants; (3) using these structured relational representations to construct relational models of various domains (including arithmetic and physics, or video games) procedurally via processes like reinforcement learning; and (4) using those representations to understand new domains by analogy to familiar ones. 

We do not propose that all four of these operations take place \textit{de novo} every time a person generalizes from one domain to another. In particular, if a given domain is familiar to a person, then they will already have performed steps (1)$\ldots$(3) with respect to that domain. Moreover, even most novel domains are almost never completely novel. By the time a person learns Newton’s laws of mechanics, for instance, they have already mastered arithmetic, so although the equations themselves are new to the student, the arithmetic operations they represent are not. Although children likely engage actively in all four steps, by adulthood, the majority of cross-domain transfer—whether from arithmetic to physics, or from one game to another—likely relies to an extent on step (3), learning what known relations might be relevant in a given situation, and most heavily on step (4), using existing relational concepts and domain models to make inferences about novel domains that are themselves represented in terms of relations and objects already familiar to the learner. 

The following presents our model of cross-domain transfer, which performs all four of the steps outlined above: invariant discovery, relation isolation and predication, model construction, and relational inference based on those models. The model is an integration and augmentation of the LISA model of analogical reasoning \parencite{hummel1997distributed, hummel2003symbolic, knowlton2012neurocomputational} and the DORA model of relational learning and cognitive development \parencite{doumas2008theory, doumas2018learning}. LISA and DORA account for over 100 major findings in human perception and cognition, spanning at least seven domains: (a) shape perception and object recognition  \parencite{doumas2010computational, hummel2001complementary, hummel1992dynamic}; (b) relational thinking \parencite{choplin2002magnitude, hummel1997distributed, hummel2003symbolic, krawczyk2004structural, krawczyk2005one, kroger2004varieties, kubose2002role, taylor2009finding}, (c) relation learning \parencite{doumas2012computational, doumas2008theory, jung2015making, jung2015revisiting, livins2015recognising, livins2016shaping}, (d) cognitive development \parencite{doumas2008theory, licato2012exploring, lim2013modeling, sandhofer2008order}, (e) language processing  \parencite{doumas2018learning, martin2017mechanism, martin2019predicate, rabagliati2017representing}, (f) cognitive aging  \parencite{viskontas2004relational}, and (g) decline due to dementia, stress, and brain damage \parencite{morrison2004neurocomputational, morrison2011computational}. Accordingly, we view these systems as a promising starting point for an account of human-level cross-domain generalization. Importantly, LISA provides a solution to problem (4) above (the problem of inference), and DORA provides a solution to problem (2) (learning structured representations from non-structured inputs). The current model integrates LISA and DORA into a single framework, and then extends the resulting model to address problem (1) (the discovery of abstract relational invariants) and problem (3) (model construction via reinforcement learning). 

Our core theoretical claims and their instantiation in the proposed model are summarized in Table 1. These claims along with the core claims of LISA/DORA \parencite{doumas2008theory, hummel1997distributed, hummel2003symbolic} compose the primary assumptions of the approach. We claim that structured relational representations underlie cross-domain generalization as a natural consequence of (a) their general applicability across domains, and (b) their ability to underlie analogies between domains. Cross-domain analogical inference occurs because we build models of domains consisting of an open-ended vocabulary of relations among of the elements the domains \parencite[see also,][]{murphy1985role, carey2000origin}. These representations are structured and relational in that they express the invariant content of the relations they specify in a structured (i.e., symbolic) format that dynamically binds arguments to relational roles. We learn both the content and the format of structured relational representations from experience by explicitly comparing examples. We learn which relations are important for characterizing and acting in a domain through a process of reinforcement learning. 

{\renewcommand\labelitemi{}
\begin{table*}[ht]
  \caption{Core theoretical claims and their instantiation in DORA}
  \label{tab:theoretical}
  \begin{tabular}{ p{0.45\linewidth} p{0.50\linewidth} }
  \toprule
  \textbf{Core theoretical claim} & \textbf{Instantiation in DORA} \\
  \midrule
  
  \vspace*{-0.5\baselineskip}\begin{enumerate}[leftmargin=*]
  \item Cross-domain generalization is a natural consequence of structured relational representations, which express the invariant content of relations and compose dynamic role-argument bindings into propositions.
  
  \begin{enumerate}[leftmargin=*]
  \item[]
  \item[]
  \item Structured relational representations support cross-domain generalization because they are generally applicable across domains.
  \item[]
  \item[]
  \item[]
  \item[]
  \item Cross-domain generalization is a case of analogical inference over domain models.
  \end{enumerate}
  
  \end{enumerate}\vspace*{-\baselineskip} & \vspace*{-0.5\baselineskip}\begin{itemize}[leftmargin=*]
      \item DORA represents relations between elements in a format that makes explicit the relational invariants, the bindings of relational roles to arguments, and the integration of multiple bindings into propositions, and the capacity for cross-domain generalization follows from operations over these representations.
      \item[]
      \item DORA’s representations can be applied promiscuously to characterize genuinely new situations. DORA uses the representations that it has learned in the past to represent novel situations by dynamically binding previously learned predicate representations to objects in those situations.
      \item[]
      \item DORA performs analogical mapping, discovering correspondences between situations based on shared relational structure.
  \end{itemize}\vspace*{-\baselineskip}\\
  \midrule
  
  \vspace*{-0.5\baselineskip}\begin{enumerate}[leftmargin=*]
  \addtocounter{enumi}{1}
  \item Structured relational representations are acquired by a comparison process that reveals and isolates relational invariants and composes them into predicates that can bind dynamically to their arguments.
  \end{enumerate} & \vspace*{-0.5\baselineskip}\begin{itemize}[leftmargin=*]
      \item DORA discovers invariant relational properties by exploiting properties inherent to rate-coded representations of magnitude. DORA then learns structured relational representations of these properties through a process of comparison-based intersection discovery and Hebbian learning.
  \end{itemize}\vspace*{-\baselineskip}\\
  \midrule
  
  \vspace*{-0.5\baselineskip}\begin{enumerate}[leftmargin=*]
  \addtocounter{enumi}{2}
  \item Relations relevant for characterizing and acting in a domain are learned procedurally through reinforcement learning.
  \end{enumerate}\vspace*{-\baselineskip} & \vspace*{-0.5\baselineskip}\begin{itemize}[leftmargin=*]
      \item DORA and its representations integrate smoothly with existing methods for reinforcement learning.
  \end{itemize}\vspace*{-\baselineskip}\\
  \bottomrule
  
  \end{tabular}
\end{table*}
}

The remainder of the paper proceeds as follows. First, we summarize our integration of the LISA/DORA frameworks and describe the current extensions for invariant discovery and genralization. Next, we report simulations demonstrating that (a) the model learns structured relational representations from simple visual inputs without assuming a vocabulary of structured representations \textit{a priori}. (b) These representations support the development of more complex domain models via reinforcement learning. (c) The model uses these representations to generalize to a new domain in a single exposure, exhibiting \textit{zero-shot }transfer. (d) Generalization in the model fails without the structured format of the representations it learns. (e) The representations that the model learns from simple domains transfer readily to more complex tasks like adult analogy problems, and the representations that the model learns meet the hallmarks of human relational cognition. (f) The trajectory of the model as it learns closely mirrors the developmental trajectory of human children and that the representations learned in one domain transfer readily to different laboratory tasks, allowing the model to capture several phenomena from the developmental literature. Finally, we discuss some implications and possible future extensions of the model and contrast our approach with purely statistical accounts of human learning.

\section{The model}

As noted previously, our model of cross-domain generalization is based on an integration and augmentation of the LISA and DORA models of relational reasoning (henceforth, simply DORA). We begin by reviewing how DORA represents relational knowledge, how it uses those representations for reasoning, and how it learns structured representations of relations from unstructured vector-based inputs. 

We next present a novel algorithm for discovering invariant relational properties—that is, the semantic content of relational representations—from simple, nonrelational visual inputs. The resulting model provides the first complete account of how structured representations of visual relations can be learned \textit{de novo} from simple nonrelational inputs without feedback and without assuming a vocabulary of relations \textit{a priori}. The resulting model also provides an account of human cross-domain generalization as a natural consequence. 

The following description of the model presents published details of DORA’s operation only in broad strokes, going into detail only when those details are relevant to understanding the novel extensions of the work (e.g., relation discovery). Complete details of the model’s operation can be found in Appendix A (which includes a functional description of the model, pseudocode, and full computational details). The model’s source code is available online.\footnote{ Source code is available from \url{https://github.com/AlexDoumas/BrPong_1}.}.

\subsection{Representing relational knowledge: LISAese}

\begin{figure*}[!htbp]
\centering
\includegraphics[width=13.76cm,height=6.62cm]{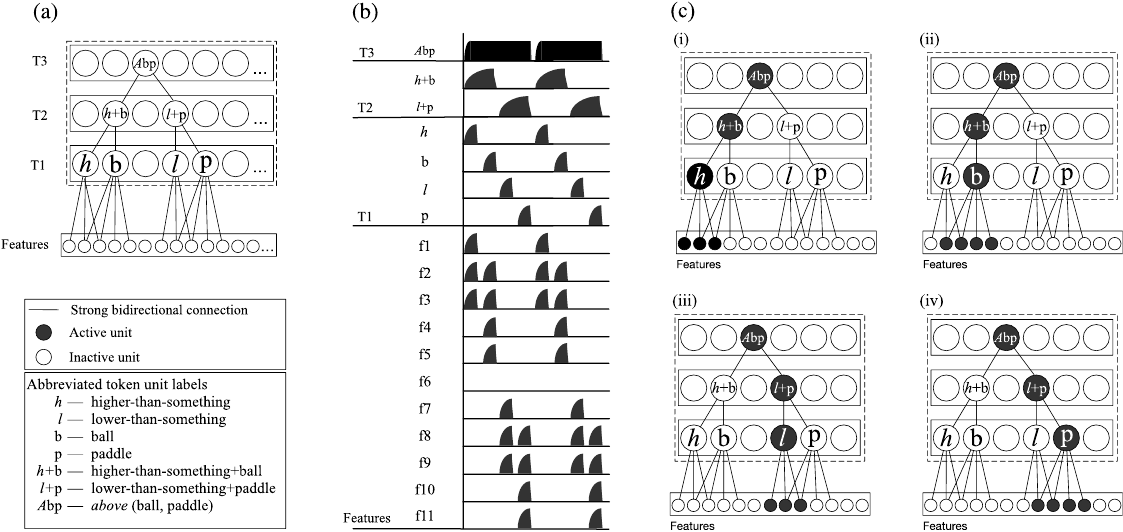}
\caption{Knowledge representation and temporal binding in DORA. (a) Representation of a single proposition (\textit{above} (ball, paddle)) in DORA. Feature units represent properties of objects and relational roles in a distributed manner. Token units in T1 represent objects and roles in a localist fashion; token units in T2 conjunctively bind roles to their arguments (e.g., objects); token units in T3 conjunctively link role-argument pairs into multi-place relations. (b) A time-series illustration of the activation of the units illustrated in (a). Each graph corresponds to one unit in (a) (i.e., the unit with the same name as the graph). The abscissa of the graph represents time, and the ordinate represents the corresponding unit's activation. (c) Time-based binding illustrated as a sequence of discrete frames (i$\ldots$iv). (i) Units encoding \textit{higher-than-something} fire. (ii) Units encoding ball fire. (iii) Units encoding \textit{lower-than-something} fire. (iv) Units encoding paddle fire. Labels in units indicate what the unit encodes (see key); the labels on the units are provided for clarity and are meaningless to DORA.} 
\label{fig:knowledge_representation_and_temporal_binding}
\label{fig:1}
\end{figure*}

We begin by describing the final (i.e., post-learning) state of DORA’s knowledge representations. These representations do not serve as the input to the model but are the result of its learning (as described below). DORA is a neural network consisting of four layers of bidirectionally connected units. DORA represents propositions using a format, \textit{LISAese}, which is a hierarchy of distributed and progressively more localist units whose activations oscillate over progressively slower time scales (moving from the bottom to the top layer of the network; Figure \ref{fig:1}).  At the bottom of the hierarchy, \textit{feature units} represent the basic features of objects and relational roles in a distributed manner. Token units (T1-T3) learn without supervision (see below) to conjunctively code collections of units from the layer below. Tokens at the lowest level of the hierarchy (T1) take inputs directly from feature units and learn to respond to objects or relational roles in a localist fashion. Tokens in the next layer (T2) take their inputs from PO tokens and learn to respond to \textit{pairs} of PO units—that is, to roles and the objects (arguments) to which they are bound. Tokens in the highest layer (T3) learn to respond to collections T2 units, instantiating multi-place relational propositions. 

When they become active, units representing relational roles and their arguments independently (i.e., features and T1 units) must be bound together dynamically. DORA represents dynamic bindings using time: T1 units representing relational roles and their arguments fire out of synchrony but in close temporal proximity. These temporal relations arise from inhibitory interactions among token units. Within a single proposition, token units are laterally inhibitory (e.g., T1 units inhibit other T1 units, T2 units inhibit other T2 units; as elaborated below, lateral inhibition between token units extends beyond propositions but for the present illustration this simplification is appropriate). Each token unit is also yoked to an inhibitory unit, which causes the token’s activation to oscillate, even in response to a fixed excitatory input. 

Together, the yoked inhibitors and lateral inhibitory interactions between tokens cause the constituents of a single proposition to fire in a systematic, hierarchical temporal pattern (Figure \ref{fig:1}b, c): When a T3 unit representing a proposition such as \textit{above} (ball, paddle) becomes active, it will excite the T2 units below itself (here, \textit{above+ball} and \textit{below+paddle}). These T2 units inhibit one another, and one of them (e.g., \textit{above+ball}) will randomly win the initial inhibitory competition, becoming active and driving the other to inactivity. After a fixed number of iterations/milliseconds, \textit{k}, the inhibitor yoked to that T2 unit (\textit{above+ball}) will become active, driving it to inactivity, and allowing the other T2 unit (\textit{below+paddle}) to become active for \textit{k} iterations/milliseconds (until its own inhibitor drives it to inactivity).

This same pattern of lateral inhibition and yoked inhibitor activity causes the T1 units below \textit{above+ball} and \textit{below+paddle}—namely, the T1 units \textit{above} and \textit{ball}, and \textit{below} and \textit{paddle}, respectively—to oscillate out of phase with one another at twice the frequency (that is, half the duration; specifically, \textit{k}/2) at which the T2 units are oscillating. T1 units activate feature units, causing the semantic units’ activity to oscillate at the same frequency as the T1 units. The result on the T1 and semantic units is a repeating temporal pattern of the form [(\textit{above}, \textit{ball}), (\textit{below}, \textit{paddle})], [(\textit{above}, \textit{ball}), (\textit{below}, \textit{paddle})], etc., where units inside parentheses are oscillating out of phase with one another with duration \textit{k}/2, and units inside brackets are out of phase with duration \textit{k} (see Figure \ref{fig:1}b, c). In this way, the network represents relational roles and fillers independently (in the features and T1 units) and simultaneously represents the binding of roles to fillers. 

% use figure* to have figure ignore columns.

\subsection{Interpreting LISAese} 

One advantage of the representations DORA learns is that they are easily interpretable. Units in T1 will learn to represent objects and relational roles, and by inspecting the features to which any given T1 unit is connected, it is possible to determine which object or role it represents. Units in T2 will learn to represent specific predicate-argument bindings, which are interpretable by inspecting the T1 units to which they are connected. And units in T3 will learn to represent complete propositions, which are interpretable by inspecting the connected T2 units. Accordingly, in the following, we will refer to the units DORA learns in terms of these interpretations. We do so solely for clarity of exposition. The labels we use have no meaning to DORA and no impact on its operation.

\subsection{Computational macrostructure}

\begin{figure*}[!htbp]
\centering
\includegraphics[width=13.76cm,height=4.68cm]{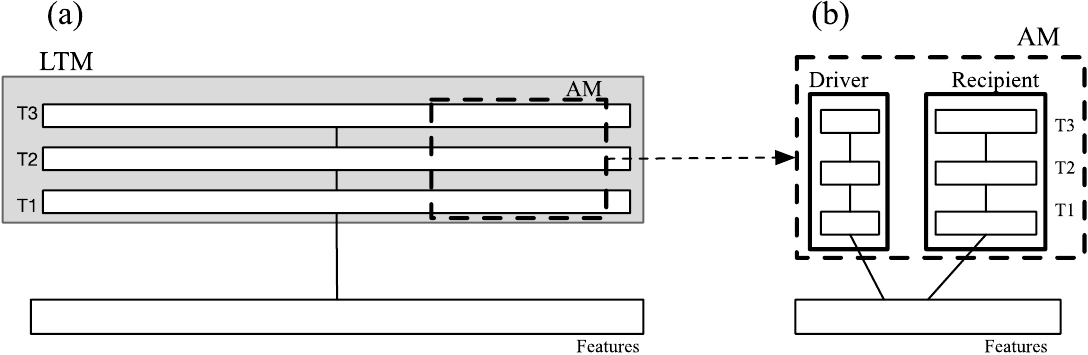}
\caption{DORA's Macrostructure. (a) DORA’s long-term-memory (LTM), consisting of layers of token units (T1-T3; black rectangles), and the feature units connected to the bottom layer of LTM. During processing, some units in LTM enter active memory (AM). (b) Expanded view of AM. AM is composed of two sets, the driver (the current focus of attention) and the recipient (the content of working memory available for immediate processing). Black lines indicate bidirectional excitatory connections.} 
\label{fig:2}
\end{figure*}

Figure \ref{fig:1} depicts the representation of an individual proposition in DORA’s working memory (as synchronized and desynchronized patterns of activation). Figure \ref{fig:2} provides an overview of DORA’s macrostructure. 

A complete analog—situation, story, schema, etc.—consists of the collection of token units that collectively encode its propositional content. Within an analog a single token represents a given object, role, role-binding, or proposition, no matter how many propositions refer to that entity. For example, in a single analog the same T1 unit for \textit{left-of} represents that role in all propositions containing \textit{left-of} as a role. However, separate analogs do not share tokens. For example, one unit would represent \textit{left-of} in DORA’s representation of the game it is currently playing (represented in one analog) and a completely separate token would represent \textit{left-of} in DORA’s representation of a game it had played in the past (represented in a separate analog). Collections of token units (i.e., T1$\ldots$T3) representing the situations and schemas (i.e., "analogs") DORA knows collectively form its LTM (Figure \ref{fig:2}a). 

For the purposes of learning and reasoning—for example when making an analogy between one situation and another—the propositions representing those analogs enter \textit{active memory} (dashed box in Figure \ref{fig:2}a), a state in which they are readily accessible for processing, but not fully active \parencite[see, e.g.,][]{cowan2001magical,hummel1997distributed}. As depicted in Figure \ref{fig:2}b, the analogs in active memory are divided into independent sets: The \textit{driver} corresponds to the current focus of attention (e.g., the state of the current game, as delivered by perceptual processing), and maps onto one or more \textit{recipients} (e.g., an analog describing the model’s emerging understanding of the game).\footnote{ The idea that mutually exclusive sets are fundamental for analogical reasoning goes back to \parencite{gentner1983structure}, and has been implemented in a variety of models (SME, LISA, etc.). As detailed in \textcite{knowlton2012neurocomputational}, we assume these sets are implemented by neurons in posterior frontal cortex with rapidly modifiable synapses that act as "proxies" for larger structures represented elsewhere in cortex. } Token units are laterally inhibitory within but not between sets. The driver/recipient distinction in DORA is different from the more familiar source/target distinction discussed in the analogical reasoning literature. The \textit{target} of analogical reasoning is the novel problem to be solved (or situation to be reasoned about), and the \textit{source} is the analog from which inferences are drawn about the target. As summarized shortly, in DORA, the target analog tends to serve as the driver (i.e., the focus of attention) during memory retrieval and in the initial stages of analogical mapping, whereas the source serves as the driver during analogical inference. 

\subsection{Operation}

In DORA activation generally starts in the driver, passing through the feature units (and any mapping connections, as detailed shortly) into other analogs in LTM (for memory retrieval), including any analogs in the recipient (for mapping, learning, and inference). Token units in the driver compete via lateral inhibition to become active (i.e., token units in the driver laterally inhibit other tokens in the driver in the same layer), generating patterns of activation on the feature units (as described previously; Figure \ref{fig:1}). Units in the recipient (or LTM) compete via lateral inhibition to respond to the resulting patterns on the feature units. This inhibitory competition is hierarchical in time, reflecting the temporal dynamics of the driver and features: T1 units (relational roles and objects) in LTM/recipient compete to respond to patterns generated by individual roles and objects in the driver; T2 units (role bindings) in LTM/recipient compete to respond to specific role/filler bindings; and T3 units compete to respond to complete propositions. The result is a winner-take-all inhibitory competition operating at multiple temporal scales and serves as the foundation of all the functions DORA performs, including memory retrieval, analogical mapping \parencite{hummel1997distributed}, analogical inference \parencite{hummel2003symbolic}, and relation discovery \parencite{doumas2008theory}. 

\paragraph{Memory Retrieval:} Patterns of activation imposed on the feature units by active tokens in the driver will tend to activate token units in LTM that have learned to respond to similar patterns (Appendix A3 for details). For example, the features activated by a paddle in the driver will tend to activate T1 units responsive to paddle features, and the features activated by \textit{leftmost }in the driver will tend to activate T1 units connected to \textit{leftmost} features. Together, these T1 units will tend to excite T2 units for \textit{leftmost+}paddle. Features consistent with ball and \textit{rightmost }would likewise activate T1 units for ball and \textit{rightmost}, which would excite a T2 unit for \textit{rightmost}+ball. Together, the T2 units for \textit{leftmost}+paddle and \textit{rightmost}+ball will tend to activate any T3 unit(s) encoding the proposition \textit{left-of} (paddle, ball): The model will have \textit{recognized} the desynchronized patterns of features as representing the fact that the paddle is \textit{left-of} the ball, which can be retrieved into the recipient. 

\paragraph{Mapping:} One of the most important operations DORA performs is analogical mapping. During mapping, DORA discovers structural correspondences between tokens in the driver and recipient. When tokens in the driver become active, similar tokens are activated in the recipient via the shared feature units. Using a kind of Hebbian learning, the model learns \textit{mapping connections} between coactive units in the same layer across driver and recipient \parencite[][see Appendix A2.2 and Appendix A4 for details]{hummel1997distributed, hummel2003symbolic}. The resulting connections serve both to represent the mappings DORA has already discovered, and to constrain its discovery of additional mappings. The algorithm provides an excellent account of human analogical mapping performance \parencite{doumas2008theory, hummel1997distributed, hummel2003symbolic}. 

\paragraph{Analogical Inference:} Augmented with a simple algorithm for \textit{self-supervised learning} \parencite{hummel2003symbolic}, DORA’s mapping algorithm also provides a psychologically and neurally-realistic account of analogical inference (making relational inferences about one situation based on knowledge of an analogous one; Appendix A2.3.3 for details). The algorithm implements a version of \textcite{holyoak1994component} copy-with-substitution-and-generalization (CWSG) framework. In CWSG, when two situations are analogically mapped, information about one situation can be inferred about the other. For example, if one knows about \textit{situation}-1, where \textit{chase} (Fido, Rosie), and \textit{scared} (Rosie) are true, and maps that onto situation-2, where \textit{chase} (Spot, Bowser) is true, one can copy the representation of the \textit{scared} predicate from situation-1 to situation-2, and then use the mapping of Bowser to Rosie, to copy Bowser as the argument of \textit{scared} to infer \textit{scared} (Bowser). As elaborated below, this process serves as the basis for our proposed solution to the problem of cross-domain generalization.

\subsection{Learning relational format: LISAese from non-structured inputs}

DORA generalizes the operations described above to address the problem of learning structured representations of relations from unstructured "flat" vector/feature-based representations  \parencite{doumas2008theory}. DORA represents relations as collections of linked roles, rather than as monolithic structures: For example, the relation \textit{above} composes the roles \textit{higher} and \textit{lower} rather than consisting strictly of the single atom \textit{above}, as it would in propositional notation or a labeled graph. This role-based approach to representing relations offers several advantages over alternative approaches \parencite[see][for a review]{doumas2005approaches}, one of which is that it makes it possible to learn relations by (a) first learning their roles and then (b) linking those roles together into multi-argument relational structures  \parencite[as described in][]{doumas2008theory}. 

\begin{figure*}[!htbp]
\centering
\includegraphics[width=12.5cm]{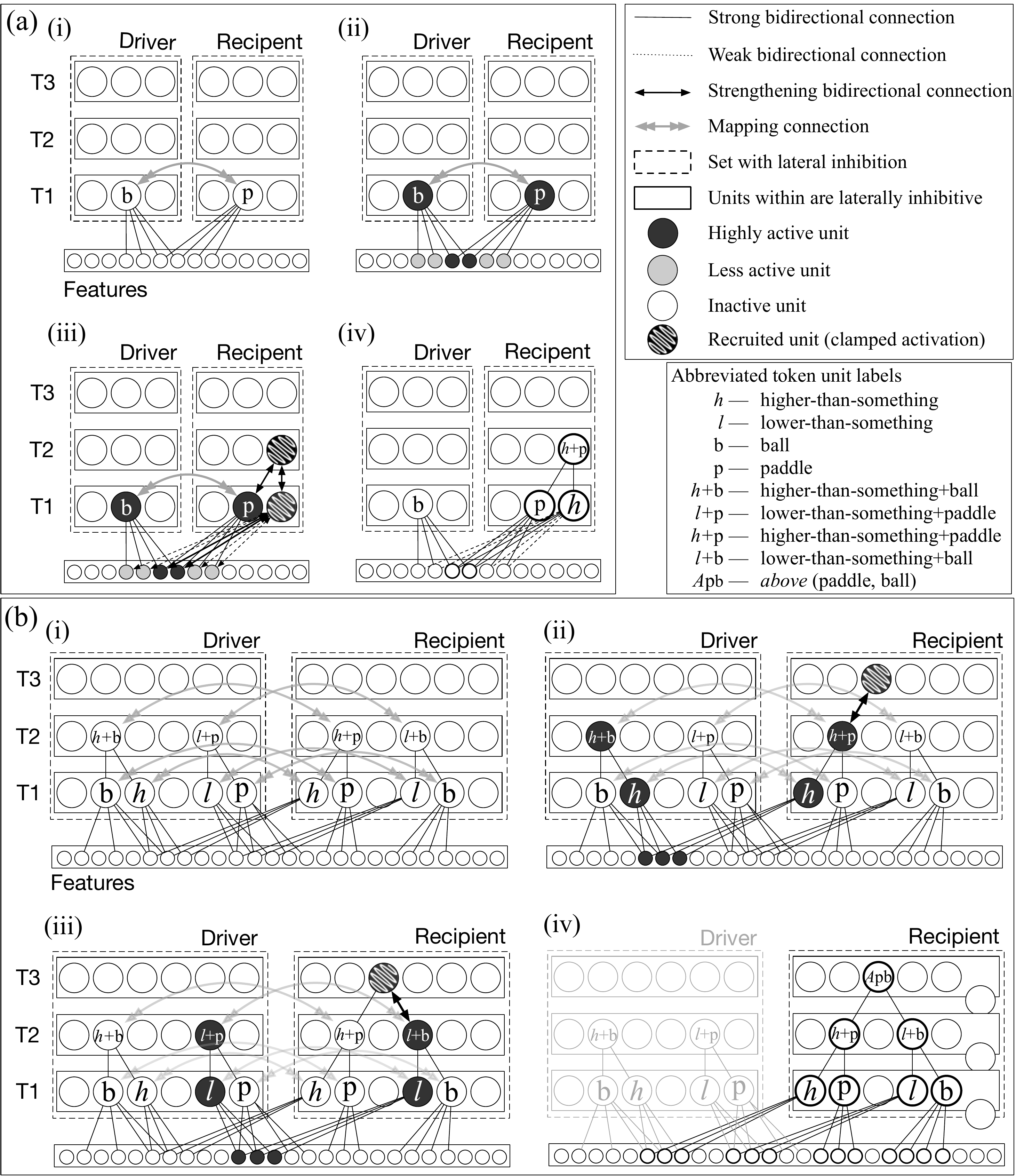}
\caption{Representation learning in DORA. (a) Learning a single-place predicate representation by comparing two objects. (i) A representation of a ball in the driver is mapped (grey double-headed arrow) to a representation of a paddle in the recipient. (ii) The representation of the ball in the driver activates the mapped unit in the recipient (through shared features and mapping connection); as units pass activation to their features, shared features become more active (dark grey units) than unshared features (light grey units). (iii) Units in T1 and T2 are recruited (activation clamped to 1; dark grey units with white squiggle) in the recipient, and weighted connections are learned via Hebbian learning (i.e., stronger connections between more active units). (iv) The result is an explicit representation of the featural overlap of the ball and paddle—in this case the property of being \textit{higher-than-something} (see main text)—that can be bound to an argument (as in Figure \ref{fig:1}). (b) Learning a multi-place relational representation by linking a co-occurring set of role-argument pairs. (i) a representation of a ball that is higher-than-something and a paddle that is lower-than-something is mapped to a different representation of a paddle that is higher-than-something, and a ball that is lower-than-something (e.g., from a different game screen). (ii) (Mapping connections, grey doubled-headed arrows, have been lightened to make the rest of the figure clearer.) When the representation of \textit{higher-than-something} (ball) becomes active in the driver it activates mapped units in the recipient; a T3 unit is recruited (activation clamped to 1; dark grey unit with white squiggle) in the recipient and learns weighted connections to units in T2 via Hebbian learning (iii) When the representation of\textit{ lower-than-something} (paddle) becomes active in the driver, it activates corresponding mapped units in the recipient; the active T3 unit learns weighted connections to T2 units. (iv) The result of learning is a LISAese representation of the relational proposition \textit{above} (ball, paddle) (see Figure \ref{fig:1}). Labels in units indicate what the unit encodes (see key). The labels on the units are provided for clarity and are meaningless to DORA. }
\label{fig:3}
\end{figure*}

DORA’s unsupervised relation learning algorithm \parencite{doumas2008theory} begins with representations of objects encoded as vectors of features. DORA learns single-place predicates—that is, individual relational roles—as follows (see Appendix A2.3.1 for details): (1) By the process of analogical mapping (summarized above) the model maps objects in one situation (the \textit{driver}; e.g., a previous state of the game of Breakout) onto objects in a similar known situation (the \textit{recipient}; e.g., the current state of a game). For example, DORA might map a T1 unit representing the ball in its previous location onto a T1 unit representing the paddle in its current location (Figure \ref{fig:3}ai). Early in learning, these tokens will be holistic feature-based representations specifying the objects' attributes (e.g., location, color, etc.) in a single vector. (2) As a result of this mapping connection, the T1 unit representing the ball in the driver will become coactive with the T1 unit representing the paddle in the recipient. T1 units in both the driver and recipient pass activation to the feature units to which they are connected, so any features connected to the T1 units in both the driver and recipient will receive about twice as much input—and therefore become about twice as active—as any features unique to one or the other (Figure \ref{fig:3}aii). As a result, the \textit{intersection} of the two instances becomes highlighted as the collection of most active features. (3) DORA recruits (activates) a T1 unit and a T2 unit in the recipient, and updates connections between units in adjacent layers via Hebbian learning (Figure \ref{fig:3}aiii). Consequently, the recruited T1 unit will learn connections to active features in proportion to their activation, thereby encoding the shared features (the intersection) of the mapped objects. If the compared objects are both, say, \textit{higher} than something—and so have the features of \textit{higher} in their vectors—then DORA will learn an explicit representation of the features corresponding to being \textit{higher} (the next section describes how such relational features can be learned from absolute location information delivered by the perceptual system). (4) The resulting representation (Figure \ref{fig:3}aiv) can now function as single-place predicate (i.e., relational role), which can be bound to new arguments (i.e., other units in T1) by asynchrony of firing (see Figure \ref{fig:1}). Applied iteratively, this kind of learning produces progressively more refined single-place predicates \parencite{doumas2008theory}. 

The same Hebbian learning algorithm links token units into complete propositions by allowing tokens units in successive layers to integrate their inputs over progressively longer temporal intervals  \parencite[][see Appendix A2.3.1 for details]{doumas2008theory}. The algorithm exploits the fact that objects playing complementary roles of a single relation will tend to co-occur in the environment. For example, the representation of a ball that is higher than something (i.e., \textit{higher-than-something } (ball)) will systematically co-occur with another object (e.g., the paddle), which is lower than something (e.g., \textit{lower-than-something} (paddle); Figure \ref{fig:3}bi). When two sets of co-occurring role-argument pairs are mapped (e.g., an instance where ball is higher-than-something and a paddle is lower-than-something is mapped to an instance where a paddle is higher-than-something and a ball is lower-than-something; Figure \ref{fig:3}bi), a diagnostic pattern of firing emerges: (1) the T2 units coding each predicate-argument binding will oscillate systematically across both driver and recipient (Figure \ref{fig:3}bii-iii). (2) In response, DORA recruits a unit in T3 that learns (via Hebbian learning) connections to the T2 units as they become active (Figure \ref{fig:3}bii-iii). (3) The resulting representation encodes a multi-place relational structure (equivalent to \textit{above} (paddle, ball); Figure \ref{fig:3}biv). \footnote{ In the above example we describe learning a 2-place relation composed of 2 role-filler pairs for the purposes of brevity. DORA learns relations of arity n by linking n role-filler pairs \parencite[e.g., a 3-place relation is composed of three role-filler pairs; see][]{doumas2012computational}. } Applied iteratively over many examples, this algorithm learns abstracted structured representations describing a domain in terms of the properties and relations that characterize that domain \parencite{doumas2008theory}. 

\textcite{doumas2008theory} demonstrated that this algorithm for relation discovery provides a powerful account of numerous phenomena in cognitive development. However, although the Doumas et al. algorithm can learn relational representations with a structured \textit{form} from repeated exposures to nonstructured inputs, it provides little basis for discovering the invariant \textit{content} of those relations—that is, the relational features themselves. Instead, their simulations were based primarily on representations of objects with hand-coded invariant features. The next section describes a novel algorithm for discovering invariant relational features from non-relational inputs.

\subsection{Learning relational content: Discovering Relational Invariance}

To learn an abstract representation of a relation that remains invariant with the relation’s arguments, there must be at least one invariant that characterizes that relation.  For example, to learn a representation of \textit{right-of} that captures every instance of \textit{right-of-ness}, there must be a detectable property(ies) that remains constant over all instances of \textit{right-of-ness} \parencite[see, e.g.,][]{biederman2013human, harnad1990symbol, hummel1992dynamic, kellman2020modeling}. Most previous work on relational perception and thinking has tacitly assumed the existence of such invariants \parencite[e.g.,][]{anderson2007can, doumas2008theory, falkenhainer1989structure, hummel2001complementary, hummel1992dynamic, hummel1997distributed, hummel2003symbolic}. But unless all these invariants are assumed to be innate, there must be some basis for discovering them from representations of values on the underlying dimensions over which the relations are defined (e.g., somehow discovering the notion of \textit{right-of} by observing examples of objects arrayed in the horizontal dimension of space). 

Part of what makes invariant discovery difficult is that it poses a kind of chicken and egg problem: An invariant only seems to be discoverable in a non-invariant input if one knows to look for that invariant in the first place. Consider an invariant like "square". Of all the possible arrangements of pixels on a computer screen, some of them form squares and others do not. Whether a set of pixels forms a square does not depend on the color of the pixels, the color of the background, the locations of the pixels on the screen, or their distances from one another: Provided they are arranged relative to one another in a way that forms a square, then they satisfy the invariant "square". "Square" is a higher-order relational property that is independent of—that is, invariant with—the properties of any of the pixels composing it. Making matters more complicated, "square" is only one of an infinity of such higher-order relational invariants one could find in visual images. Others include rhombi, various triangles, and countless random-looking clouds of points. All these configurations are defined by the spatial relations among sets of points, so any one of them could, in principle, become a perceptual invariant like "square". But not all of them do. Why do we recognize "square" as an invariant, but not any of the nearly infinite random looking (but nonetheless invariant) clouds of points? 

It is not our intent to fully answer to this question here, but one constraint that suggests itself is to \textit{start simple}: Perhaps "square" is not itself a basic ("primitive") invariant in the human cognitive architecture but is instead (at least initially) a composition of several simpler invariants (for example straight lines, equal lengths, right angles, and such) arranged in particular relations to one another (as proposed by Biederman, 1987, and many others). According to this account, the cognitive architecture might be biased to find a small number of very basic invariants \parencite[things such as \textit{equal-to}, \textit{greater-than}, and \textit{less-than}, among others;][]{doumas2008theory, kellman2020modeling}, and compose more complex relational invariants, such as \textit{above }(\textit{x}, \textit{y}), \textit{right-of }(\textit{x}, \textit{y}), and \textit{square} (\textit{x}), by applying the basic relations to specific perceptual and cognitive dimensions, and to other relations. 

Following this intuition, we developed a simple \textit{relational invariant discovery circuit} (henceforth, simply \textit{relational invariance circuit}) to discover the invariants \textit{greater-than}, \textit{equal-to}, and \textit{less-than} on any metric dimension, \textit{m}. This circuit exploits computational properties that naturally emerge whenever magnitudes are \textit{rate coded}, either in terms of the number of units that fire in response to a given magnitude, or in terms of the rates at which individual neurons fire. The basic idea is that for any magnitude represented as a rate code, computing relations such as \textit{greater-than}, \textit{less-than} and \textit{equal-to} is a straightforward matter of responding to the difference between two rates. The sign of this difference (+, 0, or -) becomes an invariant signature of the categorical relations \textit{greater-than}, \textit{equal-to}, and \textit{less-than}, respectively; and by summing over \textit{greater-than} and \textit{less-than}, this same operation yields the invariants \textit{same} and \textit{different} with respect to the dimension in question. 

Let \textbf{m} be an \textit{n}-dimensional vector space, for example a collection of neurons that codes a simple magnitude, \textit{m}, such as size.  The vector \textbf{a} (in \textbf{m}) then represents an object with size $a$, and \textbf{b} represents size $b$. Armed with these rate codes, the difference between sizes $a$ and $b$, $E_{a,b}$ is the directional difference \parencite[e.g.,][]{gallistel2000non, zorzi2005computational}:

\begin{equation}
E_{a,b} = \sum_{i}^{}\left(a_{i}-b_{i}\right)\\
\label{eq:1}
\end{equation} 
%equation 1

\noindent when \textbf{a} is larger than \textbf{b}, $E_{a,b}$ will be positive; when \textbf{a} is smaller than \textbf{b}, $E_{a,b}$ will be negative; and when they are equal, $E_{a,b}$ will be zero. 

Using unsupervised learning, it is straightforward to exploit this regularity to train units to respond explicitly to whether any metric values $a$ and $b$ are equal, unequal with $a>b$, or unequal with $a<b$. The resulting neurons will be invariant representations of the relations \textit{equal}, \textit{greater-than}, and \textit{less-than} for any rate-coded metric dimension \textit{m}. In the language of LISAese, the resulting units could serve as feature units in dimension-specific relations such as \textit{right-of}, \textit{above}, \textit{larger}, etc.\footnote{ We have left the arguments $a$ and $b$ out of the relational expressions here because these units represent invariant relational /emph{content}, but by themselves do not specify how the roles of those relations bind to arguments; that is, they do not specify the relational /emph{format}. However, as noted above, the problem of learning representations with relational format from representations without this format is already solved in DORA. }  

\subsubsection{Circuit Architecture} 

\begin{figure*}[!htbp]
\centering
\includegraphics[width=16cm]{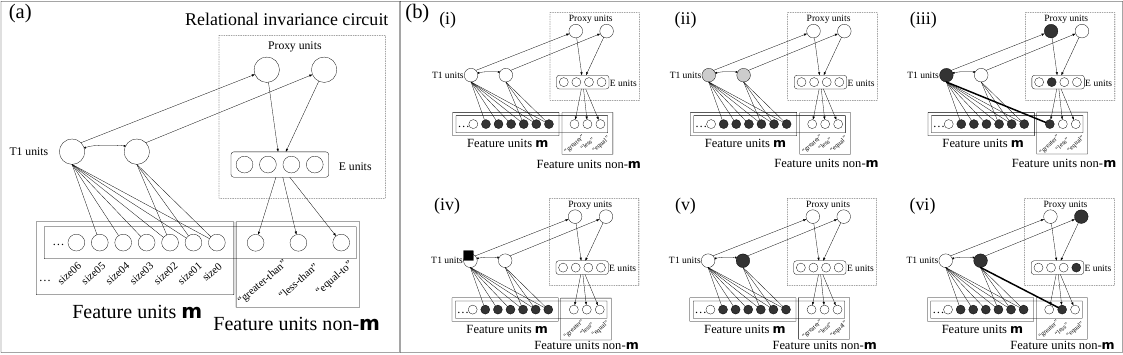}
\caption{(a) The relational invariance circuit. (b) (i) Activation flows from clamped feature units encoding a dimension or property to T1 units (dark grey units indicate more active units).(ii) T1 units compete via lateral inhibition to respond to the active feature units (light grey units indicate less active units). (iii) T1 units activate proxy units, which feed activation to E units. E units pass activation to a subset of feature units non-m and connections between active feature units non-m and T1 units are updated via Hebbian learning. (iv) The active T1 unit is inhibited to inactivity by its inhibitor (black square). (v-vi) The process repeats for the second active T1 unit.}
\label{fig:4}
\end{figure*}

The circuit begins with a rate-coded representation of a metric dimension, \textit{m} (e.g., size, or location in the horizontal dimension of the visual field; "Feature units m" in Figure \ref{fig:4}a). These units share bidirectional excitatory connections with a collection of T1 units (e.g., role units in DORA), which mutually inhibit one another, and each of which excites a single "proxy" unit. The proxy units in turn excite a collection of four \textit{E} units (for their relation to Eq. \ref{eq:1}), which excite feature units outside the set of features representing \textbf{m} ("Feature units non-\textbf{m}" in Figure \ref{fig:4}a). All other connections depicted in Figure \ref{fig:4} start with weights of 1.0, with the exception of connections between Feature units non-\textbf{m} (henceforth non-\textbf{m} features) and \textit{E} units, and the connections between feature units non-\textbf{m} and T1 units. The connections to and from feature units change during learning. Initially, the connection weights, $w_{ij}$, from each \textit{E} unit, \textit{j}, to a collection of 10 non-\textbf{m} features, \textit{i}, are initialized randomly with values between 0 and 1. As detailed below, after learning each \textit{E} unit is most strongly connected to a different subset of the non-\textbf{m} feature units, and these subsets become representations \textit{greater-than}, \textit{less-than}, and \textit{equal-to} (as in Figure \ref{fig:4}a). Connections between non-\textbf{m} feature units and T1 units are initially 0. 

\subsubsection{Circuit Operation} 

Processing in the relational invariance circuit is assumed to begin after perceptual processing has segmented the image into objects and encoded each object in terms of its various attributes (e.g., size, location in the horizontal and vertical dimensions, etc.). As elaborated under Simulations, DORA is equipped with a simple perceptual preprocessor that accomplishes this segmentation and encoding. The T1 units depicted in Figure \ref{fig:4} correspond to a subset of this encoding (i.e., representing each object’s size). For the purposes of illustration, we shall assume that each T1 unit in Figure \ref{fig:4} represents the size of one object (in a display containing two objects, \textit{a} and \textit{b}). (The T1 units in this circuit are otherwise identical to other T1 units in DORA.) For simplicity, we also assume that the pattern of activation on the feature units representing dimension \textbf{m} is the superposition (i.e., sum) of the vectors, \textbf{m}$_i$, representing the sizes of the two objects in the display. The learning algorithm does not require the feature inputs,  \textbf{m}$_i$, to be segmented into separate objects, \textit{i}; instead, it is sufficient to encode this information in the connections from \textbf{m}$_i$ to T1\textit{\textsubscript{i}}. 

Once the objects and their attributes are encoded by the preprocessor, the superimposed vector \textbf{m}$_t$ = \textbf{m}$_a$ + \textbf{m}$_b$ is clamped on the subset of feature units, \textbf{m}, representing \textbf{m} (Figure \ref{fig:4}bi). This vector serves as input to the T1 units, which compete via lateral inhibition to become active. The input to T1 \textit{unit} \textit{i} is given by: 

\begin{equation}
n_{i} = \sum_{j}^{}a_{j}w_{ij}-\sum_{k\neq i}^{}a_{k}-LI
\label{eq:2}
\end{equation} 
%equation 2

\noindent where $j$ are feature units connected to T1 unit $i$, $k$ are other active T1 units $k \neq\ i$, and $LI$ is the activation of the local inhibitor (a refresh signal given when no active T1 units are active in the driver \parencite[as in,][see Appendix A1.4 for details]{doumas2008theory, hummel1997distributed, hummel2003symbolic, horn1990excitatory, horn1991segmentation, usher1996modeling, von1992sensory}. Activation of T1 units is calculated as: 

\begin{equation}
\Delta a_{i} = \gamma n_{i}\left(1.1-a_{i}\right)-\delta a_{i} \\
\label{eq:3}
\end{equation} 
%equation 3

\noindent where $\Delta a_{i}$ is the change in activation of unit $i$, $\gamma=0.3$ is a growth parameter, $n_{i}$ is the net input to unit $i$ and $\delta=0.1$ is a decay parameter. At this juncture, a note about parameter values is warranted: All the standard DORA parameters have the same values as reported in previous papers, and previous work has shown that DORA’s behavior is robust to the values of these parameters \parencite[see][]{doumas2008theory}. Where we have had to introduce new parameters for the purposes of the relational invariant circuit, their values were set solely for the purposes of having the circuit’s behavior match the main idea expressed in Eq. \ref{eq:1}. No attempt was made to optimize their values. 

Since the vectors \textbf{m}\textsubscript{a} and \textbf{m}\textsubscript{b} are superimposed on the feature units, the T1 unit with more connections to the feature units (in this case, the unit coding for the larger size), say T1\textsubscript{a}, will initially win the inhibitory competition, inhibiting T1\textsubscript{b} to inactivity. For this reason, the T1 unit with the larger input vector will always win the initial inhibitory competition (Figure \ref{fig:4}bii). 

Each proxy unit, $i$, has a connection weight of $1$ from T1$_i$, and a weight of $0$ from all other T1$_{j \neq i}$. A proxy unit is simple binary threshold unit whose activation is given by: 

\begin{equation}
p_{i} =\begin{Bmatrix}
1, & n_{i}\geq 0.4 \\ 
0, & otherwise \\ 
\end{Bmatrix}
\label{eq:4}
\end{equation} 
%equation 4

\noindent where $p_i$ is the activation of proxy unit $i$. 

Input, $n_i$, to proxy unit $i$ is calculated as: 

\begin{equation}
n_{i} = \sum_{j}^{}a_{j}w_{ij}-\rho_{i}
\label{eq:5}
\end{equation} 
%equation 5

\noindent where $j$ is an active T1 unit and $\rho_{i}$ is the refraction of unit $i$. The refraction, $\rho_{i}$ is given:

\begin{equation}
\rho_{i} =\frac{1}{.1+\iota e^{x}}
\label{eq:6}
\end{equation} 
%equation 6

\noindent where $x$ is the number of iterations since unit $i$ last fired, and $\iota=1\times10^{-7}$ is a scaler. Proxy unit $i$ will be active if and only if T1$_i$ is active (i.e., $a_i > 0.4$) and proxy unit i has not recently been active.

\textit{E} units take their inputs from the proxy units. The connections from proxy units to \textit{E} units have temporal delays built into them, so that each E unit has a Gaussian receptive field in the three-dimensional space formed by the two T1 cells’ activations, plus time. Input to \textit{E} units is given by Eq. \ref{eq:5} such that $j$ are proxy units, and change in activation is calculated using: 

\begin{equation}
    \Delta a_{i} = \gamma e^{-\frac{(n_i-\theta_E)^2}{k^2}}-.1a_{i}-.5 \sum_{j}a_{j}-LI
\label{eq:7}
\end{equation}
%equation 7

\noindent where $\gamma$ is a growth parameter, $\theta_E$ is the threshold on unit $E$, $k=.2$, $j$ are all other units in $E$ $j \neq i$. The circuit contains four $E$ units each with a $\gamma$ of $.1$ or $.3$, and a $\theta_E$ of $1$ or $2$, such that all four combinations of $\gamma$ and $\theta_E$ values are present in a single $E$ unit. As a result, some $E$ units respond preferentially to proxy units firing early in processing, others respond preferentially to proxy units firing later, and still others respond preferentially to the two proxy units firing at the same time. Like the T1 units, $E$ units laterally inhibit one another to respond in a winner-take-all fashion so that only one $E$ unit tends to become active in response to any (temporally-extended) pattern of activation over the proxy units. 

$E$ units are randomly connected to a collection of feature units that are not part of the vector space $\mathbf{m}$ (Figure \ref{fig:4}). Active $E$ units, $i$, both excite the non-$\mathbf{m}$ feature units, $j$, and during learning, update their connections to them. Feature units connected to $E$ units update their input by: 

\begin{equation}
n_{i} = \sum_{j}^{}a_{j}w_{ij}
\label{eq:8}
\end{equation} 
%equation 8

\noindent where $i$ and $j$ are feature units and $E$ units, respectively. Feature unit activation is updated as:

\begin{equation}
a_{i} =\frac{n_{i}}{\max\left(n_{j}\right)}
\label{eq:9}
\end{equation} 
%equation 9

\noindent where $a_i$ is the activation of feature unit $i$, $n_i$ is the net input to feature unit $i$, and $\max(n_j)$ is the maximum input to any feature unit. There is physiological evidence for divisive normalization in the feline visual system \parencite[e.g.,][]{bonds1989role, heeger1992normalization, foley1994human} and psychophysical evidence for divisive normalization in human vision \parencite[e.g.,][]{thomas1997contrast}. While the circuit is being learned, connections between units in $E$ and feature units are updated by the equation: 

\begin{equation}
\Delta w_{ij} = a_{i}\left(a_{j}-w_{ij}\right)\gamma 
\label{eq:10}
\end{equation}
%equation 10

\noindent where $i$ and $j$ refer to units in $E$ and feature units, respectively, and and $\gamma=.3$ is the growth parameter. 

When the circuit is running, connections between the non-\textbf{m} feature units and active T1 units are updated by Eq. \ref{eq:10} (Figure \ref{fig:4}biii). As a result, T1$_a$ (the T1 unit we assumed won the initial inhibitory competition) will learn positive connections to whatever feature unit(s) the active $E$ unit has learned to activate (recall that the connections from $E$ units to non-\textbf{m} features are initially random). Importantly, the active non-\textbf{m} feature units, $j$, to which T1$_a$ has learned connections effectively represent \textit{greater-than}: These features will become active whenever the E unit that responds to early firing proxy units. In short, T1$_a$ has gone from representing a particular value on \textit{m} to representing the conjunction of that value (it is still connected to the features in \textbf{m}) along with the relational invariant \textit{greater-than}. 

Recall that the T1 units are oscillators (see also Appendix A). As a result, after some iterations have transpired with T1$_a$ active, that unit’s inhibitor will become active, inhibiting T1$_a$ and allowing T1$_b$ to become active (Figure \ref{fig:4}biv). The same operations described with respect to T1$_a$ will take place with respect to T1$_b$, which will learn connection(s) to feature unit(s) representing \textit{less-than} (those non-\textbf{m} features that are strongly connected to the $E$ unit that responds later in firing; Figure \ref{fig:4}bv-vi). 

After these operations have taken place on a single pair of objects, DORA will have constructed a representation of that pair of objects, each with a specific value on metric dimension \textit{m} (represented by T1$_a$ and T1$_b$ in Figure \ref{fig:4}) explicitly tagged with an invariant specifying that its value is either \textit{greater-than} (in the case of T1$_a$ or \textit{less-than} (T1$_b$) some other value on \textit{m}. Importantly, these representations do not yet constitute an explicit representation of the relation \textit{greater-than} (a, b), because they are not linked into a propositional structure (e.g., by T2 and T3 units) specifying that the individual roles, \textit{greater-than} and \textit{less-than}, are linked into a single relation. Moreover, the emerging representation of these roles, as instantiated in the feature units connected to T1$_a$ and T1$_b$, still retain a full representation of the specific metric values of T1$_a$ and T1$_b$ on \textit{m}. In other word, T1$_a$ and T1$_b$ do not represent \textit{greater-than} and \textit{less-than} in the abstract, but instead represent something closer to \textit{greater-than-and-value-a} and \textit{less-than-and-value-b}. However, this kind of \textit{almost-relational} representation, in which relational invariants are present, but (a) are still associated with other, nonrelational features (e.g., specific values on \textit{m}), and (b) are not yet composed into an explicitly relational structure, are precisely the kind of representations DORA uses as the starting point for learning explicitly structured relations (see above). 

To illustrate, consider what will happen when a different pair of objects, \textit{c} and \textit{d}, are engaged in the process described above. For the purposes of illustration, assume that \textit{c} has a larger value on \textit{m} than \textit{d} does, but both have different values than \textit{a} and \textit{b}. The processes described previously will attach T1$_c$ to the same invariant \textit{greater-than} feaure(s) as T1$_a$ and T1$_d$ to the same \textit{less-than} feature(s) as T1$_b$. It is in this sense that those feature units represent \textit{greater-than} and \textit{less-than}, respectively: The relational invariance circuit will, by virtue of the E unit-to-feature connections learned in the context of objects \textit{a} and \textit{b}, be biased to activate the feature unit(s) recruited for \textit{greater-than} in the context of T1$_a$ in response to the "\textit{greater-than}-ness" of T1$_c$, and the feature unit(s) recruited for \textit{less-than} in response to the "\textit{less-than}-ness" of T1$_d$. If, subsequently, DORA compares the pair [\textit{a}, \textit{b}] to the pair [\textit{c}, \textit{d}], it will learn predicates (T1 units) strongly connected to \textit{greater-than} and \textit{less-than}, and only weakly connected to the specific values of \textit{a} and \textit{c}, and \textit{b} and \textit{d}, respectively (see above). After exposure to as few as two pairs of objects, DORA has started to explicitly predicate the invariant relation \textit{greater-than} (\textit{x}, \textit{y}) with respect to dimension \textit{m}. 

In fact, DORA has learned something much more general than that, because the invariant features \textit{greater-than} and \textit{less-than}, learned in the context of metric dimension \textit{m}, will generalize to any other rate-coded metric dimension, \textit{n} $\neq$ \textit{m}. The reason is that the rate code that serves as the input to the relational invariance circuit operates on the magnitudes of the T1 units’ inputs, regardless of their origin: So long as the T1 units in question are (a) coupled oscillators that compete with one another to become active, and (b) receive rate-coded inputs from whatever metric dimension they represent, so that (c) the T1 unit with the larger value fires earlier than the T1 unit connected to the smaller value, the T1 unit connected to the larger value of the dimension will become connected to \textit{greater-than} and the T1 unit connected to the smaller value on the dimension will become connected to \textit{less-than} (or, if the two T1 units encode the same value on \textit{n}, and become simultaneously co-active therefore activating the E unit that responds when multiple T1 units are active, they will become connected to the non-\textbf{m} features strongly connected to the active E unit, or \textit{equal-to}). 

This same property of the relational invariance circuit renders it vulnerable to incorrectly assigning the invariants \textit{more-than} and \textit{less-than} to \textit{any} pair of object properties (represented as T1 units) that get passed into the circuit, even if those properties do not lie on a metric dimension. For this reason, there need to be constraints on when the relational invariance circuit is invoked. One obvious constraint is that it should only be invoked when the property coded by a T1 unit is a value on a metric dimension. As discussed previously, additional constraints, for example, regarding which metric dimensions are most likely to invoke the circuit under what circumstances, are likely also important, but consideration of what those constraints are is beyond the scope of the current work \parencite[but see][]{spelke2007core}. 

\subsection{Learning the content and the format of relational representations}

By integrating the relational invariance circuit with the DORA algorithm for learning structured representations of relations  \parencite{doumas2008theory}, we have developed a single system that explains learning structured relational representations of similarity and relative magnitude from very simple (non-relational) beginnings without assuming any structured representations, or even relational invariants, \textit{a priori}. The model starts with flat feature vector representations of object properties. These vectors contain no relational features, just absolute information about properties and magnitudes along dimensions. As described above, when objects with these feature encodings are compared, invariant patterns emerge, which mark similarities and differences in featural encoding and absolute magnitudes. The relational invariance circuit exploits these patterns to identify relational instances and return invariant features identifying those relations. 

The DORA learning algorithm identifies invariant features of compared objects and learns structured representations of those features in a format akin to a single-place predicate. The model then links systematically co-occurring predicate-argument bindings to form functional multi-place predicate representations. That is, over a series of progressive comparisons, the model isolates collections of object features, represents these as functional single-place predicates, links systematically co-occurring single-place predicates to form multi-place predicates, and produces increasingly more refined versions of these representations. When the representational content of these objects is relational, DORA will learn structured representations of this relational content. 

\subsection{Mechanism for generalization}

We propose that operations on relational representations underlie human generalization and that generalization based on relations occurs in (at least) two ways. First, relational representations learned in one context are readily applicable to characterize new contexts. Relational representations are useful for characterizing multiple domains because the same relations apply across domains regardless of the objects involved. Second, theories and schemas learned from one domain allow us to make inferences about other domains using analogical inference. 

\begin{figure*}[!htbp]
\centering
\includegraphics[width=13.76cm,height=6.11cm]{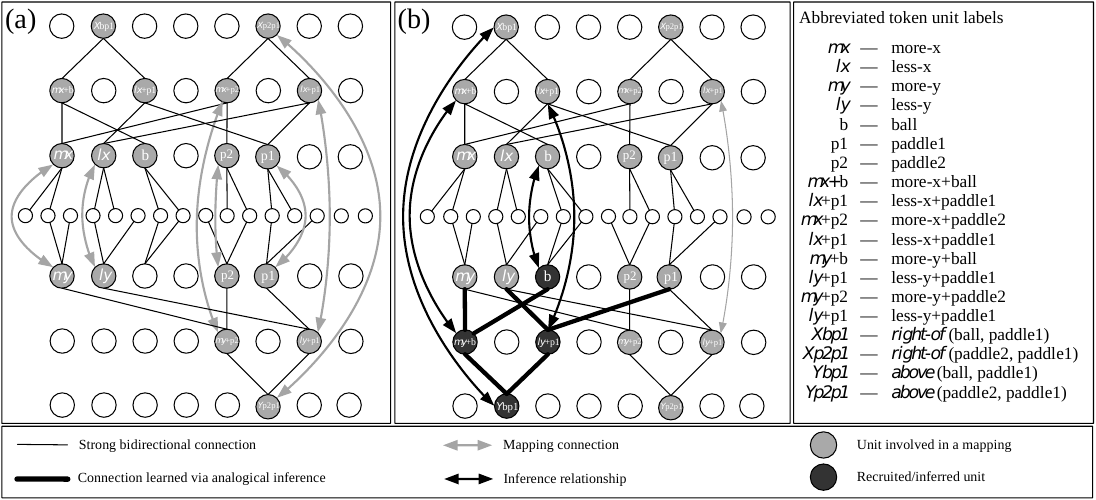}
\caption{Analogical inference in DORA. (a) The representation of the \textit{right-of} (ball, paddle1), and \textit{right-of} (paddle2, paddle1) in the driver maps to the representation of \textit{above} (paddle2, paddle1) in the recipient (grey double-arrowed lines indicate mappings). (b) As the representation of \textit{right-of} (ball, paddle1) becomes active in the driver, some active units have nothing to map to in recipient (the units representing ball, \textit{more-x}+ball, \textit{less-x}+paddle1, and \textit{right-of} (ball, paddle1)). DORA recruits and activates units to match the unmapped driver units (black units indicate recruited units; black double-arrowed lines indicate inference relationships). DORA learns connections between co-active token units in the recipient (heavy black lines). The result is a representation of the situation: \textit{above} (ball, paddle1) $\&$ \textit{above} (paddle2, paddle1) in the recipient. }
\label{fig:5}
\end{figure*}

Analogical inference—in this case, using a model of one domain to reason about another domain—follows directly from DORA’s mapping process. For example, suppose that DORA has learned about spatial relations (e.g., \textit{above}, \textit{right-of}, \textit{larger}) and then learned that when playing the game Breakout—where the goal of the game is to hit a ball with a paddle moving horizontally—relations between the ball and paddle predict actions to take. Specifically, DORA has learned that the state \textit{right-of} (ball, paddle1) supports moving right (i.e., \textit{right-of} (paddle2, paddle1); where paddle1 is the state of the paddle before the move, and paddle2 after the move), the state \textit{left-of} (paddle1, ball) supports moving left, and the state \textit{same-x} (ball, paddle1) supports making no move. When DORA encounters a game like Pong—where the goal of the game is to hit a ball with a paddle moving vertically—the moves available in Pong (up and down) might remind DORA of the moves available in Breakout (left and right). With the representation of the available Pong actions in the driver (e.g., \textit{above} (paddle2, paddle1), and representations of the Breakout strategy retrieved into the recipient (e.g., \textit{right-of} (ball, paddle)  \textit{right-of} (paddle2, paddle)), DORA performs analogical mapping. Because of the shared relational similarity, corresponding moves between Pong and Breakout map—e.g., \textit{above} (paddle2, paddle1) in the driver will map to \textit{right-of} (paddle2, paddle1) in the recipient (Figure \ref{fig:5}a, mappings depicted as double-headed arrowed lines; Appendix A2.2 and Appendix A4 for details of how such mappings are discovered). Generalization is performed on the basis of these mappings. 

During analogical generalization, propositions with unmapped elements enter the driver, and any propositions to which they map enter the recipient (information is generalized from the driver to recipient; see Hummel $\&$ Holyoak, 2003). So, if DORA has mapped \textit{above} (paddle2, paddle1) to \textit{right-of} (paddle2, paddle1), the representation of the rule from Breakout enters the driver, and the mapped representation of the move from Pong enters the recipient. Figure \ref{fig:5}b depicts a case where the rule \textit{right-of} (ball, paddle)  \textit{right-of} (paddle2, paddle) is in the driver, and the mapped \textit{above} (paddle, paddle2) is in the recipient. When a unit \textit{i}, in the driver learns an excitatory mapping condition to a given unit \textit{j}, in the recipient, it also learns a global inhibitory mapping connection to all other units, $k \neq\ j$, in the recipient. Similarly, \textit{j} learns a global inhibitory connection to all units $i \neq\ l$ in the driver. These global inhibitory connections play a vital role in the inference process. Continuing the example, all units encoding the representation of Pong in the recipient have a positive mapping connection to a corresponding unit in the driver (i.e., \textit{right-of} maps to \textit{above}, paddle1 maps to paddle1 and paddle2 maps to paddle2; Figure \ref{fig:5}). By consequence, all units in the recipient also have a global inhibitory mapping connection to all other units in the driver. Therefore, when \textit{more-x} (ball) becomes active in the driver (along with the T2 unit encoding \textit{more-x}+ball and the T3 unit encoding \textit{right-of }(ball, paddle1)), the T1 unit encoding ball inhibits all T1 units in the recipient (except for \textit{more-y}, which is excited by \textit{more-x}), the T2 unit encoding \textit{more-x}+ball inhibits all T2 units in the recipient, and the T3 unit representing \textit{right-of }(ball, paddle2) inhibits all T3 units in the recipient. 

This form of generalized inhibition occurs when all units in the recipient map to some unit in the driver, and no units in the recipient map to the currently active driver units. That is, the signal indicates that there are elements in the driver that map to nothing in the recipient. This occurrence signals DORA to initiate analogical inference. During analogical inference, DORA recruits and activates units in the recipient that corresponds to the unmapped (i.e., inhibitory) unit in the driver (e.g., DORA recruits a T1 unit in the recipient corresponding to an unmapped T1 unit in the driver). Newly recruited units are assigned positive mapping connections with the driver units that initiated their recruitment (black double-headed arrows in Figure \ref{fig:5}b), and they learn connections to other recipient units by simple Hebbian learning (e.g., active T1 units learn connections to active T2 units and active feature units; thick black lines in Figure \ref{fig:5}b). When \textit{more-}x (ball) is active in the driver, the driver T1 unit representing \textit{more-x} activates the recipient T1 unit \textit{more-y}, as they map, however, the active ball T1 unit, \textit{more-x}+ball T2 unit, and \textit{above} (ball, paddle1) T3 unit will activate nothing and inhibit all recipient units (as they map to nothing in the recipient). In response to this generalized inhibition in the recipient, DORA recruits a T1 unit corresponding to the unmapped active driver T1 unit (representing ball), a T2 unit corresponding to the active unmapped driver T2 unit (representing \textit{more-x}+ball), and a T3 unit is recruited corresponding to the active unmapped driver T3 unit (representing \textit{right-of} (ball, paddle1); black units in Figure \ref{fig:5}b). The recruited T1 unit learns connections to the active features of ball and to the recruited T2 unit (as they are all co-active; thick black connections in Figure \ref{fig:5}b). The recruited T2 unit learns connections to the recruited T3 unit (as they are co-active), and then to the T1 unit representing \textit{more-y} when it is active. As such, the recruited T1 unit becomes a representation of ball, and the recruited T2 links the representation of ball and \textit{more-y} (\textit{more-y}+ball). Similarly, when \textit{less-x} (paddle1) becomes active in the driver (activating the \textit{less-y} and paddle1 T1 units in the recipient), a T2 unit will be recruited to match the unmapped active driver T2 unit. That T2 unit will learn connections to the \textit{less-y} and paddle1 T1 units, and to the active recruited T3 unit (which remains active as its corresponding driver T3 unit remains active). The result is a representation of \textit{above} (ball, paddle1) in the recipient (Figure \ref{fig:5}b). 

\begin{figure}[!htbp]
\centering
\includegraphics[width=7.5cm]{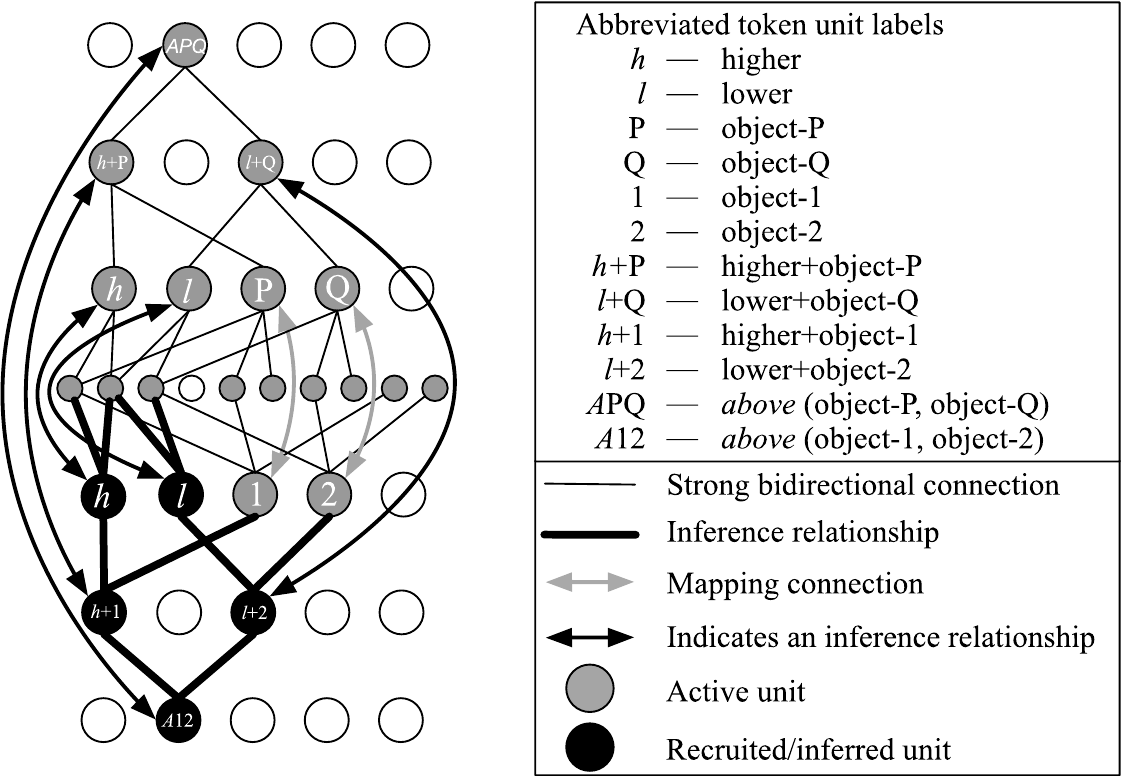}
\caption{Generalizing known relations to new situations using the analogical inference algorithm in DORA. The representations of object-P and object-Q in the driver map to the representations of object-1, object-2 in the recipient (grey double-arrowed lines). As the representations of \textit{higher} (object-P) and \textit{lower} (object-Q) become active in the driver, some active units have nothing to map to in recipient (e.g., the driver units representing \textit{higher} and \textit{higher}+object-P, and the units representing \textit{lower} and \textit{lower}+object-Q). DORA recruits and activates units to match the unmapped driver units (black units indicate recruited units; black double-headed arrows indicate inference relationships). DORA learns connections between co-active token units in the recipient (heavy black lines). The result is a representation of \textit{above} (object-1, object-2) in the recipient. }
\label{fig:6}
\end{figure}

The same process also accounts for how we might predicate (or explicitly represent) known relations about new situations. As a simple example, suppose DORA encounters two objects involved in a relation such as when object-1 is above object-2. Those objects will have properties of that relation (e.g., object-1 might have features such as "more" and "y"; delivered by the relation invariance circuit). If DORA has learned explicit structured representations of the relation \textit{above} (e.g., two linked predicates strongly connected to the features "more" and "y" and "less" and "y" respectively), then it might retrieve a representation of that relation from LTM, say \textit{above} (P, Q). The retrieved relational representation can then be projected on to object-1 and object-2 (Figure \ref{fig:6}). Specifically, P, the higher item, will correspond to object-1, and Q, the lower item, will correspond to object-2. Based on these correspondences, DORA will infer the predicates bound to P and Q about object-1 and object-2 via the analogical inference algorithm. Shifting focus to the \textit{above} (P, Q) proposition (\textit{above} (P, Q) is in the driver; Figure \ref{fig:6}), when the \textit{higher} unit becomes active it inhibits all units in the recipient, signalling DORA to recruit a T1 unit in the recipient to match the active \textit{higher} unit and a T2 in the recipient to match the active \textit{higher}+P unit. The recruited T1 unit learns connections to the active feature units, becoming an explicit representation of \textit{higher}, and the recruited T2 unit learns connections to the recruited T1 unit (representing \textit{higher}) and the object-1 unit when it becomes active (Figure \ref{fig:6}). Similarly, a representation of \textit{lower} is represented about object-2. The result is the known relation \textit{above} predicated about object-1 and object-2 (Figure \ref{fig:6}). 

In the following simulations we demonstrate the efficacy of the computational account of relation learning and relation-based generalization that we have proposed. We show how DORA learns structured relational representations from simple visual non-structured non-relational inputs, and how it then uses these representations to support human-level cross-domain generalization—by characterizing novel domains in terms of known relations, and then driving inferences about the new domain based on the systems of relations learned in previous domains. Additionally, we demonstrate that DORA captures several key properties of human representation learning and the development of generalization.

\section{Simulations}

As described previously, explaining cross-domain generalization as analogical inference entails explaining how a system: (1) detects (or learns to detect) relational content; (2) learns structured representations of that relational content; (3) uses these representations to characterize and behave in the domains it experiences (e.g., to build models of the domain to guide behavior); (4) uses the representations learned from previously experienced domains to make analogies and subsequently inferences about new domains. In addition, there is a distinction between representing a domain and acting on those representations. Extrapolation from one domain to another relies on both using representations learned in one domain to characterize another (i.e., \textit{representational} transfer), and adopting strategies from one domain for use in another (i.e., \textit{policy} transfer). 

Below we report a series of simulations evaluating these various capacities of the model. In simulation 1, we show that the model learns structured relational representations—both their form and content—from non-structured and non-relational visual inputs without assuming a vocabulary of structured representations or relational features a priori. In simulations 2-4, we show that the model can be integrated with methods for reinforcement learning to use the representations that it learns to build more complex models (or policies) for behaving in the domain, and then use its representations to perform zero-shot (i.e., first trial) cross-domain generalization. Specifically, we show that after the model learns to play one video game (Breakout), it can generalize its knowledge to play a structurally similar but featurally very different game (Pong). Moreover, we show that generalization in the model relies exquisitely on the structured format of the representations that it learns. In simulation 5, we evaluate whether the representations that the model learned in previous simulations generalize to more complex tasks like adult analogy problems, support generalization to completely novel stimuli (i.e., approximating universal generalization), and meet the hallmarks of human relational cognition. Finally, we are proposing an account of human generalization that includes relational representation learning. As such, it should be the case that our model mirrors the capacities of children as they learn relational representations. In simulations 6 and 7 we use the learning trajectory the model underwent during simulations 2-4 to simulate studies from the developmental literature on children’s magnitude reasoning (simulation 6) and relational problem solving (simulation 7). Additionally, simulations 6 and 7 provide further tests of the model’s capacity for cross-domain generalization: After learning representations in a domain like Breakout, the model extends these representations to reason in the domain of a psychology experiment. We show that the model not only generalizes the representations that it learns from one domain to reason about a new task (as children do when they enter the laboratory) but also goes through the same behavioral trajectory as children. Details of all simulations appear in Appendix B.

\subsection{Visual front end}

Generalizing DORA to work with perceptual inputs, such as pixel images necessitated supplying the model with a basic perceptual front end capable of segmenting simple objects (e.g., paddles and balls) from visual displays. We endeavored to keep this extension as simple as possible, importing existing solutions. 

We used a visual pre-processor that delivers object outlines using edge detection (via local contrast) with a built-in bias such that enclosed edges are treated as a single object. In brief, the pre-processor identifies "objects" (enclosed edges) and represents them in terms of their location on the "retina", size, and color. This information roughly corresponds to the total retinal area of the object and the enervation of the superior, inferior, lateral, and medial rectus muscles in reaching the (rough) center of the object from a reference point \parencite[see][]{demer2002orbital}. The information is encoded as the raw pixels and direction (specific muscle) between the rough object center and the reference point, and the RGB encoding of the pixels composing the object. One consequence of this encoding is that the model shows the same bias to classify along the cardinal directions observed in humans \parencite{girshick2011cardinal}. 

This pre-processor is clearly a vast oversimplification of human perception. However, we chose it because it is adequate for our current purposes, it is computationally inexpensive, and the representations it generates are at least broadly consistent with what is known about human vision. For example, the visual system detects edges by local contrast \parencite[e.g., ][]{marr1980theory}, represents objects and their spatial dimensions \parencite[e.g., ][]{wandell2007visual}, and these representations and the visual image are quasi-homomorphic \parencite[e.g.,][]{demer2002orbital, engel1994fmri, furmanski2000oblique, moore2001neural}. We certainly do not claim the pre-processor is an accurate model of human vision; only that it is not grossly inconsistent with what is known about biological vision, and that it is adequate to our current goal, which is to model learning and generalization in the domain of simple visual images, with generalization to novel domains. The output of the visual pre-processor adds two assumptions to the model: (a) that the visual system can individuate objects, and (b) that dimensional information is rate-coded. Finally, it is possible to use the pre-processor as a front-end to both DORA and to the comparison DNNs, allowing us to equate the inputs used by DORA with those used by the DNNs. 

\subsection{Simulation 1: Unsupervised discovery of relations}

The goal of simulation 1 was to evaluate the capacity of the model to learn, without supervision, structured representations of relational concepts from non-structured representations of objects that include only absolute (non-relational) information. We ran two simulations, 1a and 1b, testing the model’s capacity to learn from both simpler and more featurally complex stimuli. 

\subsubsection{Simulation 1a}

Simulation 1a served as a basic proof of concept. In this simulation, we tested whether DORA would learn structured relational concepts when presented with simple visual displays. 

\begin{figure}[ht]
\begin{center}
\includegraphics[width=0.4\linewidth]{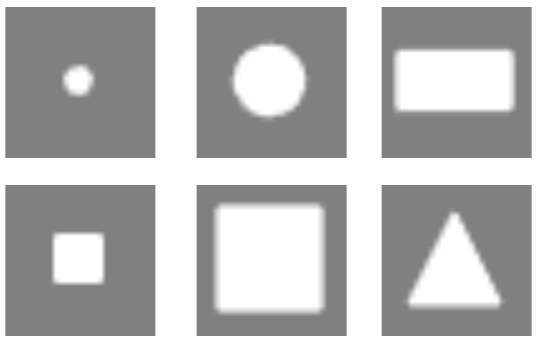}
\end{center}
\caption{Examples of the shape stimuli used for Simulation 1.}
\label{fig:7}
\end{figure}

\paragraph{Visual displays} We started with 150 two-dimensional images, each differing in shape, contrast, size, width, and height (see Figure \ref{fig:7} for examples of the images). Each of the 150 shapes was then randomly grouped with between 1 and 4 other shapes to create a total of 150 multi-object displays. The displays were processed by the visual preprocessor. As described above, the visual preprocessor identified an object as any item with a continuous and connected edge and represented that object as a T1 unit connected to a collection of features corresponding to the pixels composing its absolute width and height (x- and y-extent), size, and vertical and horizontal deviation from the edge of the screen (x- and y-deviation). As the images were greyscale, we left out the RGB information in this simulation. To add extraneous noise to the objects, each T1 unit encoding an object was also randomly connected to 10 "noise" features from a set of 1000. The result was 150 "scenes" containing between 2 and 5 objects, with each object represented as a set of absolute rate-coded spatial dimensions and noise features. 

\paragraph{Learning structured relational representations} We designed this simulation to mimic a child noticing a visual display (e.g., a scene) and attempting to use their memory of previous experiences to understand and learn about that display. DORA started with no representations (i.e., all weights set to 0). The representations of the 150 multi-object scenes were placed in DORA’s LTM. DORA attempted to learn from these stimuli, but it did not otherwise have any "task" to perform, and it received no feedback on its performance during the simulation. Rather, DORA performed 3000 "learning trials". On each learning trial, DORA randomly selected one collection of objects from LTM and placed that collection in the driver, thus simulating the perception of a visual display. DORA ran the driver representations through the relational invariance circuit, and then performed memory retrieval, analogical mapping, and representation learning (as described previously and in Appendix A). For the current simulations, we constrained DORA’s retrieval algorithm to favor more recently experienced displays \parencite[such recency effects are common in the memory literature; e.g.,][]{logie2020working}. With probability .2, DORA attempted to retrieve from the last 100 analogs that it had learned, otherwise it attempted to retrieve from LTM generally. 

In evaluating DORA’s learning, what we wanted to know was whether the model learned structured representations of relation content. That is, we wanted to know first, if the model had learned T1 units connected strongly to features defining a specific relational role concept (and only weakly to other features), and second, whether the model linked representations of complementary relational roles into multi-place relational structures. For example, if DORA learned a representation of a T1 unit connected strongly to the features encoding "more" and "x-extent" and weakly to all other features, then it had learned a relative (relational) representation of \textit{more-x-extent}. Similarly, if the model learned a representation of a T1 unit connected strongly to the features encoding for "less" and "x-extent" and weakly to all other features, then it had learned a relative (relational) representation of \textit{less-x-extent}. Finally, if the model learned a full LISAese structure wherein these T1 units (one representing \textit{more-x-extent} and another representing \textit{less-x-extent}) were bound to objects via T2 units that were linked via a T3 unit (as in Figure \ref{fig:1}), then it had learned a structured multi-place relational representation. 

To this end, we first defined a set of meaningful relational roles that the model could learn given the input images. This list comprised the set of relative encodings of the absolute dimensional information returned by the visual preprocessor: That is, the features encoding "more", "less", and "same" paired with the encoding of x-extent (["more", "x-extent"], ["less", "x-extent"], ["same", "x-extent"]), y-extent (["more", "y-extent"], ["less", "y-extent"], and ["same", "y-extent"]), size (["more", "size"], ["less", "size"], and ["same", "size"]), x-deviation (henceforth x; (["more", "x"], ["less", "x"], and ["same", "x"]), and y-deviation (henseforth y; (["more", "y"], ["less", "y"], and ["same", "y"]). 

\begin{figure}[!htbp]
\centering
\includegraphics[width=8cm]{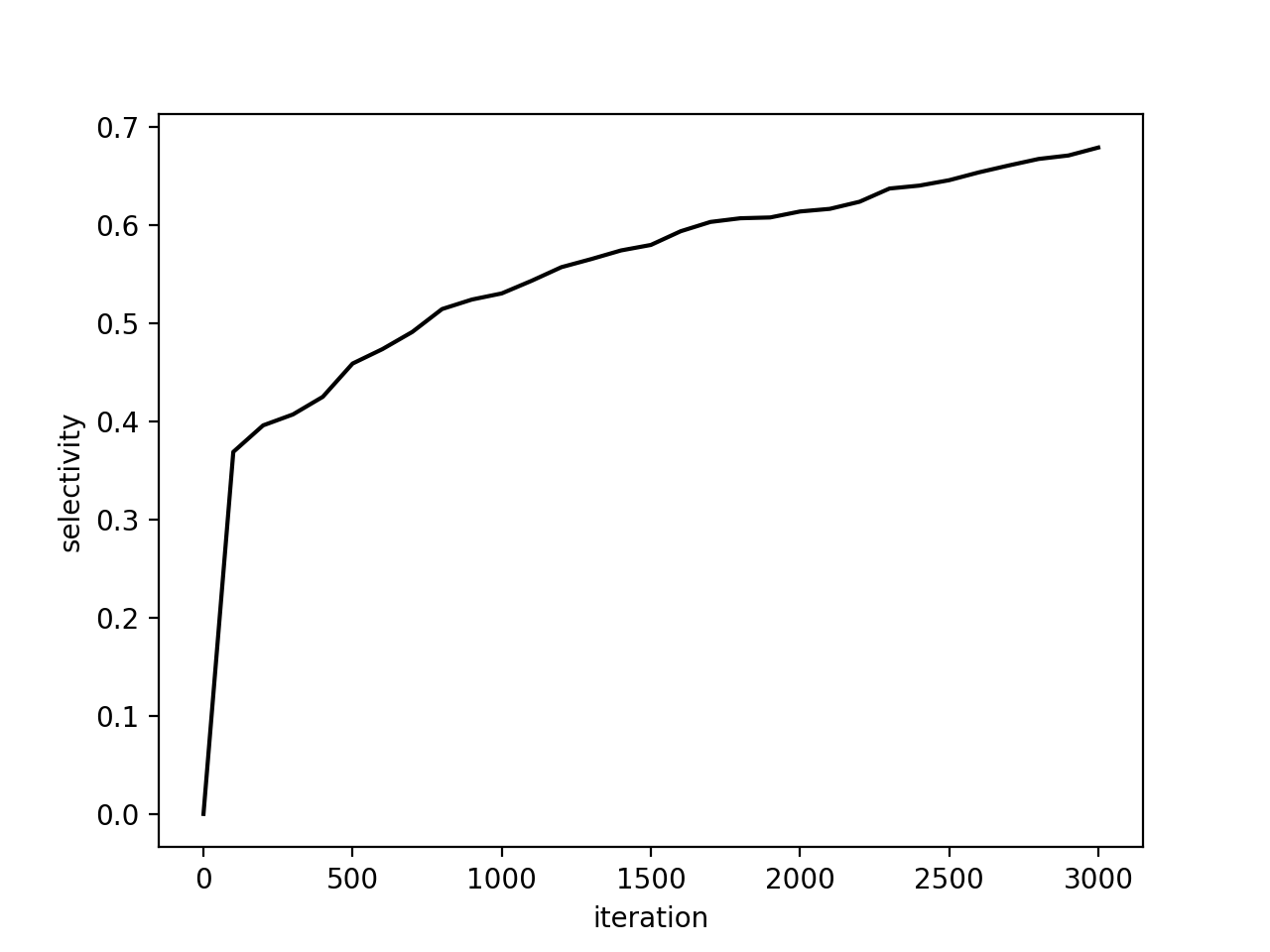}
\caption{Mean relational selectivity of T1 units (as defined in text) as a function of number of training examples, simulation 1a.}
\label{fig:mean_relational_selectivity_t1_units}
\label{fig:8}
\end{figure}

In order to evaluate DORA’s learning of the relations in the displays, we defined the \textit{relational selectivity} metric, $Q_{i}$, for a T$_1$ unit $I$ as: 

\begin{equation}
\begin{split}
r^{\prime} =  argmax_{r}\frac{1}{n}\sum_{j\in r}^{n}I \\
Q_{i} =\frac{\frac{1}{n}\sum_{j\in r^{\prime}}^{n}I_{i, j}}{1+\frac{1}{m}\sum_{k}^{m}I}
\end{split}
\label{eq:11}
\end{equation}
% equation (11)

\noindent where $ r^{\prime}$ is the relational role that maximizes the mean weight of unit $i$ to the features, $j=1\ldots n$ that make up the role’s content, and $k=1\ldots m$ are all other features. $Q_{i}$ scales with the degree to which unit $i$ codes selectively for a relational role, where $Q_{i}=1.0$ indicates that the unit responds exclusively to the features of a single relational role, $r^{\prime}$. We measured the relational specificity of the T1 units in DORA’s LTM over the course of 3000 learning trials (Figure \ref{fig:8}). As Figure \ref{fig:8} illustrates, DORA learned representations (T1 units) encoding meaningful relational roles. That is, DORA learned T1 units encoding roles like \textit{more-x-extent} (strongly connected only to features for "more" and "x-extent), \textit{less-y} (strongly connected only to "less" and "y"), or \textit{same-size} (strongly connected only to "same" and "size"). The results indicate that DORA’s learning algorithm produces representations that encode invariant relational content. 

\begin{figure}[!htbp]
\centering
\includegraphics[width=8cm]{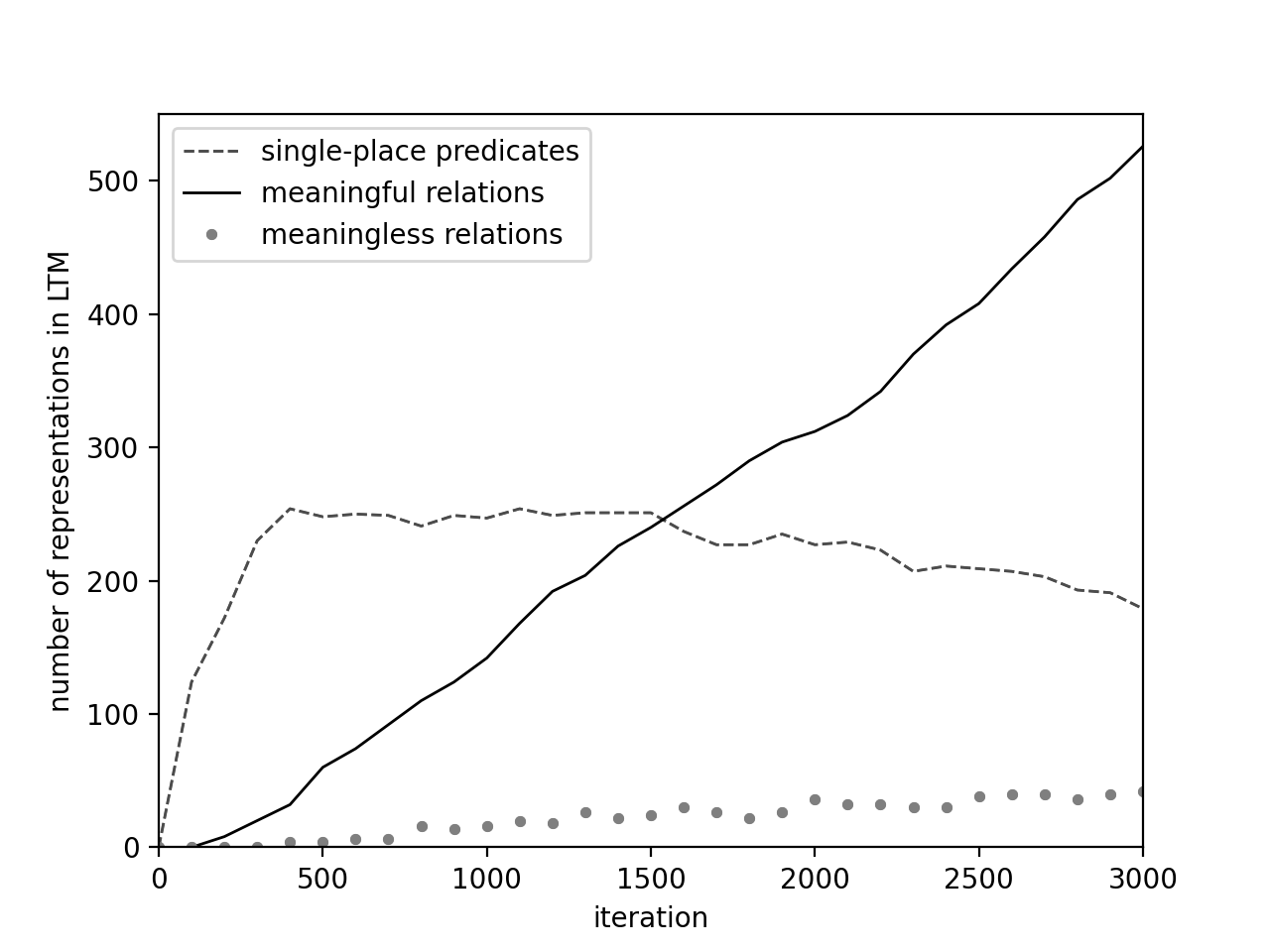}
\caption{The number of representations of single-place predicates, meaningful relations, and non-sense relational structures in DORA’s LTM after each 100 learning trails, simulation 1a.} 
\label{fig:9}
\end{figure}

Next, we checked whether the representations DORA learned were composed into meaningful relational structures (i.e., whether representations of complementary roles were linked into multi-place structures). For example, if DORA links \textit{more-y }(obj1) and \textit{less-y} (obj2) to form the relation \textit{above} (obj1, obj2), or links \textit{more-x-extent }(obj2) and \textit{less-x-extent} (obj1) to form the relation \textit{wider} (obj2, obj1), then it has learned representations of meaningful relations. To this end, we checked the number of representations in LTM representing single-place predicates (a learned T1 unit linked to another T1 unit via a T2 unit but not connected to a T3 unit), meaningful multi-place relations (T1 units representing complementary relational roles, each linked to an object T1 unit via T2 units that are also linked via a single T3 unit), and meaningless multi-place relations (T1 units not representing complementary relational roles, each linked to an object T1 unit via T2 units that are linked via a single T3 unit). As presented in Figure \ref{fig:9}, DORA learns representations of meaningful relations with experience. By the 1000\textsuperscript{th} learning trial, DORA had learned representations of all possible meaningful relations (i.e., \textit{above}, \textit{below}, \textit{same-vertical}, \textit{right-of}, \textit{left-of}, \textit{same-horizontal}, \textit{wider}, \textit{thinner}, \textit{same-width}, \textit{taller}, \textit{shorter}, \textit{same-height}, \textit{larger}, \textit{smaller}, \textit{same-size}), and it learned progressively more refined representations of these relations with additional learning trials. 

As described above, DORA learns multi-place relations by comparing sets of single-place predicates. During learning, this process runs in parallel with the discovery of the single-place predicates that will form the roles of these relations. However, because the linking operation depends on having a vocabulary of single-place predicates to combine, DORA necessarily follows a developmental trajectory in which it acquires single-place predicates before it acquires multi-place relations. As shown in Figure 9 roughly the first 300 learning trials are dominated by the discovery of single-place predicates like \textit{more-x}, \textit{less-x-extent}, and their complements. After that initial period, learning is dominated by the discovery of multi-place relations like \textit{above}. 

In contrast to error-correction learning (such as back propagation), DORA’s learning algorithm does not replace old knowledge (e.g., predicates discovered early in learning) with new knowledge (predicates learned later), but rather adds new knowledge to its existing knowledge. For example, the multi-place relations it learns do not replace the single-place predicates from which they were composed, and refined predicates and relations do not replace their less refined predecessors. However, as a consequence of DORA’s retrieval algorithm, less refined predicates become less likely to be retrieved (and thus used as the basis of new comparisons) than their (increasingly common) more-refined counterparts \parencite[the retrieval algorithm is biased in favor of retrieving the simplest pattern that fits the retrieval cue;][]{hummel1997distributed}. Effectively, the less refined predicates tend to fall out of service as they become obsolete. As a result, there is a difference between the predicates DORA knows (i.e., has stored in memory) and those it frequently uses. 

Moreover, DORA not only learns specific relations such as \textit{above} and \textit{below}, but it also discovers more abstract relations such as \textit{greater-than} and \textit{same-as} as a natural consequence. For example, when DORA compares two different instances of \textit{wider} (\textit{x}, \textit{y}), then it will learn a more refined representation of \textit{wider} (\textit{x}, \textit{y}) (as described previously). But if it compares an instance of \textit{wider} (\textit{x}, \textit{y}) to an instance of \textit{taller} (\textit{z}, \textit{w}), then it will learn a representation that retains what \textit{wider} has in common with \textit{taller}, or a generic \textit{greater} (\textit{a}, \textit{b}) relation. This result mirrors the development of abstract magnitude representations in children \parencite[e.g.,][]{sophian_2008}. 

In total, these results indicate that DORA learns structured representations of relative magnitude and similarity relations from unstructured (i.e., flat feature vector) representations of objects that include only absolute values on dimensions and extraneous noise features. However, a potential criticism of this simulation is that the starting representations are quite simple. Perhaps DORA only learns useful relational representations because each object only has five dimensions, whereas real objects have many more. Simulation 1b addresses this limitation.

\subsubsection{Simulation 1b: Scaling up}

In this simulation we tested DORA’s capacity to learn from messier examples containing more competing and extraneous information. We created 150 scenes each containing between 2 and 5 objects (using the same procedure as in simulation 1a). We then altered the objects in two important ways. First, we added 1000 distractor features from a pool of 100,000. Second, we added absolute encodings from 45 additional dimensions. That is, while in simulation 1a, each object was connected to rate-coded features encoding an absolute value on 5 dimensions (as delivered by the visual pre-processor), in this simulation each object was connected to rate-coded features encoding an absolute value on 50 dimensions (the five from simulation 1a, and 45 additional "dimensions"). As a consequence, each object was now much messier, containing not only more noise features, but also encoding more dimensions (that DORA could potentially learn explicit representations of). If DORA’s learning algorithm is indeed robust, then we would expect it to (a) learn relational representations of all dimensions (there should be nothing special about the 5 used in simulation 1a), and (b) learn these representations in a number of learning trials that scales proportionally with the number of items to be learned (i.e., the model should take roughly 10 times as long to learn comparably refined structured relational representations of 50 dimensions as it took to learn five). 

Simulation 1b proceeded like simulation 1a. A no-representations version of DORA was created. The representations of the 150 multi-object scenes were placed in DORA’s LTM. As in simulation 1a, DORA attempted to learn from these stimuli, but it did not otherwise have any "task" to perform, and it received no feedback on its performance during the simulation. In this simulation DORA performed 10,000 learning trials. 

\begin{figure}[!htbp]
\centering
\includegraphics[width=8cm]{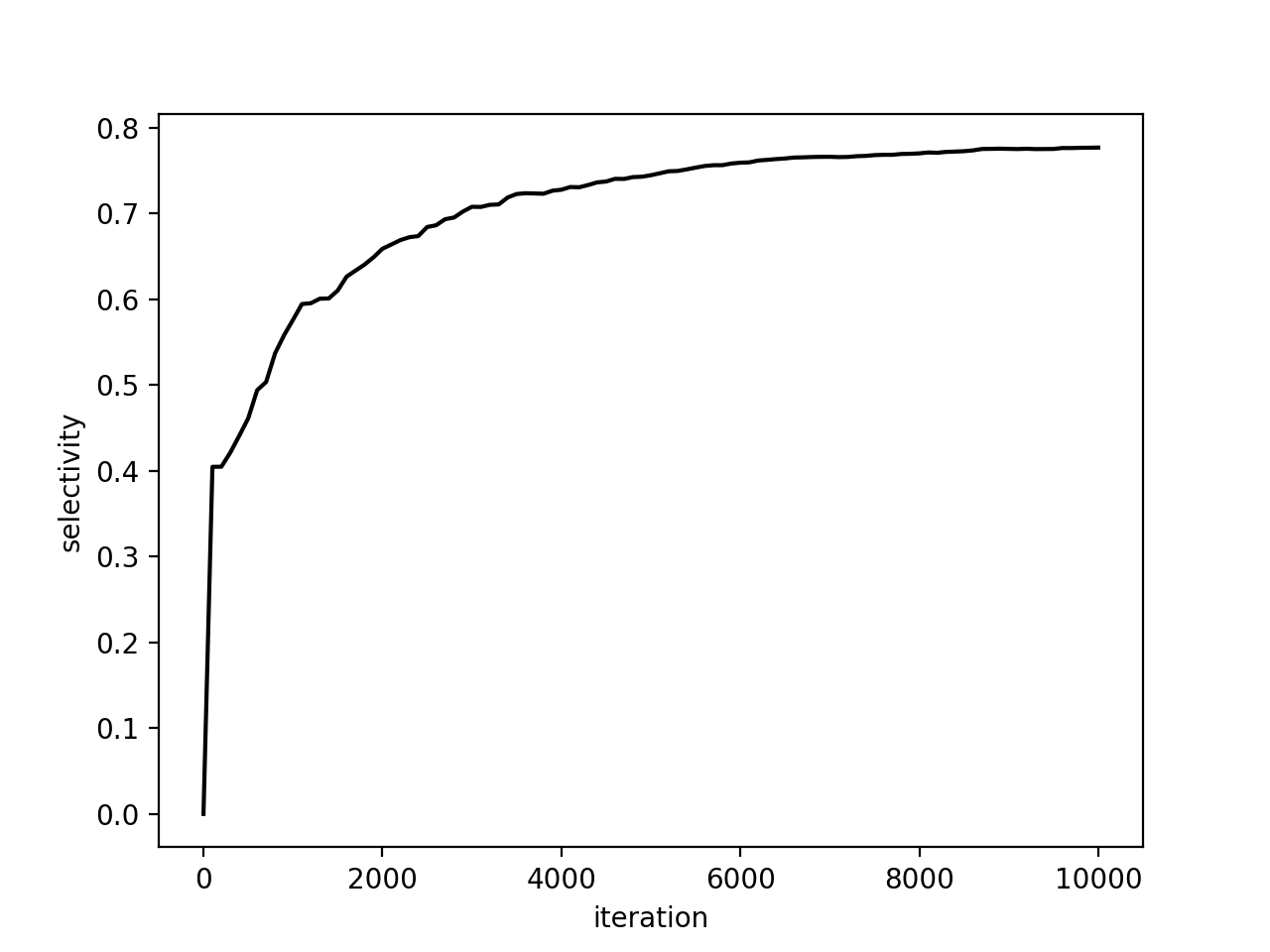}
\caption{Mean relational selectivity of T1 units (as defined in text) as a function of number of training examples, simulation 1b.}
\label{fig:10}
\end{figure}

\begin{figure}[!htbp]
\centering
\includegraphics[width=8cm]{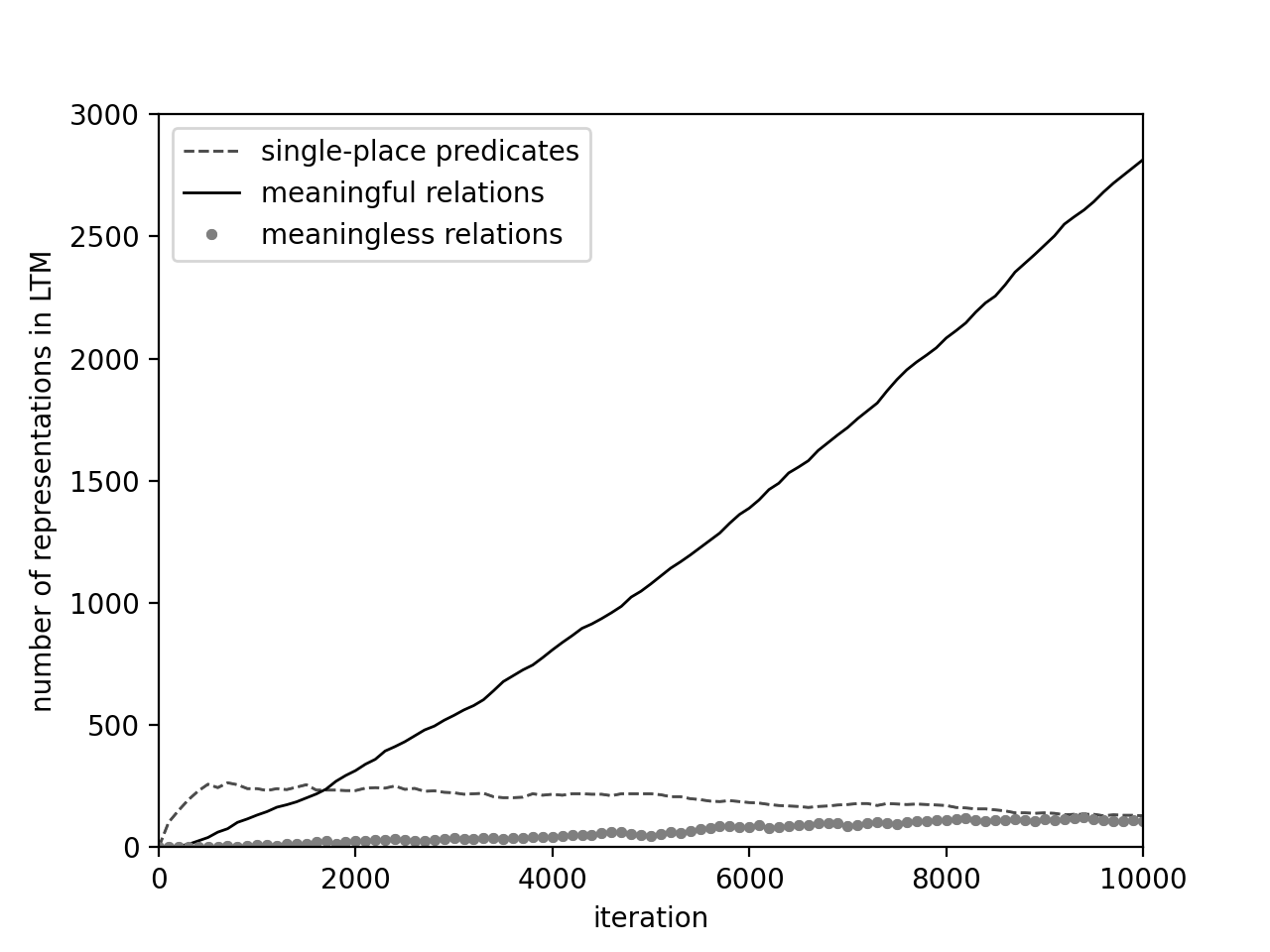}
\caption{The number of representations of single-place predicates, meaningful relations, and non-sense relational structures in DORA's LTM after each 100 learning trails, simulation 1b.}
\label{fig:11}
\end{figure}

Figure \ref{fig:10} shows the progression of the relational selectivity of DORA’s T1 units over the course of training. Just as in simulation 1a, DORA learned progressively more refined representations of relational content. Vitally, DORA learned structured relational (relative; more/less/same) representations of all 50 dimensions, as well as of general \textit{greater}, \textit{lesser}, and \textit{same}. In addition, as seen in Figure \ref{fig:10}, the number of learning trials required to learn refined representations of structured representations scales linearly. While DORA learned representations of all meaningful relations from five dimensions in roughly 1000 learning trials (simulation 1a), DORA learned meaningful relational representations of 50 dimensions in 10,000 trials. In addition, with more learning trials, the representations that DORA learned became progressively more refined. Just as in simulation 1a, after 3000 learning trials the relational selectivity in the model was just below 0.7, though with additional learning trials in simulation 1b, relational selectivity continued to increase. Finally, Figure \ref{fig:11} shows that just as in simulation 1a, DORA learned progressively more meaningful structured representations of relational representations with more learning trials and follows the same trajectory of learning single-place predicates first, followed by multi-place relational representations. 

The results of Simulation 1b show that DORA’s learning algorithm scales well with the complexity of the learning environment. Finally, it is worth noting that although DORA’s learning in simulations 1a and 1b was unsurprisingly slightly slow (recall that DORA received no feedback or guidance), \textcite{sandhofer2008order} showed that DORA’s learning accelerates (to the same rate as human learners) when the model receives the kind of general guidance children routinely receive from adults during the normal course of cognitive development \parencite[e.g., being guided to make specific comparisons in specific sequences;][]{sandhofer1999learning,sandhofer2000counting}. 

\subsection{Simulation 2: Cross-domain transfer in simple video games}

Our second simulation was designed as a test of cross-domain generalization. We wanted to evaluate whether the model could learn representations from a domain, use those representations to perform intelligently in that domain, and then transfer that knowledge to a new domain in a single shot (i.e., without any additional training). We used transfer between different video games as a case study. In this simulation, after DORA learned to play Breakout, we tested its capacity to generalize, without additional training, to Pong\footnote{ In this simulation we discuss DORA’s performance transferring from Breakout to Pong for expositional clarity. As shown in supplemental simulation 1, the generalization results were the same when DORA learned to play Pong and attempted to transfer to Breakout. } (which is a structurally analogous to Breakout but featurally quite different—among other differences, the player moves the paddle up and down in Pong but left and right in Breakout), and then tested its capacity to return to playing Breakout. 

For the purposes of comparison, we also trained four statistical learning systems, including (1) a Deep Q-learning Network \parencite[DQN;][]{mnih2015human} with the standard convolutional neural network front end; (2) a DQN with the same visual front end as DORA; (3) a supervised DNN with the same visual front end as DORA; and (4) a graph network \parencite[e.g.,][]{Battaglia2018RelationalIB} with the same front end as DORA. These controls allowed us to compare DORA to systems that do not have structured relational representations, and to control for the visual front-end and its assumptions: Networks 2-4 also had objects individuated and contained rate-coded dimensional information as inputs. 

In addition, we ran a small transfer study with humans. Unlike the current networks, humans come into the game situation with a wide range of knowledge beyond simple video games (let alone only Breakout), but an account of human generalization should be able to match the qualitative property that humans do transfer between things like games. To test whether humans do indeed generalize between games, human players either played 50 minutes of Breakout followed by 10 minutes of Pong, or the reverse. The results indicate cross-game transfer between Breakout and Pong and Pong and Breakout. Details appear in Supplemental Results. 

\subsubsection{Learning to play a game in DORA}

In this simulation, as in simulation 1, DORA started with no knowledge. To begin, DORA learned representations from Breakout game screens. Learning in this simulation proceeded as described in simulation 1, with the difference that we used game screens from Breakout rather than images of collections of 2D shapes. We allowed the system to play 250 games of Breakout making completely random responses, which produced game screens. These game screens ran through the visual pre-processor described above generating representations of scenes composed of simple objects (e.g., the paddle, the ball, the rows of bricks), which were stored in LTM. DORA attempted to learn from these stimuli performing 2500 learning trials. It did not have any "task" to perform, and it received no feedback during this part of the simulation. DORA successfully learned structured representations of \textit{above}, \textit{below}, \textit{same-vertical}, \textit{right-of}, \textit{left-of}, \textit{same-horizontal}, \textit{wider}, \textit{thinner}, \textit{same-width}, \textit{taller}, \textit{shorter}, \textit{same-height}, \textit{larger}, \textit{smaller}, \textit{same-size}, \textit{same-color}, and \textit{different-color} (screen images were colored). 

The next step was for DORA to use the representations that it learned from the game environment to engage intelligently with it. Several accounts of how relational representations, once available, may be used to characterize particular domains have been proposed \parencite[e.g.,][]{nye2020learning, lake2015human}. However, these probabilistic program induction approaches do not directly address the problem of building a relational model of the environment from a reward signal (these models are supervised). As learning to play video games entails learning to associate actions with states of the game based on reward signals (e.g., points), reinforcement learning methods \parencite{sutton2018reinforcement} are a natural starting point to solve this problem. Reinforcement learning (RL) has been widely applied to account for several aspects of human learning and exploratory behaviour \parencite[e.g.,][]{gershman2018deconstructing, otto2010regulatory, rich2018limits}. In tabular RL, which is the version of RL that we use in this simulation, the state-action space is represented as a table where the rows are defined by the individual states and the columns are defined by the actions. A known problem with tabular RL is that as the size of the table increases, learning becomes intractable. As relational representations can be combined, the size of the table grows exponentially on the number of relations considered when describing the state. Therefore, in our simulations we make the simplifying assumption that the agent knows what the relevant relations to build a model of the domain are (from the relations that the model had learned from game screens previously). By definition, a full account of how to build a relational model of the domain from the reward signal would need to solve the problem of selecting the relevant relations from a potentially very large set of relations. We return to this point in the general discussion. 

As mentioned above in our simulations, states were represented as the relevant relations to learn to play a game. In Breakout the relations considered were \textit{right-of, left-of, }and \textit{same-horizontal} applied over the paddle and the ball and the same relations applied over the ball at times \textit{t-1} and \textit{t}. For example, one state in Breakout could be [\textit{right-of} (ball1, paddle1)]. On the other hand, actions were represented as a relation between the object that the action was performed over at time \textit{t}, and the same object at time \textit{t}+1. For example, in Breakout the action \textit{move-right }was represented as \textit{right-of} (paddle2, paddle1), where paddle1 is the paddle before acting, and paddle2 is the paddle after. To associate actions with states of the game we augmented DORA with the capacity for reinforcement learning. Specifically, we used tabular \textit{Q}-learning \parencite{watkins1989learning}. RL algorithms seek to maximize the expected discounted cumulative reward, or return, by interacting with the environment. In each iteration of this process the environment produces a state $S_t$ and a reward $R_t$ and the agent takes an action $A_t$ in response. The goal of reinforcement learning is to find the optimal policy $\pi^*$ that maximizes the return. To do this Q-learning utilizes action-values as the basis for this search. The action-value of a state-action pair $Q(s, a)$ is the return when the agent is in state $s$ at time $t$, $S_t=s$ and takes action $a$, $A_t=a$. Q-learning follows an epsilon-greedy policy, where most of the time the action is selected greedily regarding the current action values and with a small probability the action is selected randomly, while updating the action-values according to the equation: 

\begin{equation}
    Q(S_{t},A_{t})\leftarrow Q(S_{t},A_{t}) + \alpha  
\Big[R_{t+1} + \gamma \max_{a}Q(S_{t+1},a) - Q(S_{t},A_{t})\Big]
\label{eq:12}
\end{equation}
% Equation (12)

\noindent where $gamma$ is a discount factor (in all simulation we set this value to .99). 

Applied iteratively, this algorithm approximates the true action-values and, therefore, the output policy (greedy regarding these values) will approximate the optimal policy. We trained DORA for 1000 games using tabular Q-learning. The model learning to associate relational states with relational representations of the available actions. Importantly for our purposes, because the states are relational the resulting policy corresponds to a set of relational rules that can be used as a basis for analogical inference (see below). 

\subsubsection{Generalizing to a new game in DORA}

As described above, analogical inference occurs when a system uses analogical correspondences between two situations to flesh out one situation based on knowledge of the other. This method is precisely how DORA infers how to play a game like Pong based on its experience with a game like Breakout. While learning to play Breakout, DORA had learned that relations between the ball and paddle predicted actions. Specifically, DORA learned that the state \textit{right-of} (ball, paddle1) supported moving \textit{right} (i.e., \textit{right-of} (paddle2, paddle1)), that the state \textit{left-of} (paddle1, ball) supported moving \textit{left}, and that the state \textit{same-x} (ball, paddle1) supported making \textit{no move}. With the representation of the available Pong actions in the driver, these Breakout representations could be retrieved into the recipient. DORA then performed analogical mapping. Because of the shared relational similarity, corresponding moves between Pong and Breakout mapped—e.g., \textit{above} (paddle2, paddle1) in the driver mapped to \textit{right-of} (paddle2, paddle1) in the recipient. DORA then performed analogical generalization on the basis of these mappings. 

As described previously and illustrated in Figure \ref{fig:5}, after mapping the moves in Breakout and Pong, DORA infers the relational configurations that might reward specific moves in Pong based on the relational configurations that reward specific moves in Breakout. For example, given that \textit{right-of} (ball, paddle) tends to reward a\textit{ right} response (\textit{right-of} (paddle2, paddle1) in Breakout, and the mapping between the \textit{right} response and the \textit{up} response (\textit{above} (paddle2, paddle1) in Pong, DORA inferred that \textit{above} (ball, paddle) tends to reward a\textit{ up} response in Pong. The same process allowed DORA to generalize other learned rules (e.g., \textit{below} (paddle2, paddle1) then move \textit{down}). 

Importantly, like DORA’s representation learning algorithm, mapping and analogical inference are completely unsupervised processes: The model discovers the correspondences between the games on its own, and based on those correspondences, makes inferences about what kinds of moves are likely to succeed in the new situation.

\subsubsection{Simulation results}

The DQN, the DQN with the same visual front end as DORA, and the graph network were all trained for 31,003, 20,739, and 10,000 games respectively. The DNN was trained via back-propagation for 4002 games. Models that use structured representations often require far fewer training examples than networks trained with traditional feature-based statistical learning algorithms \parencite[see e.g.,][]{bowers2017parallel,hummel2011getting}. It is therefore unsurprising that DORA learned to play Breakout much faster than the other networks. 

\begin{figure*}[!htbp]
\centering
\includegraphics[width=13.76cm,height=5.21cm]{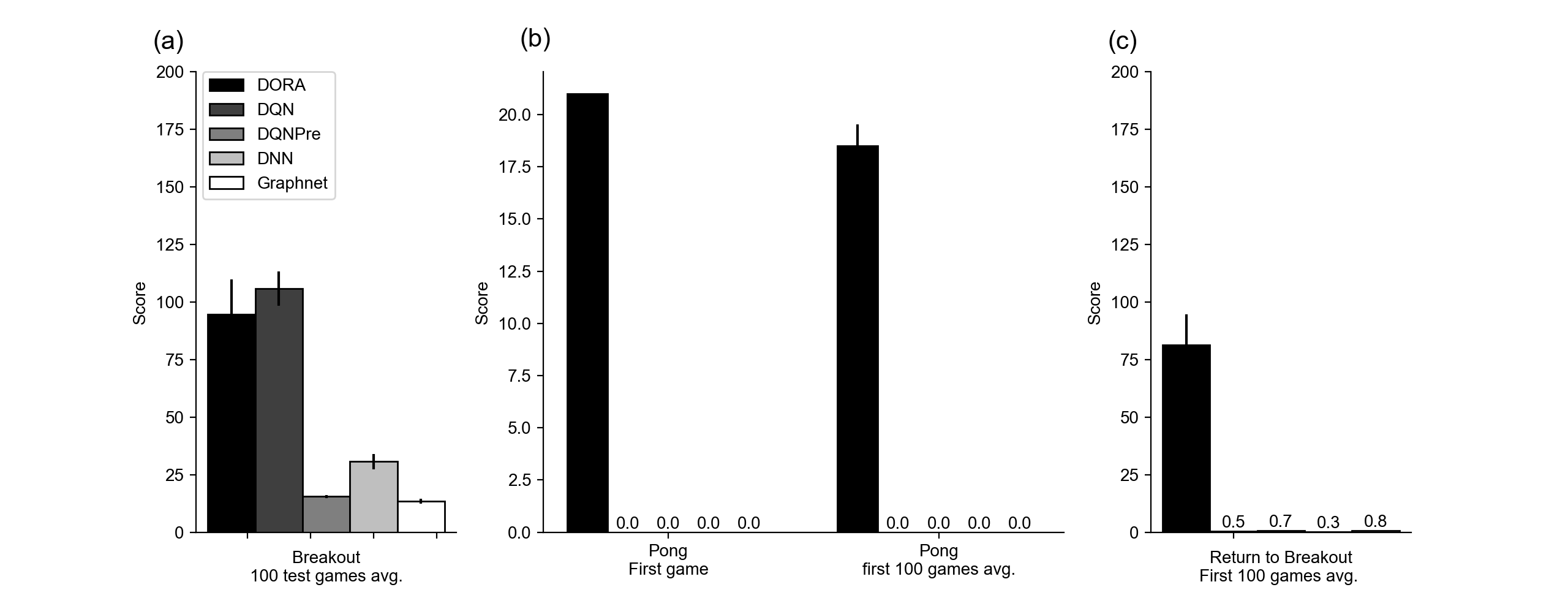}
\caption{Results of game play simulations with DORA, the DQNs and the DNNs. Error bars represent 2 stderrors. (a) Performance humans and networks on Breakout as an average of 100 test games. (b) Results of networks playing Pong after training on Breakout as score on the first game played and mean score over the first 100 games played. (c) Results of networks when returning to play Breakout after playing or learning to play Pong as an average of the first 100 games played.}
\label{fig:12}
\end{figure*}

Figure \ref{fig:12}a shows the mean score over the last 100 games of Breakout for all five networks. As expected, all the networks performed well. 

We then had the networks play a new game, Pong, for 100 games. Figure \ref{fig:12}b show the models’ zero-shot (i.e., immediate) transfer from Breakout to Pong. DORA performed well on the very first game of Pong (left columns) and over its first 100 games (beyond making an analogical inference, DORA did not engage in any additional learning during these test games). That is, DORA demonstrated \textit{zero-shot transfer} between the games: Having learned to play Breakout, generalized to how to play Pong. In contrast, the statistical learning algorithms did not transfer from Breakout to Pong: Having learned to play Breakout, the statistical algorithms knew nothing at all about Pong. (The bars for these other networks are not missing from Figure \ref{fig:12}c, they are simply at zero.) This result is largely unsurprising: One does not expect a lookup table for addition to do subtraction, and one does not expect a lookup table for Breakout to play Pong.

The reason for DORA’s zero-shot transfer from Breakout to Pong is straightforward. As described above, during its first game of Pong, DORA represented the game state using the relations it had learned playing Breakout. Armed with these relations, the model used analogical mapping to discover the correspondences between the two games, and based on those correspondences, made inferences about what kinds of moves were likely to succeed in the new situation. The model’s prior experience with Breakout thus allowed it to play its first game of Pong like a good rookie rather than a rank novice. 

As a final test, we trained the DNNs (but not DORA) to play Pong until they could play with competence, and then retested the DNNs and DORA for their ability to play Breakout. Of interest in this simulation was whether the various systems, upon learning to play Pong, would still know how to play Breakout. Figure \ref{fig:12}c shows the performance of the networks on the first 100 games of Breakout after learning Pong. DORA returned to Breakout with little difficulty (t(198) $=$ 1.26, p>.05; again DORA engaged in no learning during these test games). By contrast, the deep DNNs showed extremely poor performance, indicating that learning to play Pong had completely overwritten their ability to play Breakout \parencite[i.e., the networks suffered interference from Pong to Breakout; see][]{french1999catastrophic}. \footnote{Catastrophic forgetting can be avoided by interleaved training \parencite[i.e., training on to-be-learned tasks simultaneously, with “batch” updating of connection weights; e.g.,][]{kirkpatrick2017overcoming}. Sequential training of the type people routinely encounter continues to produce catastrophic forgetting in DNNs}.  

It is important to stress that the supervised DNN, one of the DQNs, and the graph net used the same visual pre-processor as DORA, so the differences in generalization performance cannot be attributed to differences in inputs (e.g., individuated objects and rate-coded dimensions). Rather, the DNNs’ generalization failure reflects the purely statistical nature of their representations. For a DNN screens from Breakout and screens from Pong are simply from different distributions, and therefore, it has no reason to generalize between them. By contrast, relation-based learning—like a variablized algorithm—naturally generalizes to novel values (arguments) bound to the variables (relational roles) composing in the algorithm (model of the task).

\subsection{Simulation 3: Cross-domain transfer from shape images to video games}

A serious limitation of Simulation 2 is that all the models embarked on learning Breakout as blank slates, with no prior knowledge of any kind. We adopted this practice to remain consistent with the tradition in the prior literature on neural networks for game play \parencite[e.g.,][]{mnih2015human}, which is to start with completely untrained networks. This convention no doubt reflects the fact that DNNs do not profit from cross-domain transfer, so there is no point in training them on any other kind of task (e.g., to teach them basic spatial relations) before training them to play video games. 

People, by contrast, learn very differently than DNNs. Rather than approaching each new task as a \textit{tabula rasa}, people bring their prior knowledge to the learning of new tasks. By the time a person plays their first video game, they have no doubt had extensive experience with such basic spatial relations as \textit{above} and \textit{left-of}, not to mention years of experience in numerous other domains. This difference between how people and DNNs learn is important as it speaks directly to the importance of cross-domain transfer: Whereas purely associative systems such as DNNs suffer from retraining on a new task, people \textit{rely} on it. 

Accordingly, Simulation 3 explored a slightly more realistic course of learning in DORA. Instead of learning the relations relevant to playing the game from the game itself, DORA first learned representations from a different domain, namely the first 300 images from the CLEVR data set \parencite[pictures consisting of multiple objects on a screen;][]{johnson2017clevr}. We started with a version of DORA with no knowledge. CLEVR images were run through the pre-processor and the results were encoded into DORA’s LTM. We then ran DORA for 2500 unsupervised learning trials (as in simulation 1). As in previous simulations, DORA successfully learned structured representations of the relations present in the stimuli—here \textit{above}, \textit{below}, \textit{same-vertical}, \textit{right-of}, \textit{left-of}, \textit{same-horizontal}, \textit{wider}, \textit{thinner}, \textit{same-width}, \textit{taller}, \textit{shorter}, \textit{same-height}, \textit{larger}, \textit{smaller}, \textit{same-size}, \textit{same-color}, and \textit{different-color}. We do not claim that this pre-training with the CLEVR images provides DORA with a realistic approximation of a person’s pre-video game experience. On the contrary, DORA’s pretraining is a pale imitation of the rich experiences people bring to their first video game experience. But the crucial question in Simulation 3 is not whether we can endow DORA with all the advantages a person’s prior experiences bring to their ability to learn video games, but merely whether DORA, like people, is capable of profiting from prior exposure to relevant spatial relations even if that exposure comes from a completely different domain. 

Following representation learning from the CLEVR images, DORA learned to play Breakout via Q-learning for 800 games. Again, the key difference from previous simulation was that DORA used the representations that it had learned from the CLEVR images to encode game screens. No additional representation learning occurred from experience with Breakout. Only associations between the previously learned representations and successful moves were updated via Q-learning. As in Simulation 2, after training with Breakout, we tested the model’s ability to generalize to playing Pong, and then return to playing Breakout. 

\begin{figure*}[!htbp]
\centering
\includegraphics[width=13.76cm,height=5.17cm]{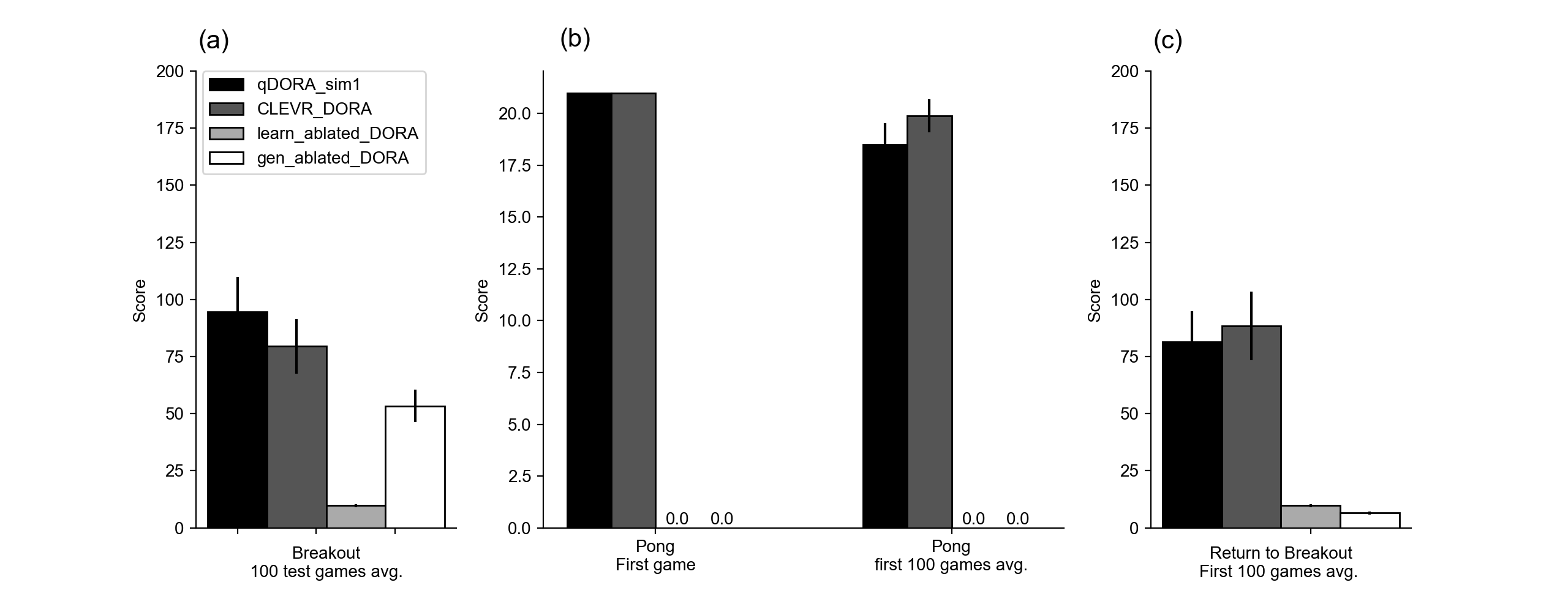}
\caption{Results of game play performance for DORA from simulation 1 and after CLVR learning (simulation 2; blue and light blue bars), and DORA after ablation (simulation 4; green and light green bars). Error bars represent 2 stderrors. (a) Performance of DORA on Breakout as an average of 100 test games. (b) Results of DORA playing Pong after training on Breakout as the score of the first game played and an average score of the first 100 games played. (c) Results of DORA when returning to play Breakout after playing or learning to play Pong, as an average score for the first 100 games played.}
\label{fig:13}
\end{figure*}

Using representations learned from CLEVR, DORA learned to play Breakout and transferred learning form Breakout to Pong and back to Breakout in a manner very similar to the results of Simulation 2 (Figure \ref{fig:13}a-c, black and dark grey bars). However, DORA learned to play Breakout in fewer games when it started with the representations learned from the CLEVR images than it did starting with a blank slate in Simulation 1 (800 vs. 1250 games, respectively; as it did not need to learn representations, only a policy for associating representational states with actions). This simulation demonstrates that DORA—like a human learner—exploits cross-domain transfer rather than suffering from it. DORA’s capacity to do so is a direct reflection of its ability to represent the domain-relevant relations explicitly, bind them to arguments, and map them onto corresponding elements between the familiar and novel games. 

\subsection{Simulation 4: The centrality of binding and structured representations in cross-domain transfer in DORA}

DORA relies on neural oscillations to dynamically bind distributed representations of objects and relational roles into relational structures. According to our account, these oscillations play a central role in learning and generalization because without them, DORA’s representations would be non-structured feature lists—akin to the representations used by DNNs and other associative learning algorithms—and its generalization ability would be correspondingly limited. To explore the role of neural oscillations—that is, structured relational representations—in DORA’s performance, we reran Simulation 2 (allowing the model to learn to play Breakout, and then attempting to generalize to Pong), but with two different ablated versions of the model. In both ablated versions we disrupted the lateral inhibition between token units (specifically, we reduced the weight of the inhibitory lateral connections between tokens from -1 to -0.1), disrupting the model’s ability to maintain systematic oscillatory behavior. In the first ablated model (A1), we ablated the inhibitory connections from the onset of the simulation. As a result, neural oscillations were disrupted both during predicate learning and thereafter. In the second ablated model (A2), we ablated the inhibitory connections after the model had learned to play Breakout: Although the model was intact when it learned to play Breakout, the neural oscillations, and thus role-argument bindings, were disrupted during generalization to Pong. 

The current simulations were otherwise identical to simulations 2. As expected, model A1 failed to learn any useful predicate representations. Disrupting the model’s neural oscillations eliminated its capacity to learn predicates, and thus greatly reduced its capacity to learn Breakout. The model resorted to learning based on the absolute features of the stimuli, and thus learned much like a less sophisticated DQN. Based on these representations, the model struggled with Breakout even after 20,000 training games and failed to generalize to Pong (Figure \ref{fig:13}, light grey bars). Model A2, which was intact during predicate learning and Breakout training, learned predicate representations and achieving good performance on Breakout within 1000 games (Figure \ref{fig:13}a, white bar). However, when the oscillations were disabled after training, the model failed to generalize to Pong (Figure \ref{fig:13}b, white bar). This result demonstrates the centrality of systematic oscillations in the model’s capacity to learn relational representations (model A1) and of structured relational representations to its ability to perform generalization (models A1 and A2). 

\subsection{Simulation 5: Transfer from games to more complex tasks}

This simulation was designed to further challenge the capacity of the representations that DORA learns. In this simulation we investigated whether the representations DORA learned playing video games and from CLEVR (i.e., simulations 2 and 3) would allow the model to generalize to the very different domain of analogical reasoning. To this end, we used the same model and representations from simulations 2 and 3 and set to it tasks representing characteristics of human-level analogical reasoning \parencite{bassok1995judging, gick1983schema, holyoak2012analogy, holyoak1995mental}.  Specifically, after learning representations from Breakout and CLEVR, we tested whether it could immediately (i.e., with no additional experience) use those representations to: (i) solve analogical cross mappings; (ii) analogically map similar, but non-identical predicates; (iii) analogically map objects with no featural overlap—including completely novel objects—that play similar roles; and (iv) map the arguments of an \textit{n}-place relation onto those of an \textit{m}-place relation even when \textit{n} and \textit{m} are unequal \parencite[i.e., called violation the \textit{n-ary restriction};][]{hummel1997distributed}. As such, the simulation had two purposes: (a) to evaluate the capacity of the representations the model learns to support human level analogical reasoning; (b) to provide a further test of the model’s capacity for cross-domain generalization: Just like humans do, the model had to learn representations in one domain, and use these representations to reason in a novel (laboratory) task. 

During a cross-mapping, an object (object1) is mapped to a featurally less similar object (object2) rather than a featurally more similar object (object3) because it (object1) plays the same role as the less similar object. For example, if cat1 chases mouse1 and mouse2 chases cat2, then the structural cross-mapping places cat1 into correspondence with mouse2 because both are bound to the \textit{chaser} role. The ability to find such mappings is a key property of human relational (i.e., as opposed to feature-based) reasoning \parencite[e.g.,][]{bassok1995judging,gick1983schema,holyoak2012analogy, richland2006children}. Cross-mappings serve as a stringent test of a computational system’s structure sensitivity as they require the system to discover mappings based on relational similarity in the face of competing featural or statistical similarity. 

We tested the representations DORA learned in Simulations 2 and 3 for their ability to support cross-mapping. DORA randomly selected two of the predicates (T1 units) it had learned during Simulations 2 and 3, such that both predicates coded for the same relation (e.g., both coded for \textit{above} or both coded for \textit{same-width}). DORA bound the relations to new objects, to form two new propositions, P1 (e.g., \textit{above} (object1, object2)) and P2 (e.g., \textit{above} (object3, object4). We manipulated the objects such that the agent of P1 (object1) was featurally identical to the patient of P2 (object 4) and the patient of P1 (object 2) was featurally identical to the agent of P2 (object 3). Based on the objects’ featural similarity, DORA would therefore map object1 to object4 and object2 to object3, but based on their roles in the relational structures, it should map object1 to object3 and object2 to object4. We repeated this procedure 100 times (each time with different randomly chosen T1 units) using representations from Simulations 2 and 3. In every case, DORA successfully mapped object1 to object3 and object2 to object4 (the structurally consistent mappings) rather than object1 to object4 and object2 to object3 (the feature-based mappings). This result demonstrates that the relations DORA learned in Simulations 2 and 3 immediately transfer to analogy tasks and support relational cross-mapping. 

We then tested whether DORA’s relational representations support mapping similar but non-identical relations (such as mapping \textit{above} to \textit{greater-than}) and mapping objects with no featural overlap based only on their bindings to similar roles. DORA randomly selected two of the relations, R1 and R2 (e.g, \textit{above} (\textit{x},\textit{y}) or \textit{wider} (\textit{x},\textit{y})), that it had learned during Simulations 2 and 3 such that each role of R1 shared roughly half of its features with the corresponding role of R2 (e.g., the role \textit{more-y} has half of its features in common with the role \textit{more-length}). The objects serving as arguments of the relations had no featural overlap at all. We repeated this process 100 times and each time, DORA mapped the agent role of R1 to the agent role of R2 and the patient role of R1 to the patient role of R2. Even though the objects had no features in common, and even though the relations to which they were bound were not identical, DORA found the structurally correct object and role mappings. 

Next, we tested whether the representations DORA learned can violate the \textit{n}-ary restriction, mapping the arguments of an \textit{n}-place predicate onto those of an \textit{m}-place predicate when $n \neq m$. In each of these simulations, DORA randomly selected a relation, R1, that it had learned in Simulations 2 and 3, and we created a single-place predicate (r2) that shared 50$\%$ of its features with the agent role of R1 and none of its features with the patient role. DORA then bound two objects to the roles of R1 to form the proposition \textit{R1} (object1, object2), and bound a third object to r2, to form the proposition \textit{r2} (object3). Object3 shared half its features with object1 and the other half with object2 (i.e., it was equally similar to both object1 and object2). DORA attempted to map \textit{r2} (object3) onto \textit{R1} (object1, object2). If the model can violate the \textit{n}-ary restriction, then it should consistently map object3 to object1 based on the similarity of \textit{r2} to the first (agent) role of \textit{R1} (recall that \textit{R1} is represented as a linked set of roles). This process was repeated 100 times using a different randomly chosen R1 each time. Each time DORA successfully mapped object3 to object1, along with corresponding relational roles (i.e., DORA maps the predicate representing one of the roles of \textit{R1} and the predicate representing the single-place predicate \textit{r2}). We then ran 100 simulations in which r2 shared half its features with the second (patient) role of R1 rather than the first (agent) role. In these 100 additional simulations, DORA successfully mapped the patient role of R1 to r2 (along with their arguments). 

Finally, we tested whether the representations that DORA learns support generalization to completely novel (i.e., never before experienced) stimuli. The ability to make generalizations about completely novel items is the hallmark of the capacity for universal generalization (see, e.g., Marcus, 2001). In this simulation, we created six objects with completely novel features (features units grafted onto DORA). DORA randomly selected two instances of the same relation that it had learned in Simulations 2 and 3 (e.g., two instances of \textit{bigger}). We bound one instance of the relation to three of the objects—object1, object2, and object3—to create three propositions R1-R3 that instantiated a transitive relation such that object2 served as both the patient of one proposition and the agent of the other. For example, if the relation DORA had selected was \textit{bigger}, then it bound that relation to object1 and object2 to make the proposition R1, \textit{bigger }(object1, object2), bound the same relation to object2 and object3 to make the proposition R2, \textit{bigger }(object2, object3), and finally bound the same relation to object1 and object3 to make the proposition R3, \textit{bigger }(object1, object3) (recall that within a single analog, or story, token units are shared between proposition, so the same T1 unit instantiating, say, \textit{more-size} was bound to both object1 and object2). From the remaining three objects— object4, object5, object6—we used the second instance of the relation to make the propositions R4 and R5 such that object5 served as both the patient in one proposition and the agent in the other. For example, we bound object4 and object5 to \textit{bigger} to make R4, \textit{bigger} (object4, object5), and bound object5 and object6 to \textit{bigger} to make R5, \textit{bigger }(object5, object6). Importantly, none of the objects had any features (including metric features) that DORA has previously experienced. As such the simulation was akin to telling DORA that a transitive relation held between three novel objects, and that some of the same relations held between a second three objects. We placed R1 and R2 and R3 in the driver, and R4 and R5 in the recipient. We wanted to see whether (a) DORA could integrate relations such that it would map R1 to R4 and R2 to R5 (i.e., it would map the instances where the same object was a patient and an agent); and (b) whether it would use these mappings (if discovered) to generalize the transitive relation in R3 to the objects in the recipient (i.e., would it complete the transitive set of relations about objects4-6). If DORA found a mapping, it then attempted to perform analogical inference. We ran the simulation 100 times. In each simulation, DORA mapped the representation of R1 to R4, and R2 to R5. Moreover, DORA generalized R3 to the recipient to create a transitive proposition R6—e.g., to continue the above example, DORA inferred that \textit{bigger} (object4, object6). That is, armed with the knowledge that some objects it had never before experienced played particular relational roles, DORA generalized that a proximal object (that it had also never experienced before) might play a complementary role. The inability to reason about completely novel features (i.e., features not part of the training space) is a well-known limitation of traditional neural networks \parencite[e.g.,][]{bowers2017parallel}. However, this limitation does not apply to DORA. Not only can DORA represent that novel objects can play certain roles (because it can dynamically bind roles to fillers), but it can also use its representations to make inferences about other completely novel objects. 

In total, simulation 5 demonstrates that the relational representations DORA learned during Simulations 2 and 3 immediately support performance of an unrelated task (analogical reasoning) even with completely novel objects. After learning representations in one domain (game play and images of shapes), DORA, with no additional experience (zero-shot), used these representations to solve a set of analogical reasoning tasks representing several hallmarks of human analogical thinking, and then used these representations to generalize to completely novel objects. The results provide further evidence that DORA’s representations support cross-domain transfer and highlight the generality of the DORA framework. 

\subsection{Simulation 6: Development of representations of relative magnitude}

We have previously shown that DORA’s format learning algorithm provides a good account of several developmental phenomena in representational development \parencite[e.g.,][]{doumas2008theory}. The purpose of simulations 6 and 7 was to examine whether the representations that DORA learned, and the trajectory of the representation learning mirror human development when DORA is learning both relational content \textit{and} relational format. Additionally, these simulations provided another opportunity to evaluate generalization of representations across domains: Learning representations in one domain and deploying those representations to reason about a new domain (as human often do when they engage in laboratory experiments). 

Children develop the ability to reason about similarity and relative magnitude on a variety of dimensions \parencite[e.g.,][]{smith1984young}. The development of children’s capacity to reason about basic magnitudes is well demonstrated in a classic study by  \textcite{nelson1974comprehension}. In their experiment, children aged three to six years-old were given a simple identification task. An experimenter presented the child with two pictures of similar objects that differed on some dimensions. The experimenter then asked the child to identify the object with a greater or lesser value on some dimension. For example, the child might be shown pictures of two fences that differed in their height, their size, and their color, and then asked which of the two fences was taller or shorter. The developmental trajectory was clear: Children between 3-years-10-months and 4-years-4 months (mean age $\sim$48 months) made errors on 34$\%$ of trials, children between 4-years-7-months and 5-years-5-months (mean age $\sim$60 months) made errors on 18$\%$ of trials, and children aged 5-years-6-months and 6-years-6-months (mean age $\sim$73 months) made errors on only 5$\%$ of trials. In short, as children got older, they developed a mastery of simple magnitude comparisons on a range of dimensions. 

If DORA is a good model of human representational development, then it should be the case that DORA’s representations follow a similar developmental trajectory. To test this claim, we used the representations that DORA had learned during Simulation 2. If DORA develops like a human child, then early in the learning process, DORA’s performance on the Nelson and Benedict task should mirror 3-4 year-old children, later in the learning process DORA’s perform should mirror 4-5 year-old children, and later in the learning process DORA’s performance should mirror 5-6 year old children. 

To simulate children of different ages we stopped DORA at different points during learning and used the representations that it had learned to that point (i.e., the state of DORA’s LTM) to perform the magnitude reasoning task. To simulate each trial, we created two objects instantiated as T1 units attached to features. These features included 100 random features selected from the pool of 10,000, along with features encoding height, width, and size (dimensions used in Nelson $\&$ Benedict) in pixel format (as in the simulations above; e.g., for an object 109 pixels wide, one feature unit describing "width0" and 109 features encoded "109 pixels wide"). A dimension was selected at random as the question dimension for that trial. DORA sampled at random a representation from its LTM that was strongly connected to that dimension (with a weight of .95 or higher). If the sampled item was a relation or a single-place predicate, DORA applied it to the objects, and placed that representation in the driver. For example, if the key dimension was size, the two objects (obj1 and obj2) were then run through the relational invariance circuit on the dimension of size, marking one (assume obj1) as relatively larger and the other (assume obj2) as relatively smaller. If DORA had sampled a representation of the relation \textit{larger }(\textit{x},\textit{y}), then the \textit{more-size} T1 unit was bound to the obj1 T1 unit and the \textit{less-size} T1 unit was bound to the obj2 T1 unit. To simulate a dimensional question, DORA randomly sampled a representation of the question dimension from LTM and placed that in the recipient. For example, if the question was, "which is bigger", DORA sampled a representation of a T1 unit encoding \textit{more-size} from LTM. DORA then attempted to map the driver and recipient representation. If DORA mapped a representation in the driver to a representation in the recipient, the mapped driver item was taken as DORA’s response on the task. If DORA failed to find a mapping, then an item was chosen from the driver at random and taken as DORA’s response for that trial (implying that DORA was guessing on that trial). The probability of guessing the correct item by chance was 0.5. 

\begin{figure}[!htbp]
\centering
\includegraphics[width=8cm]{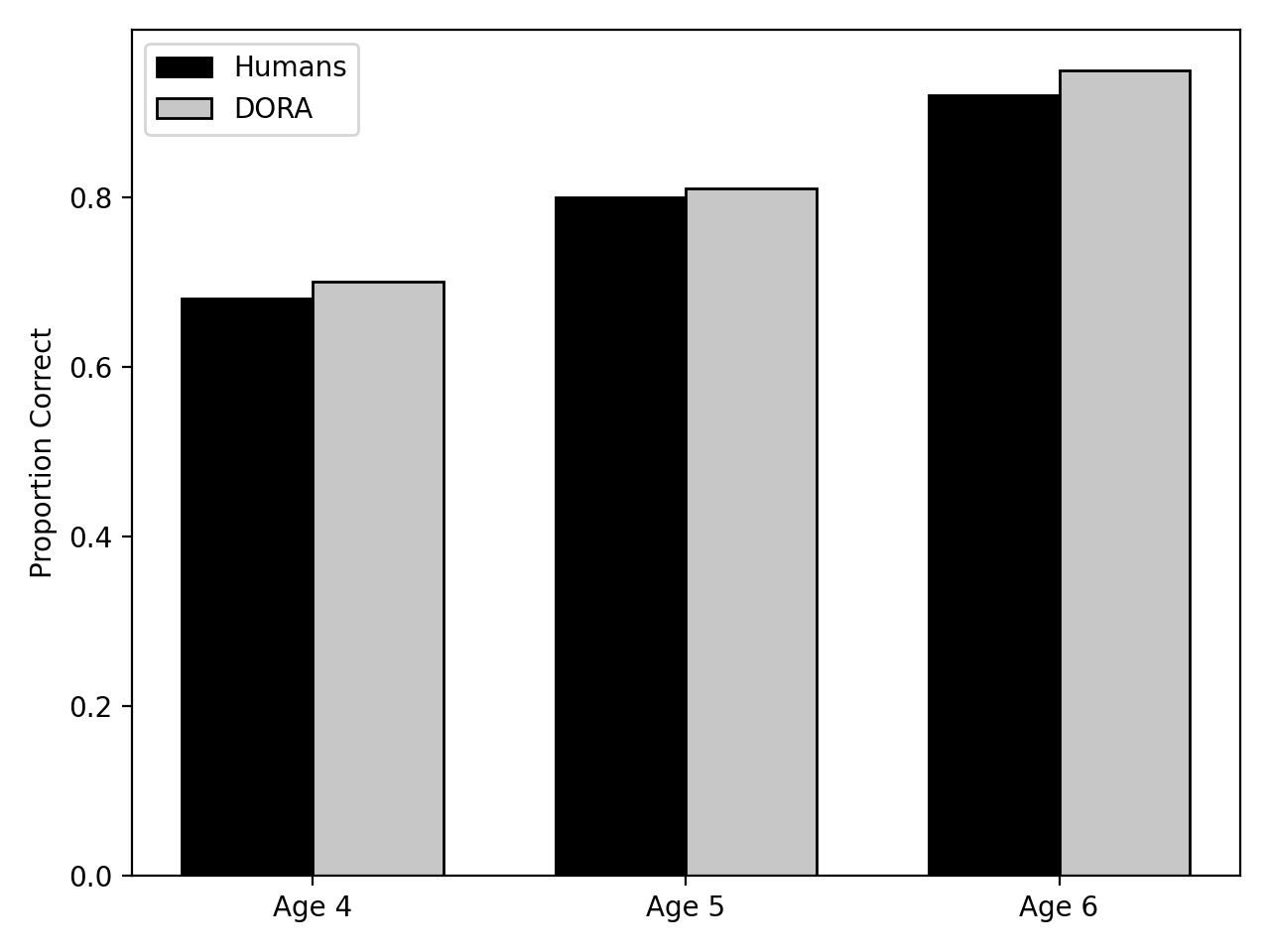}
\caption{Results of simulation of \textcite{nelson1974comprehension}.}
\label{fig:14}
\end{figure}

To simulate 4 year-olds, we used the representations in DORA’s LTM after 1000 total training trials, to simulate 5 year-olds we used the representations in DORA’s LTM learned after 1000 additional training trials (2000 total trials), and to simulate 6 year-olds we used the representations in DORA’s LTM learned after 1000 additional training trials (3000 total trials). We ran 50 simulations each with 20 trials at each age level (each simulation corresponding to a single child). The results of the simulation and the original results of Nelson and Benedict are presented in Figure \ref{fig:14}. 

The qualitative fit between DORA’s performance and the performance of the children in Nelson and Benedict’s study is close. Just like the children in the original study, DORA is better than chance, but still quite error prone early during learning, but gradually comes to learn representations that support very successful classification of dimensional magnitudes. These simulation results provide evidence that the trajectory of the development of DORA’s representations of relative dimensional magnitude mirrors that of humans. 

\subsection{Simulation 7: The relational shift}

One of the key findings from work on the development of analogical reasoning in children, is that children go through a relational shift \parencite[e.g.,][]{gentner1995two}. The relational shift describes a qualitative change in children’s reasoning wherein they progress from making analogies based on the literal features of things, to making analogies based on the relations that objects are involved in \parencite[e.g.,][]{richland2006children}. With development, children learn progressively more powerful representations of similarity and relative magnitude relations that support more proficient relational generalization \parencite{smith1984young}. In addition, children develop the capacity to integrate multiple relations in the service of reasoning \parencite[e.g.,][]{halford1980category}, and their relational representations grow more robust with learning, and allow them to overcome ever more excessive featural distraction \parencite[][]{halford1980category,rattermann1998effect}. 

\begin{figure}[!htbp]
\centering
\includegraphics[width=6.82cm,height=7.31cm]{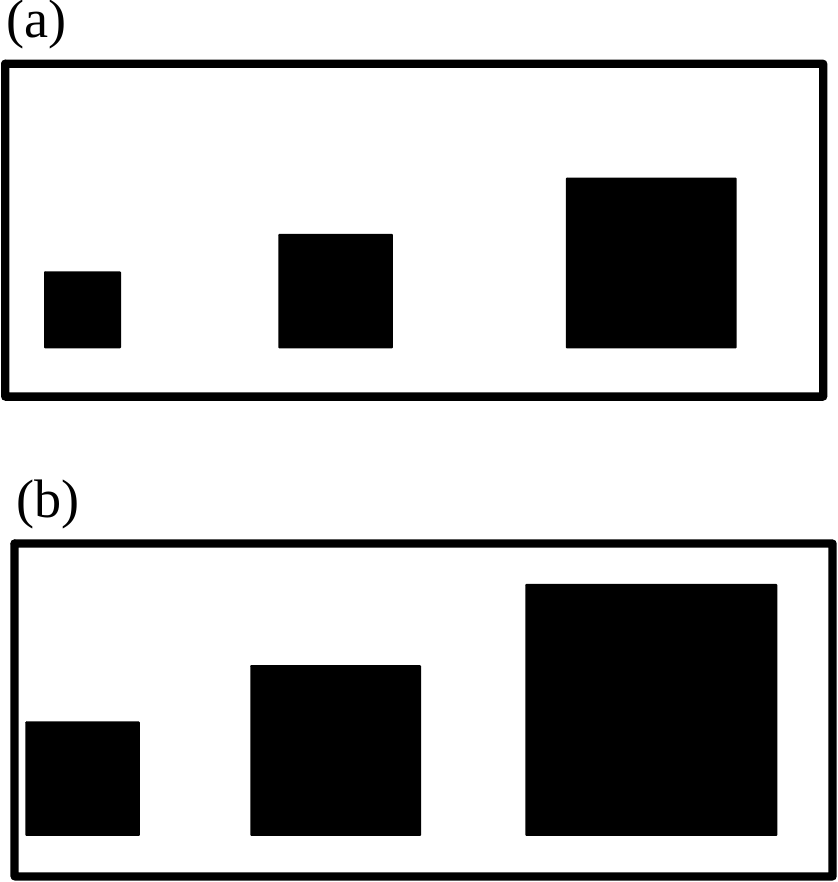}
\caption{A recreated example of the stimuli used in \textcite{rattermann1998effect}.}
\label{fig:15}
\end{figure}

One of the classic examples of the relational shift and the associated phenomena is given in  \textcite{rattermann1998effect}. In their experiment, \citeauthor{rattermann1998effect} had 3-, 4-, and 5-year-old children participate in a relational matching task. Children were presented with two arrays, one for the child and one for the experimenter. Each array consisted of three items that varied on some relative dimension. For example, the three items in each array might increase in size from left to right or decrease in width from left to right. The dimensional relation in both presented arrays was the same (e.g., if the items in one array increased in size from left to right, the items in the other array also increased in size from left to right). The items in each array were either sparse (simple shapes of the same color) or rich (different shapes of different colors). The child watched the experimenter hide a sticker under one of the items in the experimenter’s array. The child was then tasked to look for a sticker under the item from the child’s array that matched the item selected by the experimenter. The correct item was always the relational match—e.g., if the experimenter hid a sticker under the largest item, the sticker was under the largest item in the child’s array. Critically, at least one item from the child’s array matched one of the items in the experimenter’s array exactly except for its relation to the other items in its array. To illustrate, the experimenter might have an array with three squares increasing in size from left to right (Figure \ref{fig:15}a). The child might have an array of three squares also increasing in size from left to right, but with the smallest item in the child’s array identical in all featural properties to the middle item in the experimenter’s array (Figure \ref{fig:15}b). Thus, each trial created a cross-mapping situation, where the relational choice (same relative size in the triad) was at odds with the featural choice (exact object match). The child was rewarded with the sticker if she chose correctly.

\citeauthor{rattermann1998effect} found a relational shift. Children between 3 and 4-years-old were very drawn by featural matches and had trouble systematically making relational matches (making relational matches 32$\%$ of the time in the rich condition and 54$\%$ of the time in the sparse condition). Children between 4 and 5-years-old were quite good at making relational matches with sparse objects—making relational matches 62$\%$ of the time—but still had trouble with rich objects when featural matches were more salient—making relational matches 38$\%$ of the time. Children between 5 and 6-years-old were quite good at making relational matches in both the rich and the sparse conditions, with the rich condition providing more trouble than the sparse condition—making relational matches 68$\%$ for rich and 95$\%$ of the time for sparse stimuli. 

We simulated the results of \textcite{rattermann1998effect} in as in the simulation above, using the representations learned during simulation 2. Again, to simulate children of different ages we stopped DORA at different points during learning and used the representations that it had learned to that point. To simulate 3-year-olds we used the representations in DORA’s LTM after 850 training trials, to simulate 4.5-year-olds we used the representations in DORA’s LTM after 1500 training trials, and to simulate 5.5-year-olds we used the representations in DORA’s LTM after 2500 training trials. To simulate each trial, we created two arrays of three objects, each object instantiated as a T1 unit connected to features. For the sparse trials, each object was connected to feature units such that some features encoded absolute size, height, width, x-position, y-position, color, 10 features described shape, and four features were chosen at random from a pool of 1000. The identical objects from both arrays matched on all features. For the rich trials, each object was attached to additional features: some features encoding absolute size, height, width, x-position, y-position, color, 10 features describing shape, four features describing object kind (e.g, "shoe", "train", "bucket"), and 40 features chosen at random from a pool of 1000. The identical objects from both arrays matched on all features. 

We ordered the objects in both arrays according to some relation (e.g., increasing size, decreasing width). DORA sampled four representations from its LTM that were strongly connected to that dimension (with a weight of .95 or higher) and applied two of the sampled representations to each of the two arrays. If the sampled representation was a relation or a single-place predicate, it applied to the objects. For example, if the key dimension was size, and DORA sampled a representation of the relation \textit{larger }(\textit{x},\textit{y}), it applied that representation to the objects, binding the larger object to the \textit{more-size} role and the smaller object to the \textit{smaller} role (as described in simulation 6). If the sampled representation was a single-place predicate like \textit{more-size }(\textit{x}), then it was bound to the larger object. As each array consisted of two instances of the key relation (e.g., \textit{larger} (object1, object2), and \textit{larger} (object2, object3)), DORA applied one of the two sampled items to one of the relations in the array, chosen at random, and the other sampled item to the other relation in the array. For simplicity, the model only considered relations between adjacent objects. 

The representation of the child’s array entered the driver, and the experimenter’s array the recipient. An item from the recipient was chosen at random as the "sticker" item (i.e., the item under which the sticker was hidden). The capacity to ignore features is a function of the salience of those features, and so richer objects with more features are harder to ignore \parencite[see, e.g.,][]{goldstone2012similarity}. To simulate the effect of the rich vs. the sparse stimuli, on each rich trial, DORA made a simple similarity comparison before relational processing started. It randomly selected one of the items in the driver and computed the similarity between that item and the "sticker" item in the recipient using the equation: 

\begin{equation}
sim_{ij} =\frac{1}{1+\sum_{i}^{}\left(1-s_{i}\right)} \\ 
\label{eq:13}
\end{equation}
% Equation 13

\noindent where, $sim_{ij}$ is the 0 to 1 normalized similarity of PO unit $i$ and PO unit $j$, and $s_{i}$ is the activation of feature unit $i$. If the computed similarity was above .8, then DORA learned a mapping connection between the two items. Finally, DORA attempted to map the items in the driver to the items in the recipient. If any driver representation was mapped to the "sticker" item in the recipient, the mapped item was taken as DORA’s response on the task. If DORA failed to find a mapping, then it selected an item from the recipient at random as a response for that trial (implying that DORA was guessing on that trial). The probability of guessing the correct item by chance was 0.33. 

\begin{figure*}[!htbp]
\centering
\includegraphics[width=13.76cm,height=5.85cm]{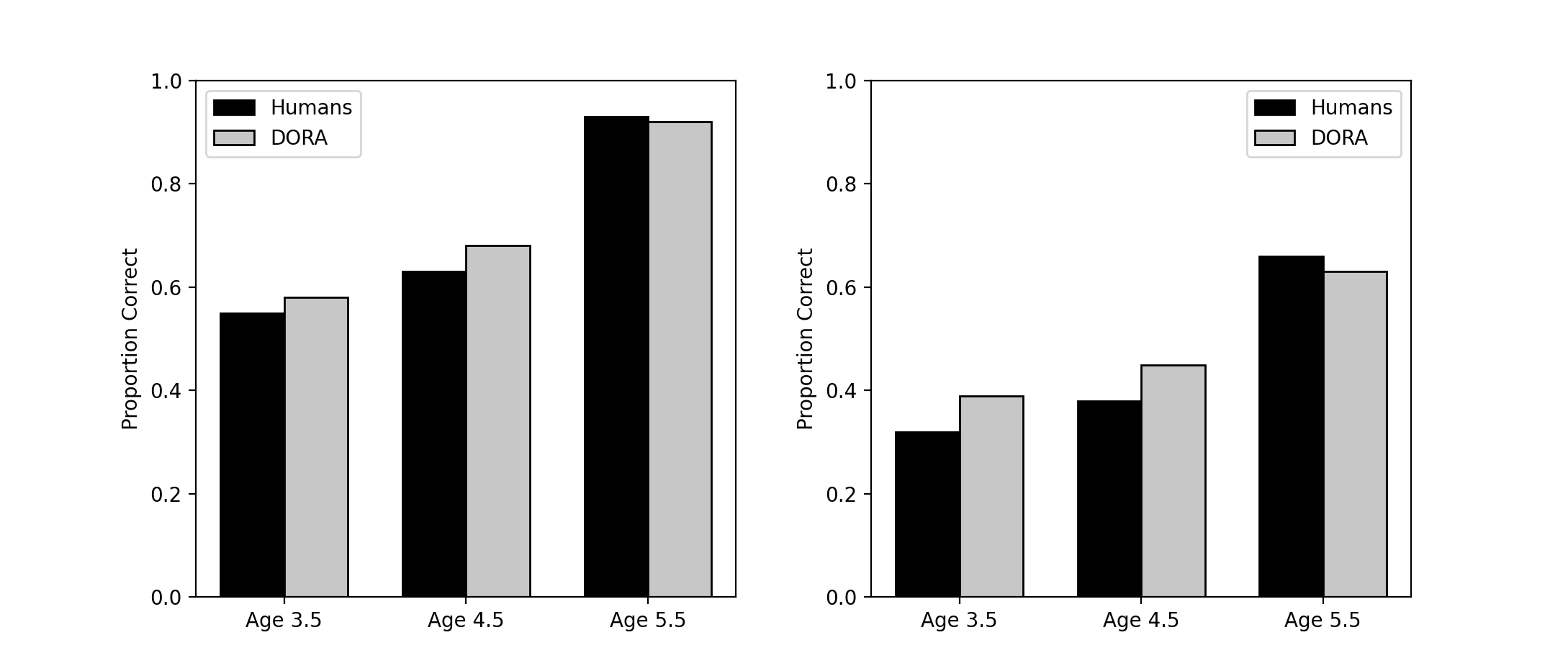}
\caption{Results of simulation of \textcite{rattermann1998effect}. (a) Performance of children and DORA on sparce trials. (b) Performance of children and DORA on rich trials.}
\label{fig:16}
\end{figure*}

We ran 50 simulations each consisting of 20 trials at each age level. The results of the simulation as well as those from the original Rattermann and Gentner experiment with both sparse and rich trials are presented in Figure \ref{fig:16}a and \ref{fig:16}b respectively. 

As Figure \ref{fig:16} shows, there is a close qualitative fit between DORA’s performance and the performance of the children in  \textcite{rattermann1998effect}. Initially, DORA, just as the 3-year-old children in the original study, had some trouble correctly mapping the items in the driver and the recipient, and struggled to solve the cross-mapping. As DORA learned more refined representations (after more training), like the 4-year-old children in the original study, DORA began to solve the sparse problems more successfully, while still struggling with the rich problems. Finally, like the 5-year-old children in the original study, after even more learning, DORA was quite successful at both rich and sparse trials, reaching ceiling level performance on the sparse problems. These simulation results indicate that, like humans, the trajectory of the development of DORA’s representations of relative dimensional magnitude undergoes a relational shift with learning. Additionally, the representations that DORA learns during its development support the same kind of performance on relational matching tasks that is evidenced by human children during their development.

\section{General Discussion}

\subsection{Summary and overview}

We have presented a theory of human cross-domain generalization instantiated in the DORA computational framework. Our proposal is that people represent knowledge domains as models consisting of structured representations of the relations among the elements of those domains. These representations specify the invariant content of the relations and their arguments, and representations of relational roles are dynamically bound to their fillers while maintaining role-filler independence. DORA learns both the content and structure of these relations from non-relational inputs, such as visual displays, without supervision. DORA uses comparison to bootstrap learning both the content and structure of relational representations. By integrating these representations with a capacity for reinforcement learning, DORA learns which relations to use in what contexts in the service of problem solving (e.g., game play). The resulting representations can be applied to new domains, including completely novel ones, by a process of analogical inference. That is, the model generalizes across domains as a natural consequence of its ability to represent relations in a manner that is invariant with both the arguments of the relations and the specific circumstances in which those relations arise. 

A series of simulations demonstrated that this approach to learning and knowledge representation greatly facilitates cross-domain generalization. Simulation 1 showed that the model is capable of learning structured representations of relations from unstructured, non-relational visual inputs. Simulation 2 showed that, as a result of learning to play one video game, DORA learns representations that support immediate (zero-shot) transfer to a different game (Pong), and the capacity to move between games successfully. By contrast, four different associative networks (two DQNs, a DNN, and a GNN) both (a) failed to transfer knowledge from one game to another and (b) lost their ability to play the first game after training on the second. Simulation 3 demonstrated that the representations DORA learns in one domain (images of 3D shapes) support learning to play one video game play and immediately generalize to a new game. Simulation 4 demonstrated the essential role of structured representations in the model’s learning and generalization. Simulation 5 showed that the representations DORA learns from domains like video games and pictures also support successful zero-shot transfer to unrelated reasoning tasks (cross-mapping, mapping non-identical predicates, mapping novel objects, and violating the \textit{n}-ary restriction), and, importantly, support generalization to completely novel (i.e., never previously experienced) stimuli. Finally, simulations 6 and 7 showed that DORA follows the same developmental trajectory as children as it learns representations. That is, DORA accounts for results from the literature on children’s reasoning as it learns to play video games. 

\subsection{LISAese, relational databases, and generalization}

We have argued that the reason that people can learn relations and apply them to new domains is because we learn representations of those relations that specify relational invariants in a form that permits the binding of relational roles to arguments without changing the representation of either \parencite[see also][]{doumas2008theory, halford1998processing, halford1998induction, hummel1992dynamic, hummel1997distributed, hummel2003symbolic, phillips1995processing, phillips2018going, phillips2021reconstruction}. It turns out that our proposal has an analog in computer science in the form of relational databases.\footnote{We are indebted to Reviewer 3 for their help in making the discussion of correspondence between LISAese and relational schemas much stronger.} 

In mathematical logic, a relation is defined as subset of the Cartesian product of two or more potentially infinite sets. For example, the relation \textit{larger-than}(), defined over the integers, is a matrix (a Cartesian product), with integers in the rows and columns, and 1s and 0s in the cells, such that a 1 appears in every cell whose row is larger than its column.  More generally, a binary relation between sets $A$ and $B$ is a subset of the Cartesian product $A \times B $ for each pair (a, b) over which the relation holds. A relation is thus represented by a characteristic function, $\chi_R(a, b)$, which maps to $1$ if the relation is true for (a, b) and $0$ otherwise. 

The characteristic function captures the same information as the relational invariants we described in the Introduction. In other words, the characteristic function specifies the \textit{content} of the relation, so learning the invariant that defines a relation is a matter of executing the characteristic function. The circuit DORA uses to discover relational invariants is nothing more than an implementation of the characteristic function of, for example, \textit{larger-than} () over a rate-coded neural representation of magnitude. 

However, we have argued that simply expressing an invariant (the output of the characteristic function) is not sufficient to support cross-domain generalization. It is also necessary to somehow represent the dynamic binding of arguments to roles of the relation in a way that preserves the identity of both the relational roles and their arguments. The representational format we have employed for this purpose, LISAese, is isomorphic to a \textit{relational schema} (or \textit{relational database}) developed in computer science. 

A relational schema \parencite[see, e.g.,][]{phillips2018going} corresponds to the headings of  a table that describe the roles of the so-related elements. The table consists of a set of rows, representing instances of the relation, and columns, corresponding to the roles of the relation. For example, the relational schema for the relation \textit{larger-than} (\textit{x}, \textit{y}) includes two columns, one specifying the larger item and the other specifying the smaller, with each row of the table an instance of the \textit{larger-than} relation. This representational format has the property that relations (tables) are represented explicitly, as are their roles (columns), and arguments (cells), while simultaneously expressing the bindings of arguments to roles without altering the meaning of either. 

Halford, Phillips, and colleagues \parencite[][]{halford1998processing, halford1998induction, phillips1995processing, phillips2018going, phillips2021reconstruction} have argued that relational schemas are a good model of human mental representations. Specifically, (a) they identify a relation symbolically, (b) the roles (or argument slots) of the relation are represented independently of the fillers of those roles, (c) binding of roles to fillers is explicit, (d) the format supports representing higher-order relations (i.e., relations between relations), and (e) the resulting representations have the property of systematicity, meaning that they permit simultaneous expression of the meaning of (i) the relation, (ii) its roles, and (iii) their composition into a larger expression \parencite[see][]{halford1998processing, halford1998induction, phillips1995processing}. \textcite{phillips2018going, phillips2021reconstruction} has observed that LISAese is a representational format akin to a relational schema. And indeed, the representations that DORA learns (i.e., LISAese) satisfy all the properties of a relational database. In DORA relational representations are specified as sets of linked single-place predicates (columns) composing a header, and values bound to those predicates instantiating (rows) of the specific relation. 

In a set of experiments, \textcite{halford1980category} demonstrated that people’s inferences in a complex learning task are better captured by representations based on relational schemas than simple associations. In their Experiment 2, participants learned to associate a shape and trigram pair with another trigram. The stimuli were composed from three shapes (say, square, circle, triangle) and three trigrams (say, BEJ, FAH, PUV). The shape acted like an operator, and the mapping from shape-trigram pair to the output (i.e., the other trigram) was given by a simple rotation rule. For example, one shape when linked with a trigram mapped to the identical trigram. The second shape when linked to a trigram, mapped to the next trigram from the list (trigram-1 mapped to trigram-2, trigram-2 mapped to trigram-3, and trigram-3 mapped to trigram-1). The third shape when linked to a trigram mapped to the trigram two jumps away (trigram-1 mapped to trigram-3, trigram-2 mapped to trigram-1, and trigram-3 mapped to trigram-2). As expected, participants learned to perform the task after several exposures. The experimenters reasoned that if participants had learned the mappings by association, then when given new trigrams and new shapes that followed the same rotation rule, the participants would need roughly the same number of exposures to learn them. However, if participants had learned the relation (i.e., rotation) between shape-trigram pair and trigram output, then they should be able to apply the relation two new shape and trigram sets within a few exposures. Participants did apply the rule to new shapes and trigram within a few trials, indicating that they had learned the relational table. 

More recently, \textcite{phillips2018going, phillips2021reconstruction}, using tools from Category Theory, showed that performance on tasks like the relational schema inference task (above) requires structured relational representations (like a relational database) and cannot be accounted for by association alone. Our simulation results resonate with this claim, and suggest the capacity extends to cross-domain transfer. As demonstrated in simulations 2-4, after learning structured relational representations, DORA can learn to play a video game and then immediately generalize to a relational similar (but featurally different) game. However, if these representations are removed, the model fails utterly at any kind of generalization. 

DORA provides an account of human generalization because it can learn explicit representations of relational concepts—both their content and their format—and then leverage those representations to solve problems. In addition, the model provides an account of how such knowledge structures can be implemented in a distributed neural system, and how they can be learned from non-relational inputs.

\subsection{Other classes of relations}

In the work reported here, we have focused on transitive relations, those that can be defined by differences on a single dimension, such as \textit{larger-than} () and \textit{left-of} (). It is natural to ask whether these same principles apply to other, nontransitive relations such as \textit{chases} (), \textit{mother-of} (), and \textit{loves} (). The short answer is yes. Starting with whatever regularities it is given or can calculate from the environment, DORA’s learning algorithm will isolate those invariants, learn structured representations (i.e., functional predicates) that take arguments, and, where appropriate, compose them into relational structures. In short, given a set of invariants (or a means to calculate them), DORA’s learning mechanisms will produce explicit predicate and relational representations of those invariants. DORA will learn structured representations of concepts based on their invariant properties, whether the invariants the system detects are instances of stimulus magnitude or romantic love \parencite[e.g.,][]{doumas2008theory}. The hard part is finding the invariants. And for this problem, the human visual system may give us a leg up. 

Nontransitive relations like \textit{chase}, \textit{support}, or \textit{love} fall on a gradient in terms of how spatial they are. A relation like \textit{chase} is comparatively easy to reduce to spatial properties: There are two objects, \textit{a} and \textit{b}, such that the vector characterizing the movement of \textit{a} is in the direction of the vector characterizing the location of\textit{ b}, and this configuration is maintained through time. The representations necessary to induce this relation are delivered by the visual system. \textcite{michotte1963perception} showed that participants would overwhelming interpret chasing occurring in a situation where two dots moved across a screen (or, in Michotte’s original version, two lights moved on a grid of lights) such that one stayed in front of the other as they moved. Similarly, relations such as \textit{support} (one object \textit{above} and in \textit{contact} with another object) or \textit{lift} (one object \textit{supports} and \textit{raises} another object), are definable in spatial terms. Even a relation like \textit{loves} might reduce, at least in part, to spatial relations, though. In a study by \textcite[][]{richardson2003spatial}, participants asked to use configurations of objects to represent a relation produced overwhelmingly similar spatial arrangements for relations like \textit{love}, \textit{admire}, and \textit{hate}. 

We do not claim that all relations are spatial in origin, or that there are no invariants (e.g., characterizing social relations such as \textit{love}, \textit{hate}, \textit{friend}, \textit{adversary}, etc.) that have non-spatial origins. On the contrary, we are completely agnostic about the number and nature of the psychological dimensions over which relational invariants might be computed. What we do claim is that any psychologically privileged dimension, whether it be spatial, auditory, social, or what have you, is subject to the kind of invariant isolation and structure-inducing processes embodied in DORA: If there is an invariant, wherever it originates, intersection-discovery can find it and DORA can predicate it and use it for inference and cross-domain generalization. 

\textcite{doumas2005neural} proposed a \textit{compression} mechanism, complementary to DORA’s refinement algorithm, which is a form of chunking. During compression, multiple roles attached to the same object fire together, and a unit learns to respond to that new conjunction as a unitary predicate. Compression allows DORA to combine multiple representations of the same object. For example, if DORA encounters situations in which one element is both \textit{larger} and \textit{occludes} some second object, DORA can compress the roles \textit{larger} and \textit{occluder} and the roles \textit{smaller} and \textit{occluded} to form a representation like \textit{cover} (\textit{a}, \textit{b}). Such a procedure might serve as a basis for combining primitive transitive relations into more complex relations. 

A second question is whether the role-filler representational system DORA uses is sufficient to represent all the relations people learn. Again, the short answer is, at least in principle, yes (as pointed out originally by Leibniz). Formally, any multi-place predicate is representable as a linked set of single-place predicates \parencite{mints2001arithmetic}. Therefore, a role-filler system can, at least in principle, be used to represent higher-arity predicates (or relations; recall the distinction, noted above, between a relation qua a relational schema and a function). Models based on role-filler representations accounte for a large number of phenomena in analogy making, relational learning, cognitive development, perceotion, and learning \parencite[e.g.,][]{doumas2010computational, doumas2008theory, hummel1992dynamic, hummel1997distributed, hummel2001complementary, morrison2004neurocomputational, lim2013modeling, livins2015varying, martin2017mechanism, morrison2011computational, sandhofer2008order,son2010words}. Moreover, access to and use of role-based semantic information is quite automatic in human cognition, including during memory retrieval \parencite[e.g.,][]{gentner1995two, ross1989distinguishing}, and analogical mapping and inference
\parencite{bassok1995judging, kubose2002role, krawczyk2004structural,holyoak2008no, ross1989distinguishing}. Indeed, the meanings of relational roles influence relational thinking even when they are irrelevant or misleading \parencite[e.g.,][]{bassok1995judging, ross1989distinguishing}. Role information appears to be an integral part of the mental representation of relations, and role-filler representations provide a direct account for why. Moreover, role-filler systems appear uniquely capable of accounting for peoples’ abilities to violate the $n$-ary restriction, i.e., mapping $n$-place predicates to $m$-place predicates, where $n \neq m$ \parencite[e.g., mapping “x murdered y” onto “a caused b to die”;][]{hummel2003symbolic}\footnote{One might wonder how the role-based learning approach works for symmetrical relations like \textit{equals} or \textit{antonym}. In short, these kinds of relations are not a problem for DORA (for example, as demonstrated above the model has no problem learning relations like \textit{same-as}). If the relation is interpreted as referential then the relation is not symmetrical, and the roles are distinct. For example, in \textit{antonym}(\textit{x}, \textit{y}), \textit{y} is the referent term, playing the \textit{referent-of-something} role, and \textit{x} is the antonym of that term, playing the \textit{opposite-of-something} role. Alternately, if the relation is symmetrical and both arguments play the same role, then in a LISAese representation of that relation, there will only be a single token for the role (recall T1 tokens are not repeated within a propostion). For example, if \textit{antonym}(\textit{x}, \textit{y}) is symmetrical, and both arguments play the same role---like \textit{opposite-of}---then a single T1 token unit will represent that role in the proposition and both \textit{x} and \textit{y} will be bound to that role by distinct T2 role-binding units in LTM and by asynchrony of firing in WM. That is, the relation does not need two distinct roles, but rather is representable (and learnable) as a single role involved in two (or more) role-bindings, with those role-bindings linked to form a higher-arity proposition.} 

\textcite{livins2016shaping} showed that we can affect the direction of a relation by manipulating which item one looks at first, for both obvious similarity and magnitude relations, and the other kinds of relations. \citeauthor{livins2016shaping} showed participants images depicting a relation that could be interpreted in different forms (e.g., \textit{chase}/\textit{pursued-by}, \textit{lift}/\textit{hang}). Before the image appeared on screen, a dot appeared on the screen drawing the participants attention to a location that one of the objects involved in the relation would appear. For example, the image might show a monkey hanging from a man’s arm, and the participant might be cued to the location where the monkey would appear. The relation that the participant used to describe the image was strongly influenced by the object that they attended to first. That is, if the participant saw the image of the monkey hanging from the man’s arm, and she was cued to the monkey, they would describe the scene using a \textit{hanging} relation. However, if the participant was cued to the man, she would describe the scene using a \textit{lifting} relation. This result follows directly from a system based on role-filler representations wherein complementary relations are represented by a similar set of roles, but the predicate, or role, that fires first, defines the subject of the relation.

\subsection{Limitations and future directions}

People routinely learn structured representations from experience, an ability we argue is fundamental to our understanding of the world and our ability to use the knowledge we have gained in one context to inform our understanding of another. We offer an account of this process that is based on minimal assumptions, assumptions that, with the exception of the capacity for dynamic role-filler binding, are standard in neural networks such as DNNs. Our account is, of course, limited in several ways. In the following, we outline some of the limitations of our model and suggest ways to address these limitations. 

First, the constraints on learning in DORA are underdetermined. DORA learns when it can and stores all the results of its learning. We have implemented a crude form of recency bias in our simulations (biasing retrieval of the most recently learned representations during learning; see simulation 1), but future work should focus on development of more principled mechanisms for constraining learning and storage. Such mechanisms might focus on either constraining when learning takes place, or when the results of learning are stored for future processing. Most likely, though, it will be necessary to account for both. 

Constraining when DORA learns amounts to constraining when it performs comparison. We have previously proposed several possible constraints on comparison such as language (e.g., shared labels) and object salience, and have shown how direction to compare (i.e., instruction) serves as a very powerful constraint on learning \parencite[see][]{doumas2008theory, doumas2013comparison}. These constraints may also serve to limit when the results of learning are stored in memory. DORA might be extended or integrated with existing accounts of language or perceptual (feature) processing to implement such constraints \parencite[see][]{martin2017mechanism}. 

Both these limitations might be addressed by refining DORA’s control structure. The quality of comparisons DORA makes and the representations it learns may serve as important constraints on the control process it uses. Reinforcement learning provides a useful tool for implementing these constraints. 

Second, as pointed out in Simulation 2, we don’t have yet a complete solution for the problem of how to select the right representations to build a relational model of a domain from the reward signal when the domain of potential relations is large. In artificial intelligence the problem of learning relational (a.k.a., first-order) policies has been studied under the name of relational reinforcement learning \parencite{dvzeroski2001relational, driessens2003relational, driessens2004integrating}, but these early models do not scale well to large problems involving multiple relations. However, recent models based on differentiable versions of inductive logic programming
\parencite{evans2018learning, jiang2019neural} seem a promising approach to this problem. These systems have shown that it is possible to use gradient descent methods to prune a prebuilt large set of rules to obtain a program (i.e., a refined sets of rules) that allows the agent to interact effectively with the environment. We are currently working to toward integrating this kind of error-correction learning with DORA. 

The discovery of invariance has relevance beyond the few problems presented here. For example, detecting invariants in speech and language is a defining and unsolved problem in language acquisition and adult speech processing, including in automatic speech recognition by machines. Similarly, whether the generalization of grammatical rules can be fully accounted for in systems that rely on statistical learning alone remains contentious. The account of learning invariance from experience offered here, combined with principles like the compression of role information \parencite{doumas2005neural}, may present new computational vistas on these classic problems in the language sciences \parencite[see][]{martin2017mechanism,martin2016language}. Systems with the properties of DORA may offer an inroad to representational sufficiency across multiple domains, built from the same mechanisms and computational primitives.

\subsection{Conclusion}

A cognitive architecture that is prepared to learn structured representations of relations is prepared to generalize broadly based on those relations. This kind of generalization includes cross-domain transfer as a special case. In fact, it is a mundane consequence of the way people conceptualize the world. 

Purely statistical learning systems will most likely continue to outperform people at any single task on which we choose to train them. But people, and cognitive architectures capable learning relations in an open-ended fashion, will continue to outperform any finite set of purely statistical systems as generalists. And general intelligence, we argue, is not the capacity to be optimal at one task, but is instead the capacity to excel, albeit imperfectly, at many. 

\printbibliography

@article{Battaglia2018RelationalIB,
    title={Relational inductive biases, deep learning, and graph networks},
    author={Peter W. Battaglia and Jessica B. Hamrick and Victor Bapst and Alvaro Sanchez-Gonzalez and Vinicius Zambaldi and Mateusz Malinowski and Andrea Tacchetti and David Raposo and Adam Santoro and Ryan Faulkner and Caglar Gulcehre and Francis Song and Andrew Ballard and Justin Gilmer and George Dahl and Ashish Vaswani and Kelsey Allen and Charles Nash and Victoria Langston and Chris Dyer and Nicolas Heess and Daan Wierstra and Pushmeet Kohli and Matt Botvinick and Oriol Vinyals and Yujia Li and Razvan Pascanu},
    year={2018},
    journal={arXiv preprint arXiv:1806.01261}
}

@article{doumas2018learning,
  title={Learning structured representations from experience},
  author={Doumas, Leonidas A and Martin, Andrea E},
  journal={Psychology of Learning and Motivation},
  volume={69},
  pages={165--203},
  year={2018},
  publisher={Academic Press}
}

@article{doumas2013comparison,
  title={Comparison and mapping facilitate relation discovery and predication},
  author={Doumas, Leonidas A and Hummel, John E},
  journal={PloS one},
  volume={8},
  number={6},
  pages={e63889},
  year={2013},
  publisher={Public Library of Science}
}

@article{doumas2008theory,
  title={A theory of the discovery and predication of relational concepts.},
  author={Doumas, Leonidas A and Hummel, John E and Sandhofer, Catherine M},
  journal={Psychological review},
  volume={115},
  number={1},
  pages={1},
  year={2008},
  publisher={American Psychological Association}
}

@incollection{doumas2005approaches,
  title={Approaches to modeling human mental representations: What works, what doesn’t and why},
  author={Doumas, Leonidas A and Hummel, John E},
  booktitle={The Cambridge handbook of thinking and reasoning}, editor={KJ Holyoak and RG Morrison},
  publisher={Cambridge University Press Cambridge},
  pages={73--94},
  year={2005}
}

@incollection{doumas2012computational,
  title={Computational models of higher cognition},
  author={Doumas, Leonidas A and Hummel, John E},
  booktitle={Oxford handbook of thinking and reasoning},
  editor={KJ Holyoak and RG Morrison},
  publisher={Oxford University Press},
  pages={52--66},
  year={2012}
}

@article{falkenhainer1989structure,
  title={The structure-mapping engine: Algorithm and examples},
  author={Falkenhainer, Brian and Forbus, Kenneth D and Gentner, Dedre},
  journal={Artificial intelligence},
  volume={41},
  number={1},
  pages={1--63},
  year={1989},
  publisher={Elsevier}
}

@article{halford1998processing,
  title={Processing capacity defined by relational complexity: Implications for comparative, developmental, and cognitive psychology},
  author={Halford, Graeme S and Wilson, William H and Phillips, Steven},
  journal={Behavioral and Brain Sciences},
  volume={21},
  number={6},
  pages={803--831},
  year={1998},
  publisher={Cambridge University Press}
}

@inproceedings{hill2018learning,
    title={Learning to Make Analogies by Contrasting Abstract Relational Structure},
    author={Felix Hill and Adam Santoro and David Barrett and Ari Morcos and Timothy Lillicrap},
    booktitle={International Conference on Learning Representations},
    year={2019},
}

@article{holyoak2012analogy,
  title={Analogy and relational reasoning},
  author={Holyoak, Keith J},
  journal={The Oxford handbook of thinking and reasoning},
  pages={234--259},
  year={2012}
}

@article{hummel1997distributed,
  title={Distributed representations of structure: A theory of analogical access and mapping.},
  author={Hummel, John E and Holyoak, Keith J},
  journal={Psychological review},
  volume={104},
  number={3},
  pages={427},
  year={1997},
  publisher={American Psychological Association}
}

@article{hummel2003symbolic,
  title={A symbolic-connectionist theory of relational inference and generalization.},
  author={Hummel, John E and Holyoak, Keith J},
  journal={Psychological review},
  volume={110},
  number={2},
  pages={220},
  year={2003},
  publisher={American Psychological Association}
}

@article{leech2008analogy,
  title={Analogy as relational priming: A developmental and computational perspective on the origins of a complex cognitive skill},
  author={Leech, Robert and Mareschal, Denis and Cooper, Richard P},
  journal={Behavioral and Brain Sciences},
  volume={31},
  number={4},
  pages={357--378},
  year={2008},
  publisher={Cambridge University Press}
}

@article{lu2012bayesian,
  title={Bayesian analogy with relational transformations.},
  author={Lu, Hongjing and Chen, Dawn and Holyoak, Keith J},
  journal={Psychological review},
  volume={119},
  number={3},
  pages={617},
  year={2012},
  publisher={American Psychological Association}
}

@article{martin2017mechanism,
  title={A mechanism for the cortical computation of hierarchical linguistic structure},
  author={Martin, Andrea E and Doumas, Leonidas AA},
  journal={PLoS Biology},
  volume={15},
  number={3},
  pages={e2000663},
  year={2017},
  publisher={Public Library of Science}
}

@article{medin1993respects,
  title={Respects for similarity.},
  author={Medin, Douglas L and Goldstone, Robert L and Gentner, Dedre},
  journal={Psychological review},
  volume={100},
  number={2},
  pages={254},
  year={1993},
  publisher={American Psychological Association}
}

@inproceedings{mikolov2013distributed,
  title={Distributed representations of words and phrases and their compositionality},
  author={Mikolov, Tomas and Sutskever, Ilya and Chen, Kai and Corrado, Greg S and Dean, Jeff},
  booktitle={Advances in neural information processing systems},
  pages={3111--3119},
  year={2013}
}

@article{penn2008darwin,
  title={Darwin's mistake: Explaining the discontinuity between human and nonhuman minds},
  author={Penn, Derek C and Holyoak, Keith J and Povinelli, Daniel J},
  journal={Behavioral and Brain Sciences},
  volume={31},
  number={2},
  pages={109--130},
  year={2008},
  publisher={Cambridge University Press}
}

@inproceedings{santoro2017simple,
  title={A simple neural network module for relational reasoning},
  author={Santoro, Adam and Raposo, David and Barrett, David G and Malinowski, Mateusz and Pascanu, Razvan and Battaglia, Peter and Lillicrap, Timothy},
  booktitle={Advances in neural information processing systems},
  pages={4967--4976},
  year={2017}
}

@article{shastri1993simple,
  title={From simple associations to systematic reasoning: A connectionist representation of rules, variables and dynamic bindings using temporal synchrony},
  author={Shastri, Lokendra and Ajjanagadde, Venkat},
  journal={Behavioral and brain sciences},
  volume={16},
  number={3},
  pages={417--451},
  year={1993},
  publisher={Cambridge University Press}
}

@article{martin2019predicate,
  title={Predicate learning in neural systems: using oscillations to discover latent structure},
  author={Martin, Andrea E and Doumas, Leonidas AA},
  journal={Current Opinion in Behavioral Sciences},
  volume={29},
  pages={77--83},
  year={2019},
  publisher={Elsevier}
}

@article{kim2018not,
  title={Not-So-CLEVR: learning same--different relations strains feedforward neural networks},
  author={Kim, Junkyung and Ricci, Matthew and Serre, Thomas},
  journal={Interface focus},
  volume={8},
  number={4},
  pages={20180011},
  year={2018},
  publisher={The Royal Society}
}

@article{funke2021five,
  title={Five points to check when comparing visual perception in humans and machines},
  author={Funke, Christina M and Borowski, Judy and Stosio, Karolina and Brendel, Wieland and Wallis, Thomas SA and Bethge, Matthias},
  journal={Journal of Vision},
  volume={21},
  number={3},
  pages={16--16},
  year={2021},
  publisher={The Association for Research in Vision and Ophthalmology}
}

@article{MESSINA202175,
title = "Solving the same-different task with convolutional neural networks",
journal = "Pattern Recognition Letters",
volume = "143",
pages = "75 - 80",
year = "2021",
issn = "0167-8655",
author = "Nicola Messina and Giuseppe Amato and Fabio Carrara and Claudio Gennaro and Fabrizio Falchi"
}

@article{ricci2021same,
  title={Same-different conceptualization: a machine vision perspective},
  author={Ricci, Matthew and Cad{\`e}ne, R{\'e}mi and Serre, Thomas},
  journal={Current Opinion in Behavioral Sciences},
  volume={37},
  pages={47--55},
  year={2021},
  publisher={Elsevier}
}

@article{stabinger2020evaluating,
  title={Evaluating the Progress of Deep Learning for Visual Relational Concepts},
  author={Stabinger, Sebastian and David, Peer and Piater, Justus and Rodr{\'i}guez-S{\'a}nchez, Antonio},
  journal={arXiv preprint arXiv:2001.10857},
  year={2020}
}

@article{hummel1992dynamic,
  title={Dynamic binding in a neural network for shape recognition.},
  author={Hummel, John E and Biederman, Irving},
  journal={Psychological review},
  volume={99},
  number={3},
  pages={480},
  year={1992},
  publisher={American Psychological Association}
}

@inproceedings{johnson2017clevr,
  title={Clevr: A diagnostic dataset for compositional language and elementary visual reasoning},
  author={Johnson, Justin and Hariharan, Bharath and Van Der Maaten, Laurens and Fei-Fei, Li and Lawrence Zitnick, C and Girshick, Ross},
  booktitle={Proceedings of the IEEE conference on computer vision and pattern recognition},
  pages={2901--2910},
  year={2017}
}

@inproceedings{gamrian2019transfer,
  title={Transfer learning for related reinforcement learning tasks via image-to-image translation},
  author={Gamrian, Shani and Goldberg, Yoav},
  booktitle={International Conference on Machine Learning},
  pages={2063--2072},
  year={2019},
  organization={PMLR}
}

@inproceedings{kansky2017schema,
  title={Schema networks: Zero-shot transfer with a generative causal model of intuitive physics},
  author={Kansky, Ken and Silver, Tom and M{\'e}ly, David A and Eldawy, Mohamed and L{\'a}zaro-Gredilla, Miguel and Lou, Xinghua and Dorfman, Nimrod and Sidor, Szymon and Phoenix, Scott and George, Dileep},
  booktitle={International Conference on Machine Learning},
  pages={1809--1818},
  year={2017},
  organization={PMLR}
}

@article{zhang2018study,
  title={A study on overfitting in deep reinforcement learning},
  author={Zhang, Chiyuan and Vinyals, Oriol and Munos, Remi and Bengio, Samy},
  journal={arXiv preprint arXiv:1804.06893},
  year={2018}
}

@article{bowers2017parallel,
  title={Parallel distributed processing theory in the age of deep networks},
  author={Bowers, Jeffrey S},
  journal={Trends in cognitive sciences},
  volume={21},
  number={12},
  pages={950--961},
  year={2017},
  publisher={Elsevier}
}

@article{geirhos2020shortcut,
  title={Shortcut learning in deep neural networks},
  author={Geirhos, Robert and Jacobsen, J{\"o}rn-Henrik and Michaelis, Claudio and Zemel, Richard and Brendel, Wieland and Bethge, Matthias and Wichmann, Felix A},
  journal={Nature Machine Intelligence},
  volume={2},
  number={11},
  pages={665--673},
  year={2020},
  publisher={Nature Publishing Group}
}

@article{carey2000origin,
  title={The origin of concepts},
  author={Carey, Susan},
  journal={Journal of Cognition and Development},
  volume={1},
  number={1},
  pages={37--41},
  year={2000},
  publisher={Taylor \& Francis}
}

@book{susan2009origin,
  title={The origin of concepts},
  author={Carey, Susan},
  year={2009},
  publisher={Oxford University Press, New York}
}

@article{murphy1985role,
  title={The role of theories in conceptual coherence.},
  author={Murphy, Gregory L and Medin, Douglas L},
  journal={Psychological review},
  volume={92},
  number={3},
  pages={289},
  year={1985},
  publisher={American Psychological Association}
}

@article{hespos2001infants,
  title={Infants' knowledge about occlusion and containment events: A surprising discrepancy},
  author={Hespos, Susan J and Baillargeon, Ren{\'e}e},
  journal={Psychological Science},
  volume={12},
  number={2},
  pages={141--147},
  year={2001},
  publisher={SAGE Publications Sage CA: Los Angeles, CA}
}

@book{piaget1954construction,
  title={The construction of reality in the child},
  author={Piaget, Jean},
  year={1954},
  publisher={New York: Basic Books}
}

@book{gelman2003essential,
  title={The essential child: Origins of essentialism in everyday thought},
  author={Gelman, Susan A and others},
  year={2003},
  publisher={Oxford University Press, USA}
}

@article{paccanaro2001learning,
  title={Learning distributed representations of concepts using linear relational embedding},
  author={Paccanaro, Alberto and Hinton, Geoffrey E.},
  journal={IEEE Transactions on Knowledge and Data Engineering},
  volume={13},
  number={2},
  pages={232--244},
  year={2001},
  publisher={IEEE}
}

@inproceedings{haldekar2017identifying,
  title={Identifying spatial relations in images using convolutional neural networks},
  author={Haldekar, Mandar and Ganesan, Ashwinkumar and Oates, Tim},
  booktitle={2017 International Joint Conference on Neural Networks (IJCNN)},
  pages={3593--3600},
  year={2017},
  organization={IEEE}
}

@inproceedings{cadene2019murel,
  title={Murel: Multimodal relational reasoning for visual question answering},
  author={Cadene, Remi and Ben-Younes, Hedi and Cord, Matthieu and Thome, Nicolas},
  booktitle={Proceedings of the IEEE/CVF Conference on Computer Vision and Pattern Recognition},
  pages={1989--1998},
  year={2019}
}

@inproceedings{ma2018visual,
  title={Visual question answering with memory-augmented networks},
  author={Ma, Chao and Shen, Chunhua and Dick, Anthony and Wu, Qi and Wang, Peng and van den Hengel, Anton and Reid, Ian},
  booktitle={Proceedings of the IEEE Conference on Computer Vision and Pattern Recognition},
  pages={6975--6984},
  year={2018}
}

@inproceedings{xu2016ask,
  title={Ask, attend and answer: Exploring question-guided spatial attention for visual question answering},
  author={Xu, Huijuan and Saenko, Kate},
  booktitle={European Conference on Computer Vision},
  pages={451--466},
  year={2016},
  organization={Springer}
}

@article{hoshen2017iq,
  title={The IQ of neural networks},
  author={Hoshen, Dokhyam and Werman, Michael},
  journal={arXiv preprint arXiv:1710.01692},
  year={2017}
}

@article{peterson2020parallelograms,
  title={Parallelograms revisited: Exploring the limitations of vector space models for simple analogies},
  author={Peterson, Joshua C and Chen, Dawn and Griffiths, Thomas L},
  journal={Cognition},
  volume={205},
  pages={104440},
  year={2020},
  publisher={Elsevier}
}

@article{lu2021probabilistic,
  title={Probabilistic Analogical Mapping with Semantic Relation Networks},
  author={Lu, Hongjing and Ichien, Nicholas and Holyoak, Keith J},
  journal={arXiv preprint arXiv:2103.16704},
  year={2021}
}

@book{anderson2007can,
  title={How can the human mind occur in the physical universe?},
  author={Anderson, John R},
  year={2007},
  publisher={Oxford University Press}
}

@article{sengupta2018robust,
  title={How robust are deep neural networks?},
  author={Sengupta, Biswa and Friston, Karl J},
  journal={arXiv preprint arXiv:1804.11313},
  year={2018}
}

@article{reichert2013neuronal,
  title={Neuronal synchrony in complex-valued deep networks},
  author={Reichert, David P and Serre, Thomas},
  journal={arXiv preprint arXiv:1312.6115},
  year={2013}
}

@inproceedings{rao2011effects,
  title={The effects of feedback and lateral connections on perceptual processing: A study using oscillatory networks},
  author={Rao, A Ravishankar and Cecchi, Guillermo A},
  booktitle={The 2011 International Joint Conference on Neural Networks},
  pages={1177--1184},
  year={2011},
  organization={IEEE}
}

@article{rao2010objective,
  title={An objective function utilizing complex sparsity for efficient segmentation in multi-layer oscillatory networks},
  author={Rao, A Ravishankar and Cecchi, Guillermo A},
  journal={International Journal of Intelligent Computing and Cybernetics},
  year={2010},
  publisher={Emerald Group Publishing Limited}
}

@book{holyoak1995mental,
  title={Mental leaps: Analogy in creative thought},
  author={Holyoak, Keith J and Holyoak, Keith James and Thagard, Paul},
  year={1995},
  publisher={MIT press}
}

@article{knowlton2012neurocomputational,
  title={A neurocomputational system for relational reasoning},
  author={Knowlton, Barbara J and Morrison, Robert G and Hummel, John E and Holyoak, Keith J},
  journal={Trends in cognitive sciences},
  volume={16},
  number={7},
  pages={373--381},
  year={2012},
  publisher={Elsevier}
}

@article{doumas2010computational,
  title={A computational account of the development of the generalization of shape information},
  author={Doumas, Leonidas AA and Hummel, John E},
  journal={Cognitive science},
  volume={34},
  number={4},
  pages={698--712},
  year={2010},
  publisher={Wiley Online Library}
}

@article{hummel2001complementary,
  title={Complementary solutions to the binding problem in vision: Implications for shape perception and object recognition},
  author={Hummel, John E},
  journal={Visual cognition},
  volume={8},
  number={3-5},
  pages={489--517},
  year={2001},
  publisher={Taylor \& Francis}
}

@article{choplin2002magnitude,
  title={Magnitude comparisons distort mental representations of magnitude.},
  author={Choplin, Jessica M and Hummel, John E},
  journal={Journal of Experimental Psychology: General},
  volume={131},
  number={2},
  pages={270},
  year={2002},
  publisher={American Psychological Association}
}

@article{krawczyk2004structural,
  title={Structural constraints and object similarity in analogical mapping and inference},
  author={Krawczyk, Daniel C and Holyoak, Keith J and Hummel, John E},
  journal={Thinking \& reasoning},
  volume={10},
  number={1},
  pages={85--104},
  year={2004},
  publisher={Taylor \& Francis}
}

@article{krawczyk2005one,
  title={The one-to-one constraint in analogical mapping and inference},
  author={Krawczyk, Daniel C and Holyoak, Keith J and Hummel, John E},
  journal={Cognitive science},
  volume={29},
  number={5},
  pages={797--806},
  year={2005},
  publisher={Wiley Online Library}
}

@article{kroger2004varieties,
  title={Varieties of sameness: The impact of relational complexity on perceptual comparisons},
  author={Kroger, James K and Holyoak, Keith J and Hummel, John E},
  journal={Cognitive Science},
  volume={28},
  number={3},
  pages={335--358},
  year={2004},
  publisher={Wiley Online Library}
}

@article{kubose2002role,
  title={The role of textual coherence in incremental analogical mapping},
  author={Kubose, Tate T and Holyoak, Keith J and Hummel, John E},
  journal={Journal of memory and language},
  volume={47},
  number={3},
  pages={407--435},
  year={2002},
  publisher={Elsevier}
}

@article{taylor2009finding,
  title={Finding similarity in a model of relational reasoning},
  author={Taylor, Eric G and Hummel, John E},
  journal={Cognitive Systems Research},
  volume={10},
  number={3},
  pages={229--239},
  year={2009},
  publisher={Elsevier}
}

@article{jung2015revisiting,
  title={Revisiting Wittgenstein’s puzzle: hierarchical encoding and comparison facilitate learning of probabilistic relational categories},
  author={Jung, Wookyoung and Hummel, John E},
  journal={Frontiers in psychology},
  volume={6},
  pages={110},
  year={2015},
  publisher={Frontiers}
}

@article{jung2015making,
  title={Making probabilistic relational categories learnable},
  author={Jung, Wookyoung and Hummel, John E},
  journal={Cognitive science},
  volume={39},
  number={6},
  pages={1259--1291},
  year={2015},
  publisher={Wiley Online Library}
}

@article{livins2015recognising,
  title={Recognising relations: What can be learned from considering complexity},
  author={Livins, Katherine A and Doumas, Leonidas AA},
  journal={Thinking \& Reasoning},
  volume={21},
  number={3},
  pages={251--264},
  year={2015},
  publisher={Taylor \& Francis}
}

@article{livins2015varying,
  title={Varying variation: the effects of within-versus across-feature differences on relational category learning},
  author={Livins, Katherine A and Spivey, Michael J and Doumas, Leonidas AA},
  journal={Frontiers in psychology},
  volume={6},
  pages={129},
  year={2015},
  publisher={Frontiers}
}

@article{livins2016shaping,
  title={Shaping relations: Exploiting relational features for visuospatial priming.},
  author={Livins, Katherine A and Doumas, Leonidas AA and Spivey, Michael J},
  journal={Journal of experimental psychology: learning, memory, and cognition},
  volume={42},
  number={1},
  pages={127},
  year={2016},
  publisher={American Psychological Association}
}

@inproceedings{licato2012exploring,
  title={Exploring the role of analogico-deductive reasoning in the balance-beam task},
  author={Licato, John and Bringsjord, Selmer and Hummel, John E},
  booktitle={Rethinking Cognitive Development: Proceedings of the 42nd Annual Meeting of the Jean Piaget Society},
  year={2012}
}

@inproceedings{lim2013modeling,
  title={Modeling the Relational Shift in Melodic Processing of Young Children},
  author={Lim, Ahnate and Doumas, Leonidas AA and Sinnett, Scott},
  booktitle={Proceedings of the Annual Meeting of the Cognitive Science Society},
  volume={35},
  number={35},
  year={2013}
}

@article{sandhofer2008order,
  title={Order of presentation effects in learning color categories},
  author={Sandhofer, Catherine M and Doumas, Leonidas AA},
  journal={Journal of Cognition and Development},
  volume={9},
  number={2},
  pages={194--221},
  year={2008},
  publisher={Taylor \& Francis}
}

@article{rabagliati2017representing,
  title={Representing composed meanings through temporal binding},
  author={Rabagliati, Hugh and Doumas, Leonidas AA and Bemis, Douglas K},
  journal={Cognition},
  volume={162},
  pages={61--72},
  year={2017},
  publisher={Elsevier}
}

@article{viskontas2004relational,
  title={Relational integration, inhibition, and analogical reasoning in older adults.},
  author={Viskontas, Indre V and Morrison, Robert G and Holyoak, Keith J and Hummel, John E and Knowlton, Barbara J},
  journal={Psychology and aging},
  volume={19},
  number={4},
  pages={581},
  year={2004},
  publisher={American Psychological Association}
}

@article{morrison2004neurocomputational,
  title={A neurocomputational model of analogical reasoning and its breakdown in frontotemporal lobar degeneration},
  author={Morrison, Robert G and Krawczyk, Daniel C and Holyoak, Keith J and Hummel, John E and Chow, Tiffany W and Miller, Bruce L and Knowlton, Barbara J},
  journal={Journal of cognitive neuroscience},
  volume={16},
  number={2},
  pages={260--271},
  year={2004},
  publisher={MIT Press}
}

@article{morrison2011computational,
  title={A computational account of children’s analogical reasoning: balancing inhibitory control in working memory and relational representation},
  author={Morrison, Robert G and Doumas, Leonidas AA and Richland, Lindsey E},
  journal={Developmental science},
  volume={14},
  number={3},
  pages={516--529},
  year={2011},
  publisher={Wiley Online Library}
}

@article{cowan2001magical,
  title={The magical number 4 in short-term memory: A reconsideration of mental storage capacity},
  author={Cowan, Nelson},
  journal={Behavioral and brain sciences},
  volume={24},
  number={1},
  pages={87--114},
  year={2001},
  publisher={Cambridge University Press}
}

@article{gentner1983structure,
  title={Structure-mapping: A theoretical framework for analogy},
  author={Gentner, Dedre},
  journal={Cognitive science},
  volume={7},
  number={2},
  pages={155--170},
  year={1983},
  publisher={Elsevier}
}

@incollection{holyoak1994component,
  title={Component processes in analogical transfer: Mapping, pattern completion, and adaptation},
  author={Holyoak, Keith J and Novick, Laura R and Melz, Eric R},
  booktitle={Advances in connectionist and neural computation theory, Vol. 2. Analogical connections},
  editor = {KJ Holyoak and JA Barnden},
  pages={113--180},
  year={1994},
  publisher={Ablex Publishing}
}

@incollection{biederman2013human,
  title={Human object recognition: Appearance vs. shape},
  author={Biederman, Irving},
  booktitle={Shape perception in human and computer vision},
  pages={387--397},
  year={2013},
  publisher={Springer}
}

@article{harnad1990symbol,
  title={The symbol grounding problem},
  author={Harnad, Stevan},
  journal={Physica D: Nonlinear Phenomena},
  volume={42},
  number={1-3},
  pages={335--346},
  year={1990},
  publisher={Elsevier}
}

@inproceedings{kellman2020modeling,
  title={Modeling perceptual learning of abstract invariants},
  author={Kellman, Philip J and Burke, Timothy and Hummel, John E},
  booktitle={Proceedings of the Twenty First Annual Conference of the Cognitive Science Society},
  pages={264--269},
  year={2020},
  organization={Psychology Press}
}

@article{gallistel2000non,
  title={Non-verbal numerical cognition: From reals to integers},
  author={Gallistel, Charles R and Gelman, Rochel},
  journal={Trends in cognitive sciences},
  volume={4},
  number={2},
  pages={59--65},
  year={2000},
  publisher={Elsevier}
}

@article{zorzi2005computational,
  title={Computational modeling of numerical cognition},
  author={Zorzi, Marco and Stoianov, Ivilin and Umilt{\`a}, Carlo},
  journal={Handbook of mathematical cognition},
  volume={19},
  pages={67--84},
  year={2005},
  publisher={Psychology Press Hove, England}
}

@article{horn1991segmentation,
  title={Segmentation, binding, and illusory conjunctions},
  author={Horn, D and Sagi, Dov and Usher, Marius},
  journal={Neural computation},
  volume={3},
  number={4},
  pages={510--525},
  year={1991},
  publisher={MIT Press}
}

@article{horn1990excitatory,
  title={Excitatory--inhibitory networks with dynamical thresholds},
  author={Horn, David and Usher, Marius},
  journal={International Journal of Neural Systems},
  volume={1},
  number={03},
  pages={249--257},
  year={1990},
  publisher={World Scientific}
}

@article{usher1996modeling,
  title={Modeling the temporal dynamics of IT neurons in visual search: A mechanism for top-down selective attention},
  author={Usher, Marius and Niebur, Ernst},
  journal={Journal of cognitive neuroscience},
  volume={8},
  number={4},
  pages={311--327},
  year={1996},
  publisher={MIT Press}
}

@article{von1992sensory,
  title={Sensory segmentation with coupled neural oscillators},
  author={von der Malsburg, Christoph and Buhmann, Joachim},
  journal={Biological cybernetics},
  volume={67},
  number={3},
  pages={233--242},
  year={1992},
  publisher={Springer}
}

@article{bonds1989role,
  title={Role of inhibition in the specification of orientation selectivity of cells in the cat striate cortex},
  author={Bonds, AB},
  journal={Visual neuroscience},
  volume={2},
  number={1},
  pages={41--55},
  year={1989},
  publisher={Cambridge University Press}
}

@article{heeger1992normalization,
  title={Normalization of cell responses in cat striate cortex},
  author={Heeger, David J},
  journal={Visual neuroscience},
  volume={9},
  number={2},
  pages={181--197},
  year={1992}
}

@article{foley1994human,
  title={Human luminance pattern-vision mechanisms: masking experiments require a new model},
  author={Foley, John M},
  journal={JOSA A},
  volume={11},
  number={6},
  pages={1710--1719},
  year={1994},
  publisher={Optical Society of America}
}

@article{thomas1997contrast,
  title={Contrast gain control and fine spatial discriminations},
  author={Thomas, James P and Olzak, Lynn A},
  journal={JOSA A},
  volume={14},
  number={9},
  pages={2392--2405},
  year={1997},
  publisher={Optical Society of America}
}

@article{spelke2007core,
  title={Core knowledge},
  author={Spelke, Elizabeth S and Kinzler, Katherine D},
  journal={Developmental science},
  volume={10},
  number={1},
  pages={89--96},
  year={2007},
  publisher={Wiley Online Library}
}

@article{demer2002orbital,
  title={The orbital pulley system: a revolution in concepts of orbital anatomy},
  author={Demer, Joseph L},
  journal={Annals of the New York Academy of Sciences},
  volume={956},
  number={1},
  pages={17--32},
  year={2002},
  publisher={Wiley Online Library}
}

@article{girshick2011cardinal,
  title={Cardinal rules: visual orientation perception reflects knowledge of environmental statistics},
  author={Girshick, Ahna R and Landy, Michael S and Simoncelli, Eero P},
  journal={Nature neuroscience},
  volume={14},
  number={7},
  pages={926--932},
  year={2011},
  publisher={Nature Publishing Group}
}

@article{marr1980theory,
  title={Theory of edge detection},
  author={Marr, David and Hildreth, Ellen},
  journal={Proceedings of the Royal Society of London. Series B. Biological Sciences},
  volume={207},
  number={1167},
  pages={187--217},
  year={1980},
  publisher={The Royal Society London}
}

@article{wandell2007visual,
  title={Visual field maps in human cortex},
  author={Wandell, Brian A and Dumoulin, Serge O and Brewer, Alyssa A},
  journal={Neuron},
  volume={56},
  number={2},
  pages={366--383},
  year={2007},
  publisher={Elsevier}
}

@article{engel1994fmri,
  title={fMRI of human visual cortex.},
  author={Engel, Stephen A and Rumelhart, David E and Wandell, Brian A and Lee, Adrian T and Glover, Gary H and Chichilnisky, Eduardo-Jose and Shadlen, Michael N},
  journal={Nature},
  year={1994},
  publisher={Nature Publishing Group}
}

@article{furmanski2000oblique,
  title={An oblique effect in human primary visual cortex},
  author={Furmanski, Christopher S and Engel, Stephen A},
  journal={Nature neuroscience},
  volume={3},
  number={6},
  pages={535--536},
  year={2000},
  publisher={Nature Publishing Group}
}

@article{moore2001neural,
  title={Neural response to perception of volume in the lateral occipital complex},
  author={Moore, Cassandra and Engel, Stephen A},
  journal={Neuron},
  volume={29},
  number={1},
  pages={277--286},
  year={2001},
  publisher={Elsevier}
}

@article{logie2020working,
  title={Working memory: The state of the science},
  author={Logie, Robert and Camos, Val{\'e}rie and Cowan, Nelson},
  year={2020},
  publisher={Oxford University Press}
}

@article{sophian_2008,
    title={Precursors to number: Equivalence relations, less-than and greater-than relations, and units},
    volume={31},
    doi={10.1017/S0140525X08005876},
    number={6},
    journal={Behavioral and Brain Sciences},
    publisher={Cambridge University Press},
    author={Sophian, Catherine},
    year={2008},
    pages={670–671}}

@article{sandhofer1999learning,
  title={Learning color words involves learning a system of mappings.},
  author={Sandhofer, Catherine M and Smith, Linda B},
  journal={Developmental Psychology},
  volume={35},
  number={3},
  pages={668},
  year={1999},
  publisher={American Psychological Association}
}

@article{sandhofer2000counting,
  title={Counting nouns and verbs in the input: Differential frequencies, different kinds of learning?},
  author={Sandhofer, Catherine M and Smith, Linda B and Luo, Jun},
  journal={Journal of child language},
  volume={27},
  number={3},
  pages={561--585},
  year={2000},
  publisher={Cambridge University Press}
}

@article{mnih2015human,
  title={Human-level control through deep reinforcement learning},
  author={Mnih, Volodymyr and Kavukcuoglu, Koray and Silver, David and Rusu, Andrei A and Veness, Joel and Bellemare, Marc G and Graves, Alex and Riedmiller, Martin and Fidjeland, Andreas K and Ostrovski, Georg and others},
  journal={nature},
  volume={518},
  number={7540},
  pages={529--533},
  year={2015},
  publisher={Nature Publishing Group}
}

@article{nye2020learning,
  title={Learning compositional rules via neural program synthesis},
  author={Nye, Maxwell I and Solar-Lezama, Armando and Tenenbaum, Joshua B and Lake, Brenden M},
  journal={arXiv preprint arXiv:2003.05562},
  year={2020}
}

@article{lake2015human,
  title={Human-level concept learning through probabilistic program induction},
  author={Lake, Brenden M and Salakhutdinov, Ruslan and Tenenbaum, Joshua B},
  journal={Science},
  volume={350},
  number={6266},
  pages={1332--1338},
  year={2015},
  publisher={American Association for the Advancement of Science}
}

@book{sutton2018reinforcement,
  title={Reinforcement learning: An introduction},
  author={Sutton, Richard S and Barto, Andrew G},
  year={2018},
  publisher={MIT press}
}

@article{gershman2018deconstructing,
  title={Deconstructing the human algorithms for exploration},
  author={Gershman, Samuel J},
  journal={Cognition},
  volume={173},
  pages={34--42},
  year={2018},
  publisher={Elsevier}
}

@article{otto2010regulatory,
  title={Regulatory fit and systematic exploration in a dynamic decision-making environment.},
  author={Otto, A Ross and Markman, Arthur B and Gureckis, Todd M and Love, Bradley C},
  journal={Journal of Experimental Psychology: Learning, Memory, and Cognition},
  volume={36},
  number={3},
  pages={797},
  year={2010},
  publisher={American Psychological Association}
}

@article{rich2018limits,
  title={The limits of learning: Exploration, generalization, and the development of learning traps.},
  author={Rich, Alexander S and Gureckis, Todd M},
  journal={Journal of Experimental Psychology: General},
  volume={147},
  number={11},
  pages={1553},
  year={2018},
  publisher={American Psychological Association}
}

@phdthesis{watkins1989learning,
  title={Learning from delayed rewards.},
  author={Watkins, Christopher John Cornish Hellaby},
  year={1989},
  school={University of Cambridge}
}

@article{hummel2011getting,
  title={Getting symbols out of a neural architecture},
  author={Hummel, John E},
  journal={Connection Science},
  volume={23},
  number={2},
  pages={109--118},
  year={2011},
  publisher={Taylor \& Francis}
}

@article{french1999catastrophic,
  title={Catastrophic forgetting in connectionist networks},
  author={French, Robert M},
  journal={Trends in cognitive sciences},
  volume={3},
  number={4},
  pages={128--135},
  year={1999},
  publisher={Elsevier}
}

@article{kirkpatrick2017overcoming,
  title={Overcoming catastrophic forgetting in neural networks},
  author={Kirkpatrick, James and Pascanu, Razvan and Rabinowitz, Neil and Veness, Joel and Desjardins, Guillaume and Rusu, Andrei A and Milan, Kieran and Quan, John and Ramalho, Tiago and Grabska-Barwinska, Agnieszka and others},
  journal={Proceedings of the national academy of sciences},
  volume={114},
  number={13},
  pages={3521--3526},
  year={2017},
  publisher={National Acad Sciences}
}

@article{bassok1995judging,
  title={Judging a book by its cover: Interpretative effects of content on problem-solving transfer},
  author={Bassok, Miriam and Olseth, Karen L},
  journal={Memory \& Cognition},
  volume={23},
  number={3},
  pages={354--367},
  year={1995},
  publisher={Springer}
}

@article{gick1983schema,
  title={Schema induction and analogical transfer},
  author={Gick, Mary L and Holyoak, Keith J},
  journal={Cognitive Psychology},
  volume={15},
  number={1},
  pages={1--38},
  year={1983},
  publisher={Elsevier}
}

@article{richland2006children,
  title={Children’s development of analogical reasoning: Insights from scene analogy problems},
  author={Richland, Lindsey E and Morrison, Robert G and Holyoak, Keith J},
  journal={Journal of experimental child psychology},
  volume={94},
  number={3},
  pages={249--273},
  year={2006},
  publisher={Elsevier}
}

@article{smith1984young,
  title={Young children's understanding of attributes and dimensions: A comparison of conceptual and linguistic measures},
  author={Smith, Linda B},
  journal={Child development},
  pages={363--380},
  year={1984},
  publisher={JSTOR}
}

@article{nelson1974comprehension,
  title={The comprehension of relative, absolute, and contrastive adjectives by young children},
  author={Nelson, Katherine and Benedict, Helen},
  journal={Journal of Psycholinguistic Research},
  volume={3},
  number={4},
  pages={333--342},
  year={1974},
  publisher={Springer}
}

@incollection{gentner1995two,
  title={Two forces in the development of relational similarity},
  author={Gentner, Dedre and Rattermann, Mary Jo and Markman, Arthur and Kotovsky, Laura},
  booktitle={Developing cognitive competence: New approaches to process modeling},
  editor = {T J Simon and G S Halford},
  pages={263--313},
  year={1995}
}

@article{halford1980category,
  title={A category theory approach to cognitive development},
  author={Halford, Graeme S and Wilson, William H},
  journal={Cognitive psychology},
  volume={12},
  number={3},
  pages={356--411},
  year={1980},
  publisher={Elsevier}
}

@incollection{rattermann1998effect,
  title={The effect of language on similarity: The use of relational labels improves young children’s performance in a mapping task},
  author={Rattermann, Mary Jo and Gentner, Dedre},
  booktitle={Advances in analogy research: Integration of theory and data from the cognitive, computational, and neural sciences},
  editor = {K Holyoak, D Gentner, \& B Kokinov},
  year={1998},
  publisher={Sophia: New Bulgarian University}
}

@incollection{goldstone2012similarity,
  title={Similarity},
  author={Goldstone, Robert L and Son, Ji Yun},
  booktitle={The Oxford handbook of thinking and reasoning},
  editor = {KJ Holyoak and RG Morrison},
  pages={155--176},
  year={2012},
  publisher={Oxford University Press}
}

@article{halford1998induction,
  title={Induction of relational schemas: Common processes in reasoning and complex learning},
  author={Halford, Graeme S and Bain, John D and Maybery, Murray T and Andrews, Glenda},
  journal={Cognitive psychology},
  volume={35},
  number={3},
  pages={201--245},
  year={1998},
  publisher={Elsevier}
}

@article{phillips2018going,
  title={Going beyond the data as the patching (sheaving) of local knowledge},
  author={Phillips, Steven},
  journal={Frontiers in psychology},
  volume={9},
  pages={1926},
  year={2018},
  publisher={Frontiers}
}

@article{phillips2021reconstruction,
  title={A reconstruction theory of relational schema induction},
  author={Phillips, Steven},
  journal={PLoS Computational Biology},
  volume={17},
  number={1},
  pages={e1008641},
  year={2021},
  publisher={Public Library of Science San Francisco, CA USA}
}

@inproceedings{phillips1995processing,
  title={The processing of associations versus the processing of relations and symbols: A systematic comparison},
  author={Phillips, Steven and Halford, Graeme S and Wilson, William H},
  booktitle={Proceedings of the Seventeenth Annual Conference of the Cognitive Science Society},
  pages={688--691},
  year={1995}
}

@article{richardson2003spatial,
  title={Spatial representations activated during real-time comprehension of verbs},
  author={Richardson, Daniel C and Spivey, Michael J and Barsalou, Lawrence W and McRae, Ken},
  journal={Cognitive science},
  volume={27},
  number={5},
  pages={767--780},
  year={2003},
  publisher={Wiley Online Library}
}

@article{michotte1963perception,
  title={The perception of causality.},
  author={Michotte, A},
  year={1963},
  publisher={Basic Books}
}

@book{doumas2005neural,
  title={A neural-network model for discovering relational concepts and learning structured representations},
  author={Doumas, Leonidas AA},
  year={2005},
  publisher={University of California, Los Angeles}
}

@incollection{mints2001arithmetic,
  title={Arithmetic, formal},
  author={Mints, GE},
  booktitle={Encyclopaedia of mathematics},
  pages={63--64},
  editor={Hazewinkel, Michiel},
  year={2001},
  publisher={Springer Berlin, Germany}
}

@article{son2010words,
  title={When do words promote analogical transfer?},
  author={Son, Ji Y and Doumas, Leonidas AA and Goldstone, Robert L},
  journal={The Journal of Problem Solving},
  volume={3},
  number={1},
  pages={4},
  year={2010}
}

@article{ross1989distinguishing,
  title={Distinguishing types of superficial similarities: Different effects on the access and use of earlier problems.},
  author={Ross, Brian H},
  journal={Journal of Experimental Psychology: Learning, memory, and cognition},
  volume={15},
  number={3},
  pages={456},
  year={1989},
  publisher={American Psychological Association}
}

@article{holyoak2008no,
  title={No way to start a space program: Associationism as a launch pad for analogical reasoning},
  author={Holyoak, Keith J and Hummel, John E},
  journal={Behavioral and Brain Sciences},
  volume={31},
  number={4},
  pages={388--389},
  year={2008},
  publisher={Cambridge University Press}
}

@article{dvzeroski2001relational,
  title={Relational reinforcement learning},
  author={D{\v{z}}eroski, Sa{\v{s}}o and De Raedt, Luc and Driessens, Kurt},
  journal={Machine learning},
  volume={43},
  number={1},
  pages={7--52},
  year={2001},
  publisher={Springer}
}

@inproceedings{driessens2003relational,
  title={Relational instance based regression for relational reinforcement learning},
  author={Driessens, Kurt and Ramon, Jan},
  booktitle={Proceedings of the 20th International Conference on Machine Learning (ICML-03)},
  pages={123--130},
  year={2003}
}

@article{driessens2004integrating,
  title={Integrating guidance into relational reinforcement learning},
  author={Driessens, Kurt and D{\v{z}}eroski, Sa{\v{s}}o},
  journal={Machine Learning},
  volume={57},
  number={3},
  pages={271--304},
  year={2004},
  publisher={Springer}
}

@article{evans2018learning,
  title={Learning explanatory rules from noisy data},
  author={Evans, Richard and Grefenstette, Edward},
  journal={Journal of Artificial Intelligence Research},
  volume={61},
  pages={1--64},
  year={2018}
}

@inproceedings{jiang2019neural,
  title={Neural logic reinforcement learning},
  author={Jiang, Zhengyao and Luo, Shan},
  booktitle={International Conference on Machine Learning},
  pages={3110--3119},
  year={2019},
  organization={PMLR}
}

@article{martin2016language,
  title={Language processing as cue integration: Grounding the psychology of language in perception and neurophysiology},
  author={Martin, Andrea E},
  journal={Frontiers in psychology},
  volume={7},
  pages={120},
  year={2016},
  publisher={Frontiers}
}

\begin{appendix}

% format the equation environment
\renewcommand{\theequation}{A.\arabic{equation}}
% reset the counter
\setcounter{equation}{0}

\section{DORA Computational Details}

For completeness, we provide full implementational details of DORA’s operation below. Code for the model is available online at  (\url{github.com/alexdoumas/BrPong_1}). 

\subsection{Parts of DORA}

As described in the main text, DORA consists of a long-term-memory (LTM) composed of three bidirectionally connected layers of units. Units in LTM are referred to as token units (or tokens). Units in the lowest layer of LTM are connected to a common pool of feature units. Token units are yoked to integrative inhibitors that integrate input from their yoked unit and token units in higher layers. 

DORA learns representations of a form we call LISAese via unsupervised learning. Propositions in LISAese are coded by hierarchy of units in layers of a neural network (see main text). At the bottom of the hierarchy, feature (or semantic) nodes code for the featural properties of represented instances in a distributed manner. At the next layer, localist predicate and object units (T1) conjunctively code collections of feature units into representations of objects and roles. At the next layer localist role-binding units (T2) conjunctively bind object and role T1 units into linked role-filler pairs. Finally, proposition units (T3) link T2 units to form whole relational structures. 

Sets, groups of potentiated units, correspond to attention or working memory (WM) within a cognitive framework. The \textit{driver} corresponds to DORA’s current focus of attention. The \textit{recipient} corresponds to active memory. Token units are laterally inhibitive (units in the same layer inhibit one another) within, but not across, sets. 

Each layer of token units is negatively connected to a local inhibitor, and all token unit are connected to a global inhibitor (\textit{I}). Active token units in a layer inhibit the local inhibitor to inactivity. When no token units in a given layer are active, the local inhibitor becomes active, and sends a refresh signal to all tokens in that layer and below across LTM (see below). When no token units in the driver are active, the global inhibitor becomes active, and sends a refresh signal to all tokens across LTM (see below). Each layer of token units is connected to a clamping unit (\textit{C}), that is excited by unclamped units in the layer below and inhibited by unclamped units in the same layer and the layer above (see below). \textit{C} units play a role in recruiting and activating token units during learning. 

We use the term \textit{analog} to refer to a complete story, event, or situation (e.g., from a single object in isolation, to a full propostion in LISAese). Analogs are represented by a collection of token units (T1-T3). Token units are not duplicated within an analog (e.g., within an analog, each proposition that refers to Don connects to the same "Don" unit). Separate analogs do have non-identical token units (e.g., Don will be represented by one T1 unit in one analog and by a different T1 in another analog). The feature units thus represent general type information and token units represent instantiations (or tokens) of those types in specific analogs. 

\subsection{Functional overview of processing in DORA}

In this section we describe DORA’s operation in strictly functional terms. For a detailed description of how these operations are instantiated in the neural network using traditional connectionist computing principles see \textcite{doumas2008theory}. 

\subsubsection{Retrieval}

When there are items in the driver (i.e., DORA is attending to something), but nothing in the rest of AM, then DORA performs retrieval. In short, some representation $b$ is retrieved into the recipient to the extent that it is similar to $a$ in the driver and prevalent in memory, and other representations $c$ in memory are not similar to $a$ and are not prevalent in memory. Functionally, retrieval works as follows: 

\begin{equation}
    Ret(B) \leftarrow f_{Ret}(sim(B, A^D), sim(C \neq B, A^D), p(B), p(C))
    \label{eq:a1}
\end{equation}
% Equation A1

\noindent where, $Ret(B)$ is a retrieved representation $B$, $A^D$ is a driver representation (i.e., a collection of connected token units instantiating a LISAese representation), $p(B)$ is the prevalence of representation $B$ in LTM, and $f_{Ret}$ is the retrieval function.

\subsubsection{Mapping}

When there are items in the driver (i.e., DORA is attending to something), and in the recipient, then DORA performs mapping. In short, representation $a$ in the driver will map to representation $b$ in the recipient to the extent that there are correspondences between $a$ and $b$, and there are not correspondences between a and any other items $c$ in the recipient. Functionally, mapping works as follows:

\begin{equation}
    M(A^D, B^R) \leftarrow f_M(sim(A^D, B^R), sim(A^D, C \neq B^R))
    \label{eq:a2}
\end{equation}
% Equation A2

\noindent where, $M(A, B)$ is a mapping between $A$ and $B$ (instantiated as a learned bidirectional weighted connection), $B^R$ is a recipient representation, $C$ are other representations in the recipient that are not $B^R$, $sim(A, B)$ is a similarity function, and $f_M$ is the mapping function.

\subsubsection{Relation learning}

If DORA has learned mapping connections between representations in driver and recipient, then DORA can learn from the mapping. During learning there are two possibilities. In the first case, if two objects (e.g., $a$ and $b$) that are not already bound to predicates (i.e., no T\textsubscript{2} units are active) are mapped, then DORA learns a single-place predicate composed of the featural intersection of $a$ and $b$. In the second case, when sets of role-filler pairs are mapped---e.g., $P_x(a)$ and $P_y(b)$ are mapped to $P_x(c)$ and $P_y(d)$---and are not already linked into multi-place relational structures (i.e., no T\textsubscript{3} units are active), then DORA links one of the mapped pairs (via a T\textsubscript{3}) unit, forming a functional multi-place relation. Functionally, learning can be defined as follows:

\begin{equation}
E_{i} = \left\{
        \begin{array}{ll}
            P_{a \cap b}^R (b^R) \leftarrow f_L(M(a^D, b^R)), & \nexists T_2 \\
            R_{i, n}^R \leftarrow f_L(M(P_1^D(a), P_1^R(b)) \dots M(P_n^D(c), P_n^R(d))), & \nexists T_3
        \end{array}
    \right.
    \label{eq:a3}
\end{equation}
% Equation A3

\noindent where $E_i$ is a learned representation, $P_i^J(a)$ is a single-place predicate $i$, in set $J (J \in [D=\text{driver}, R=\text{recipient}, M=\text{LTM}])$ that takes the argument $a$. Lowercase letters $a$, $b$, $c$, and $d$ indicate objects, $a \cap b$ indicates the intersection of the features of $a$ and $b$, $R_{i,n}^J$ is a relational structure $i$ (consisting of linked predicate argument pairs; see Eq. \ref{eq:a4}, directly below) of arity $n$ in set $J$, $T_k$ is an active token unit, and $f_L$ is the learning function. 

The relational structure $R^R_{i,n}$ is instantiated in DORA functionally as:

\begin{equation}
    R^R_{i,n} = [P_1^R(a) \Leftrightarrow \dots P_n^R(c)]
    \label{eq:a4}
\end{equation}
% Equation A4

\noindent where $\Leftrightarrow$ is a linking operator. For any $R_i$ a single T\textsubscript{3} unit instantiates all $n-1$ instances of the $\Leftrightarrow$ in $R_i$, linking (i.e., conjuncting) predicate-argument pairs $P_1^R(a) \dots P_n^R(c)$. 

\subsubsection{Refinement}

If DORA has learned mapping connections between representations in driver and recipient, then DORA can also learn a refined (or schematized) representation consisting of the featural intersection of the mapped representations. Refinement is defined as follows:

\begin{equation}
    R^{\prime M}_{i,n} \leftarrow f_R(M(R^D_n, R^R_n))
    \label{eq:a5}
\end{equation}
% Equation A5

\noindent where $R^{\prime J}_{i,n}$ (defined directly below) is a refined relational structure $i$ of arity $n$ in set $J$, and $f_R$ is the refinement function. The refined structure $R^{\prime M}_{i,n}$ is then:

\begin{equation}
    R^{\prime M}_{i,n} = [P^M_{P^D_1 \cap P^R_1}(a) \Leftrightarrow \dots P^M_{P^D_n \cap P^R_n}(b)]
    \label{eq:a6}
\end{equation}
% Equation A6

\noindent where $P_{P^D_i \cap P^R_j}$ is a single-place predicate composed of the featural intersection of mapped predicates $P_i^D$ and $P_j^R$. 

\subsubsection{Relational Generalization}

If DORA has learned mapping connections between representations in driver and recipient, then DORA can also perform relational generalization, inferring structure from the driver about items in the recipient. Generalization in DORA follows the standard copy-with-substitution-and-generalization format common in models of relational reasoning can be defined as follows:

\begin{equation}
    G^R_i \leftarrow f_G(M(R^D_j, R^R_j) \wedge \sim M(R^D_k))
    \label{eq:a7}
\end{equation}
% Equation A7

\noindent where $G_i$ is a generalized structure (see Eq. \ref{eq:a8}, directly below), $\sim M(A)$ is an unmapped structure $A$, and $f_G$ is the generalization function. The generalized structure $G_i^R$ is then: 

\begin{equation}
    G^R_i = [R^R_j \wedge R^R_k]
    \label{eq:a8}
\end{equation}
% Equation A8

\noindent where $R^R_j$ is the mapped relational structure from Eq. \ref{eq:a7}, and $R_k^R$ is generalized relational information in the recipient that matches the unmapped $R_k^D$ from Eq. \ref{eq:a7}. 

\subsection{Processing in DORA}

\begin{figure*}[!htbp]
\centering
\includegraphics{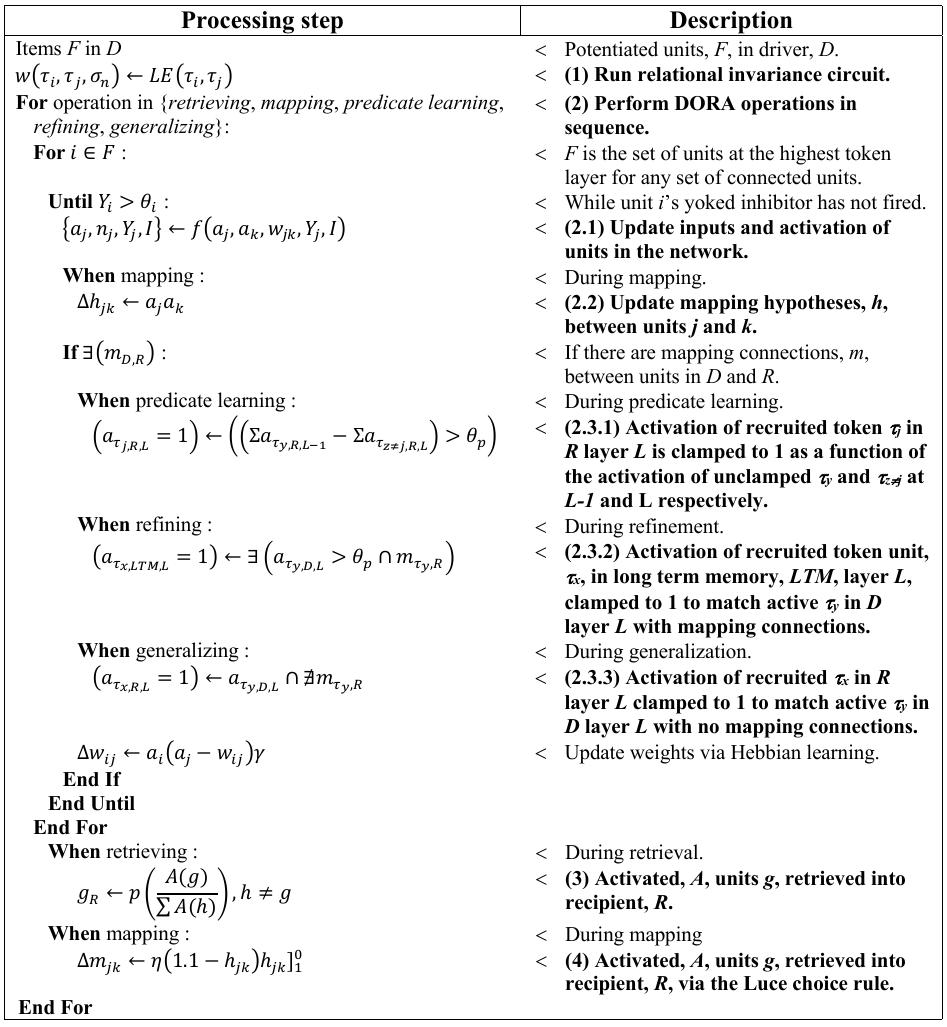}
\caption{Pseudocode of processing in DORA.}
\label{fig:A1}
\end{figure*}

DORA’s operation is outlined in pseudocode in Figure \ref{fig:A1}. The details of each step, along with the relevant equations and parameter values, are provided in the subsections that follow. DORA is very robust to the values of the parameters (see Doumas et al., 2008). For equations in this section, we use the variable $a$ to denote a unit’s activation, $n$ its (net) input, and $w_{ij}$ to denote the connection from unit $i$ to unit $j$. 

An analog, \textit{F} (selected at random, or based on the current game screen), enters the driver. Network activations are initialized to 0. Either (a) the firing order of propositions in \textit{F} is random (however, see  \parencite[however, see][]{hummel2003symbolic}, for a detailed description of how a system like DORA can set its own firing order according to the constraints of pragmatic centrality and text coherence), or (b) a roughly random firing order is instantiated by passing a top down input signal to all units $i$ in the highest layer of \textit{D} sampled from a uniform distribution with values between 0 and 0.4. DORA performs similarity and relative magnitude calculation through the relational invariance circuit, then runs retrieval from LTM, analogical mapping, and comparison-based unsupervised learning (predicate learning, refinement, and (relational) generalization). Currently, the order of operations of these routines is set to the order: retrieval, mapping, learning (predicate learning, refinement, and generalization). 

\subsubsection{Relational invariance circuit} % subsubsection 1. 

The operation of the relational invariance circuit is described in the main text. The relational invariance circuit runs when two or more items (T1 units) are present in the driver and those items are connected to similar magnitude representations (e.g., pixels, etc.). It is inhibited to inactivity when two or more circuit output features are active above threshold($=$.9) when both T1 units are active. As noted in the main text, we presently make no strong commitment to an account of dimensional salience \parencite[as discussed in, e.g., ][]{spelke2007core}. As such, two T1 units in the driver, both encoding an object or both encoding a predicate, are selected at random, and when those items are connected to multiple dimensions, a dimensional encoding becomes active at random (activation of features encoding that magnitude for the two T1 units are clamped to 1). The choice of two T1 units follows work on the WM capacity of children (e.g., Halford et al., 1998). As described in the main text, proxy units connected to the active driver T1 units update their activation and input by Eqs. \ref{eq:4} and \ref{eq:5} respectively. $E$ unit activation is updated by Eq. \ref{eq:7}. Feature units connected to $E$ units update their input by Eq. \ref{eq:8}, and their activation by Eq. \ref{eq:9}. Weights between feature units and active T1 units are updated by Eq. \ref{eq:10}. 

\subsubsection{Main DORA operations} % subsubsection 2.

Repeat the following until each token unit \textit{i} in the highest layer of \textit{F} has fired three times if mapping, or once, otherwise) (each token unit at the highest layer firing is referred to as the \textit{phase set}﻿). If a firing order has been set, select the current unit \textit{i} in the firing order and set \textit{a\textsubscript{i}} to 1.0. Otherwise, pass a top-down input (\textit{n} $=$ \textit{unif}(0,.4)) to token units in the highest layer of \textit{F}. 

\paragraph{A.2.1 Update inputs and activations of network units}

\subparagraph{A.2.1.1 Update mode of all T3 units in driver and recipient} 

T3 units in all propositions operate in one of three modes: Parent, child, and neutral \parencite{hummel1997distributed, hummel2003symbolic}. T3 mode is important for representing higher-order relations \parencite[e.g., \textit{R1}(\textit{x}, \textit{R2}(\textit{y}, \textit{z}));][]{hummel1997distributed}. As detailed below, higher-order relations are represented in DORA such that if one proposition takes another as an argument, the T3 unit of the lower-order proposition serves as the object of an T2 unit for the higher-order proposition (i.e., the lower-order T3 unit is downwardly connected to the T2 unit, as a T1 unit would be), and the T3 unit represented the lower-order proposition operates in child mode. By contrast, when a T3 unit is not acting as the argument of another proposition, it operates in parent mode. The mode \textit{m\textsubscript{i}} of T3 unit \textit{i} is updated by the equation: 

% format the equation environment
\renewcommand{\theequation}{A.\arabic{equation}}

\begin{equation}
    m_{i} =\begin{Bmatrix}
    Parent(1), & T2_{above}<T2_{below} \\ 
    Child(-1), & T2_{above}>T2_{below} \\ 
    Neutral(0), & otherwise \\ 
    \end{Bmatrix}\\ 
    \label{eq:a9}
\end{equation}
% Equation (A9)

\noindent where, $T2_{above}$ is the summed input from all T2 units to which $i$ is upwardly connected (i.e., relative to which $i$ serves as an argument), and $T2_{below}$ is the summed input from all T2 units to which $i$ is downwardly connected. In the current simulations, T3 mode did not have to change their mode. We include this step here solely for the purposes of completeness \parencite{hummel1997distributed, hummel2003symbolic}. 

\subparagraph{A.2.1.2 Update input to all units in the network}

\textit{A.2.1.2.1 Update input to all token units in driver:} Token units in the driver update their input by the equation: 

\begin{equation}
n_{i} =  \sum_{j}^{}a_{j}w_{ij}G-\sum_{k}^{}a_{k}-s\sum_{m}^{}3a_{m}-10I_{i}\\ 
\label{eq:a10}
\end{equation}
% Equation (A10)

\noindent where $j$ are all units above unit $i$ (i.e., T3 units for T2 units, T2 units for T1 units), $G$ is a gain parameter attached to the weight between the T2 and its T1 units (T1 units learned via DORA’s comparison-based predication algorithm (see below) have $G=2$ and all other T1 units have $G=1$), $k$ is all units in the driver in the same layer as $i$ (for T1 units, $k$ is only those T1 units and T3 units currently in child mode not connected to the same T2 as unit $i$; see step 2.1.1), $m$ are T1 units that are connected to the same T2 (or T2 units) as $i$, and $I_i$ is the activation of the T1 inhibitor yoked to $i$. When DORA is operating in binding-by-asynchrony mode, $s=1$; when it is operating in binding-by-synchrony mode (i.e., like LISA), $s=0$. 

\textit{A.2.1.2.2. Update input to feature units:} Feature units update their input as: 

\begin{equation}
n_{i} = \sum_{j\in S\in\left(D,R\right)}^{}a_{j}w_{ij} \\ 
\end{equation}
\label{eq:a11}
% Equation (A11)

\noindent where $j$ is all T1units in $S$, which is the set of propositions in driver, $D$, and recipient $R$, and $w_{ij}$ is the weight between T1unit $j$ and feature unit $i$. In LISA \parencite[see][]{hummel1997distributed}, when multiple propositions are in the driver simultaneously it ignores the features of the arguments. This convention follows from the assumption that people default to thinking of single propositions at a time, and the only reason to consider multiple propositions simultaneously is to consider structural constraints \parencite[e.g., ][]{hummel2003symbolic, medin1993respects}, and has been adopted in DORA. When there are multiple positions in the driver, input to semantics is taken only from T1 units $j$ acting as roles (i.e., with $mode=1$; see below). 

\textit{A.2.1.2.3. Update input to token units in recipient and LTM:} Input to all token units in recipient and LTM are not updated for the first 5 iterations after the global or local inhibitor fires. Token units in recipient and token units in LTM during retrieval update their input by the equation: 

\begin{equation}
n_{i} =  \sum_{j}^{}a_{j}w_{ij}+SEM_{i}+M_{i}-\sum_{k}^{}a_{k}-s\sum_{m}^{}3a_{m}-\sum_{n}^{}a_{n}-\Gamma_{G}-\Gamma_{L} \\ 
\end{equation}
\label{eq:a12}
% Equation (A12)

\noindent where $j$ is are any units above token unit $i$ (i.e., T3 units for T2 units, T2 units for T1 units; input from $j$ is only included on phase sets beyond the first), $SEM_i$ is the feature input to unit $i$ if unit $i$ is a PO, and 0 otherwise, $M_i$ is the mapping input to unit $i$, $k$ is all units in either recipient (if unit \textit{i} is in recipient) or LTM (if unit \textit{i} is in LTM) in the same layer as \textit{i} (for T1 units, $k$ is only those T1 units and T3 units currently in child mode not connected to the same T2 as unit $i$; see section 2.1.1), $m$ is T1 units connected to the same T2 as $i$ (or 0 for non-T1 units), $n$ is units above unit $i$ to which unit $i$ is not connected, $\Gamma_{G}$ is the activation of the global inhibitor (see below), and $\Gamma_{L}$ is the activation of the local inhibitor in the same layer as or any layer above $i$. When DORA is operating in binding-by-asynchrony mode, $s=1$; when it is operating in binding-by-synchrony mode (i.e., like LISA), $s=0$. $SEM_i$, the feature input to $i$, is calculated as: 

\begin{equation}
SEM_{i} =\frac{\sum_{j}^{}a_{j}w_{ij}}{1+num\left(j\right)} \\ 
\end{equation}
\label{eq:a13}
% Equation (A13)

\noindent where $j$ are feature units, $w_{ij}$ is the weight between feature unit $j$ and T1unit $i$, and $num(j)$ is the total number of feature units $i$ is connected to with a weight above $\theta(=0.1)$. $M_i$ is the mapping input to $i$: 

\begin{equation}
M_{i} = \sum_{j}^{}a_{j}\left(3w_{ij}-Max\left(Map\left(i\right)\right)-Max\left(Map\left(j\right)\right)\right)\\ 
\label{eq:a14}
\end{equation}
% (A14)

\noindent where $j$ are token units of the same type as $i$ in driver (e.g., if $i$ is a T2 unit, $j$ is all T2 units in driver), $Max(Map(i))$ is the highest of all unit $i$’s mapping connections, and $Max(Map(j))$ is the highest of all unit $j$’s mapping connections. As a result of Eq. \ref{eq:a14}, active token units in driver will excite any recipient units of the same type to which they map and inhibit all recipient units of the same type to which they do not map. 

\subparagraph{A.2.1.3 Update input to the yoked inhibitors}

Every token unit is yoked to an inhibitor unit $i$. T2 and T3 inhibitors are yoked only to their corresponding T2. T1 inhibitors are yoked both to their corresponding T1 and all T2 units in the same analog. Inhibitors integrate input over time as: 

\begin{equation}
n_{i}^{\left(t+1\right)} = n_{i}^{\left(t\right)}+a_{j}+\sum_{k}^{}a_{k} \\
\label{eq:a15}
\end{equation}
% Equation (A15)

\noindent where $t$ refers to the current iteration, $j$ is the token unit yoked to inhibitor unit $i$, and $k$ is any T2 units if $j$ is a T1, and 0 otherwise. Inhibitor units become active ($a_{i}=1$) when $n_i$ is greater than the activation threshold ($=220$ for T2 and T1 units; $220 \ast n$ for T3 units—where $n$ is the number of T2 units the T3 units is connected to). All T1 and T2 inhibitors become refreshed $a_{i} =0$ and $n_{i} =0$) when the global inhibitor ($\Gamma_G$; described below) fires. 

\subparagraph{A.2.1.4 Update the local and global inhibitors}

The local inhibitors, $\Gamma_L$, are inhibitory units connected to all units in a single layer of LTM (i.e., there is a local inhibitor for T1 units, another for T2 units). The local inhibitor is potentiated ($P(\Gamma_{L})=1$) when a driver unit in $\Gamma_L$’s layer is active, is inhibited to inactivity ($\Gamma_L=0$) by any driver unit in its layer with activation above $\Theta_{L}=0.5$), and becomes active ($\Gamma_{L}=10$) when no token unit in its layer has an activity above $\Theta_{L}$. A firing local inhibitor sets the activation and potentiation of all other local inhibitors below and including itself to 0. The global inhibitor, $\Gamma_{G}$, is potentiated ($P(\Gamma_{G})=1$) when any driver units are active, and is inhibited to inactivity ($\Gamma_{G}=0$) by any driver unit in its layer with activation above $\Theta_{G}$ (=0.5), and becomes active ($\Gamma_{G}=10$) when no T1 in its layer has an activity above $\Theta_{G}$. The global inhibitor sets activation and potentiation of all other local inhibitors to 0. 

\subparagraph{A.2.1.5 Update activations of all units in the network}

All token units in DORA update their activation by the leaky integrator function: 

\begin{equation}
\Delta a_{i} = \gamma n_{i}\left(1.1-a_{i}\right)-\delta a_{i}]_{1}^{0} \\ 
\label{eq:a16}
\end{equation}
% Equation (A16)

\noindent where $\Delta a_{i}$ is the change in activation of unit $i$, $\gamma$ (=0.3) is a growth parameter, $n_{i}$ is the net input to unit $i$, and $\delta$ (=0.1) is a decay parameter. Activation of all token units $i$, is hard limited to between 0 and 1 inclusive. 

Feature units update their activation by the equation: 

\begin{equation}
a_{i} =\frac{n_{i}}{\max\left(n_{j}\right)} \\ 
\label{eq:a17}
\end{equation}
% Equation (A17)

\noindent where $a_{i}$ is the activation of feature unit $i$, $n_{i}$ is the net input to feature unit $i$, and $max(n_{j})$ is the maximum input to any feature unit. There is physiological evidence for divisive normalization in the feline visual system (e.g., Bonds, 1989; Heeger, 1992) and psychophysical evidence for divisive normalization in human vision (e.g. Foley, 1994; Thomas \& Olzak, 1997). 

Token unit inhibitors, $i$, update their activations according to a threshold function: 

\begin{equation}
a_{i} =\begin{Bmatrix}
1, & n_{i}>\Theta_{IN} \\ 
0, & otherwise \\ 
\end{Bmatrix}\\ 
\label{eq:a18}
\end{equation}
% Equation (A18)

\noindent where $\Theta_{IN}=220$ for T1 and T2 units and $220 * n$ for T3 units (where $n$ is the number of T2 units to which that T3 unit is connected). 

\paragraph{A.2.2 Update mapping hypotheses}

If mapping is licensed, DORA learns mapping hypotheses between all token units in driver and token units of the same type in recipient (i.e., between T3 units, between T2 units and between T1 units in the same mode [described below]). Mapping hypotheses initialize to zero at the beginning of a phase set. The mapping hypothesis between a driver unit and a recipient unit of the same type is updated by the equation: 

\begin{equation}
\Delta h_{ij}^{t} = a_{i}^{t}a_{j}^{t}\\ 
\label{eq:a19}
\end{equation}
% Equation (A19)

\noindent where $a_{i}^{t}$ is the activation of driver unit $i$ at time $t$. 

\paragraph{A.2.3 Comparison-based unsupervised learning}

If licensed, DORA will perform comparison-based-learning (CBL). CBL is unsupervised. In the current version of the model, learning is licensed whenever 70$\%$ of driver token units map to recipient items (this 70$\%$ criterion is arbitrary, and in practice either 0$\%$ or 100$\%$ of the units nearly always map). 

\subparagraph{A.2.3.1 Predicate and relation learning}

During predicate and relation learning, DORA recruits (and clamps the activation of) token units in the recipient to respond to patterns in firing in adjacent layers. The recruitment procedure is a simplified version of ART (Carpenter $\&$ Grossberg, 1990). Each layer of token units $i$ is connected to a clamping unit $C_{i}$, which 10 iterations after any inhibitor unit has fired, is activated by the equation: 

\begin{equation}
C_{i} =\begin{Bmatrix}
1, & (\sum_{j}max(a_j) - \sum_{i}max(a_i)) \geq \theta_c \\
0, & otherwise \\ 
\end{Bmatrix} \\ 
\label{eq:a20}
\end{equation}
% Equation (A20)

\noindent where $a_{j}$ is the activation of unclamped token units in the layer below $i$ for T2 and T3 units, $a_{i}$ is the activation of unclamped token units in layer $i$, $max(a_{j})$ is the maximum activation of a unit in layer $j$, and $\theta_{c}$ is a threshold (=0.6). $C_{i}$ for the T1 layer is equal to $C_{i}$ for the T2 layer. 

An active $C_{i}$ in the recipient sends an input, ($p_{j}=1.0$), to a randomly selected token unit, $j$ (where $j$ is not connected to units in other layers), in layer $i$ ($p+{k} = 0$ for all units $k \neq j$). Token units are clamped by the equation: 

\begin{equation}
c_{j} =\begin{Bmatrix}
1, & p_{i}-3\sum_{k}^{}a_{k}>0 \\ 
0, & otherwise \\ 
\end{Bmatrix}\\ 
\label{eq:a21}
\end{equation}
% Equation (A21) 

\noindent where $c_{j}$ is the clamped activation of unit $j$ in layer $i$, and $a_k$ is the activation of all clamped token units in the same layer as $j$, where $k \neq j$, if $j$ is in T1, and all token units in the same layer as $j$, where $k \neq j$, otherwise. Unit $j$ remains clamped until $\Gamma_L$ fires and $j$ is inhibited to inactivity. If the recruited token is in T1 its mode is set to 1 (marking it as a learned representation; although the idea of units firing in modes sounds nonneural, \textcite{hummel1997distributed} described how it can be accomplished with two or more auxiliary nodes with multiplicative synapses) and connections between the recruited token unit and all active features update by the equation: 

\begin{equation}
\Delta w_{ij} = a_{i}\left(a_{j}-w_{ij}\right)\gamma \\ 
\label{eq:a22}
\end{equation}
% Equation (A22)

\noindent where $\Delta w_{ij}$ is the change in weight between the new T1 unit $i$, and feature unit $j$, $a_{i}$ and $a_{j}$ are the activations of $i$ and $j$, respectively, and $\gamma$ is a growth rate parameter. Additionally, connections between corresponding token units (i.e., between T3 and T2, or T2 and T1 units) are also updated by Eq. \ref{eq:a22}, where $i$ are recipient token units in layers adjacent to recruited unit $j$. When the phase set ends, connection weights between a T2 or T3 unit $i$ and any token unit in the adjacent lower layer $j$ (i.e., $j$ is a T2 unit when $i$ is a T3 unit, and $j$ is a T1 unit when $i$ is a T2 unit), are updated by the equation: 

\begin{equation}
w_{ij} =\begin{Bmatrix}
w_{ij}, & \sum_{k}^{}w_{ik}\geq 2 \\ 
0, & otherwise \\ 
\end{Bmatrix}\\ 
\label{eq:a23}
\end{equation}
% Equation (A23)

\noindent where $k$ is all other units, including $j$, in the same layer as $j$. This operation removes weights to redundant tokens that do not conjunct two or more units at a lower layer. 

\subparagraph{A.2.3.2 Refinement Learning}

During refinement, DORA infers token units in the LTM that match active tokens in the driver. Specifically, DORA infers a token unit in the LTM in response to any mapped token unit in the driver. If unit $j$ in the driver maps to nothing in the LTM, then when $j$ fires, it will send a global inhibitory signal to all units in the LTM (Eq. \ref{eq:a14}). This uniform inhibition, unaccompanied by any excitation in recipient, is a signal that DORA exploits to infer a unit of the same type (i.e., T1, T2, T3) in LTM. Inferred T1 units in the LTM have the same mode as the active T1 in driver. The activation of each inferred unit in the LTM is set to 1. DORA learns connections between corresponding active tokens in the LTM (i.e., between T3 and T2 units. and between T2 and T1units) by Eq. \ref{eq:a22} (where unit $j$ is the newly inferred token unit, and unit $i$ is any other active token unit). To keep DORA’s representations manageable (and decrease the runtime of the simulations), at the end of the phase set, we discard any connections between feature units and T1 units whose weights are less than 0.1. When the phase set ends, connection weights between any T2 or T3 unit $i$ and token units at a lower adjacent layer $j$ to which $i$ has connections are updated by Eq. \ref{eq:a23}. 

\subparagraph{A.2.3.3 Relational generalization}

The relational generalization algorithm is adopted from \textcite{hummel2003symbolic}. As detailed in Eq. \ref{eq:a14}, when a token unit $j$ in driver is active, it will produce a global inhibitory signal to all recipient units to which it does not map. A uniform inhibition in recipient signals DORA to activate a unit of the same type (i.e., T1, T2, T3) in recipient as the active token unit in driver. DORA learns connections between corresponding active tokens in the LTM (i.e., between T3 and T2 units. and between T2 and T1 units) by the simple Hebbian learning rule in Eq. \ref{eq:a22} (where unit $j$ is the newly active token unit, and unit $i$ is the other active token unit). Connections between T1 units and feature units are updated by Eq. \ref{eq:a22}. When the phase set ends, connection weights between any T2 or T3 unit $i$ and any token units in an adjacent lower layer $j$ to which $i$ has connections are updated by Eq. \ref{eq:a23}. 

\subsubsection{Retrieval}

DORA uses a variant of the retrieval routine described in \textcite{hummel1997distributed}. During retrieval units in the driver fire as described above for one phase set. Units in the LTM become active as in step 2.1. After one phase set representations are retrieved from LTM into the recipient probabilistically using the Luce choice axiom: 

\begin{equation}
L_{i} =\frac{R_{i}}{\sum_{j}^{}R_{j}}\\ 
\label{eq:a24}
\end{equation}
% Equation (A24)

\noindent where $L_{i}$ is the probability that T3 unit $i$ will be retrieved into working memory, $R_{i}$ is the maximum activation T3 unit $i$ reached during the retrieval phase set and $j$ are all other T3 units in LTM.  If a T3 unit is retrieved from LTM, the entire structure of tokens (i.e., connected T1 $\ldots$ T3 units) are retrieved into recipient.  

\subsubsection{Update mapping connections}

If DORA is mapping, mapping connections are updated at the end of each phase set. First, all mapping hypotheses are normalized by the equation: 

\begin{equation}
h_{ij} =\left(\frac{h_{ij}}{MAX\left(h_{i},h_{j}\right)}\right)-MAX\left(h_{kl}\right)\\ 
\label{eq:a25}
\end{equation}
% (A25)

\noindent where, $h_{ij}$ is the mapping hypothesis between units $i$ and $j$, $MAX(h_{i}, h_{j})$ is the largest hypothesis involving either unit $i$ or unit $j$, and $MAX(h_{kl})$ is the largest mapping hypothesis where either $k=i$ and $l \neq j$, or $l=j$ and $k \neq i$. That is, each mapping hypothesis is normalized divisively: Each mapping hypothesis, $h_{ij}$ between units $i$ and $j$, is divided by the largest hypothesis involving either unit $i$ or $j$. Next each mapping hypothesis is normalized subtractively: The value of the largest hypothesis involving either $i$ or $j$ (not including $h_{ij}$ itself) is subtracted from $h_{ij}$. The divisive normalization keeps the mapping hypotheses bounded between zero and one, and the subtractive normalization implements the one-to-one mapping constraint by forcing mapping hypotheses involving the same $i$ or $j$ to compete with one another. Finally, the mapping weights between each unit in driver and the token units in recipient of the same type are updated by the equation: 

\begin{equation}
\Delta w_{ij} = \eta\left(1.1-w_{ij}\right)h_{ij}]_{1}^{0}\\
\label{eq:a26}
\end{equation}
% (A26)

\noindent where $\Delta w_{ij}$ is the change in the mapping connection weight between driver unit $i$ and recipient unit $j$, $h_{ij}$ is the mapping hypothesis between unit $i$ and unit $j$, $\eta$ is a growth parameter, and $\Delta w_{ij}$ is truncated for values below 0 and above 1. After each phase set, mapping hypotheses are reset to 0. The mapping process continues for three phase sets.

\subsubsection{Learning the relational invariance circuit}

As described in the main text, the relational invariance circuit consists of three layers of nodes. At the top layer, proxy units are connected to individual T1 units in the driver. The next layer, $E$, consists of four nodes and takes input from any active proxy units. At the bottom are feature units, initially randomly connected to nodes in $E$. Weights between units in $E$ and feature units are initialized to random numbers between 0 and 0.9, and lateral connections for $E$ are set to -1. Connections between units in $E$ and feature units are updated by Eq. \ref{eq:a10} in the main text.

\subsubsection{Higher-order relations}

Although they are not necessary for the current simulations, for the purposes of completeness it is important to note that DORA easily represents higher-order relations \parencite[i.e., relations between relations; see][]{hummel1997distributed, doumas2008theory}. In short, when a proposition takes another proposition as an argument, the T3 unit of the lower-order proposition serves as the object of an T2 unit for the higher-order proposition.  For example, in the higher-order relation \textit{greater} (\textit{distance} (a, b), \textit{distance} (c, d)), the T3 unit of the proposition \textit{distance} (a, b) serves as the argument of the \textit{more} role of the higher-order \textit{greater} relation, and the T3 unit of the proposition \textit{distance} (c, d) serves as the argument of the \textit{less} role of the higher-order \textit{greater} relation. When a T3 unit serves as the object of an T2 unit, it operates in child mode (see above). The modes of T3 units change as a function of whether they are receiving top-down input. A T3 unit receiving top-down input from an T2 unit (i.e., when the T3 unit is serving as the argument of that T2 unit) will operate in child mode, while a T3 unit not receiving input from any T2 units will operate in parent mode.

\section{Simulation Details}

\subsection{DORA: Learning representations from screens}

We used DORA to simulate learning structured representations from screen shots from the game Breakout. This simulation aims to mirror what happens when a child (or adult) learns from experience in an unsupervised manner (without a teacher or guide). While we describe the results in terms of DORA learning to play Breakout and generalizing to Pong, but results were the same when run in the other direction (i.e., train on Pong and generalize to Breakout; Supplemental results, Figure S1). 

For Simulation 2, screens were generated from Breakout during 250 games with random move selection. Each screen from each game was processed with the visual pre-processor that identified objects and returned the raw pixel values as features of those objects. When learning in the world, objects have several extraneous properties. To mirror this point, after visual pre-processing, each object was also attached to a set of 100 additional features selected randomly from a set of 10000 features. These additional features were included to act as noise, and to make learning more realistic. (Without these noise features, DORA learned exactly as described here, only more quickly.)  
DORA learned from object representations in an unsupervised manner. On each learning trial, DORA selected one pair of objects from a screen at random. DORA attempted to characterize any relations that existed between the objects using any relations it has previously learned (initially, it had learned nothing, and so nothing was returned) by selecting a dimension at random and running the two objects through the relational invariance circuit (described above) over that dimension. If the features returned matched anything in LTM (e.g., "more" and "less" "x"), then DORA used that representation from LTM to characterize the current objects. DORA then ran (or attempted to run) retrieval from LTM, the relational invariance circuit, mapping, and representation learning (see above). Learned representations were stored in LTM. We placed the constraint on DORA’s retrieval algorithm such that more recently learned items were favoured for retrieval. Specifically, with probability .6, DORA attempted to retrieve from the last 100 representations that it had learned. This constraint followed our assumption that items learned more recently are more salient and more likely to be available for retrieval. 

The process was identical for Simulation 3, except that instead of screens from Breakout, we used the first 300 images from the CLVR dataset for representation learning. In simulation 4, we had two ablated versions of the model: In the first ablated model (A1), we ablated the inhibitory connections from the onset of the simulation; in the second ablated model (A2), we ablated the inhibitory connections after the model had learned to play Breakout. Representation learning for both models was as in simulation 1. 

\subsection{DORA: Q-learning for game play}

For Simulations 2, 3, and 4, for a given screen, DORA used the representations it had previously learned to represent the relations between objects on that screen and the previous screen. That is, for any pair of objects, if DORA had learned a representation that characterized the relation between the two objects (in LTM and as measured by the relational invariance circuit), DORA used that representation the characterize the objects. 

The relations were then used to form a table of encountered relational states, and Q-learning \parencite{watkins1989learning} was used to learn the approximate action-value function for Breakout. We used a rule length constraint of two relations per state, reflecting the simplicity of the game and the WM capacity exhibited by humans  \parencite{logie2020working}.We trained DORA decreasing the learning rate linearly from 0.1 to 0.05 and the exploration rate linearly from 0.1 to 0.01 throughout the training session. We saved the version of the table that yielded the maximum score during the session.  

\subsection{Deep Q-Network}

A Deep Q-Network \parencite[DQN;][]{mnih2015human} was trained to play Breakout and Pong. The raw 210 $\times$ 160 frames were pre-processed by first converting their RGB representation to grey-scale and down-sampling it to a 105 $\times$ 80 image. We stacked the last 4 consecutive frames to form the input each state. 

The input to the neural network was the 105 $\times$ 80 $\times$ 4 pre-processed state. The first hidden convolutional layer applied 16 filters of size 8 x 8 with stride 4 with a relu activation function. The second hidden convolutional layer applied 32 filters of size 4 x 4 with stride 2 with a relu activation function. The third hidden layer was fully connected of size 256 with a relu activation function. The output layer was fully connected with size 6 and a linear activation function. 

We implemented all the procedures of the DQN to improve training stability, in particular: (a) We used memory replay of size 1,000,000. (b) We used a target network which was updated every 10,000 learning iterations. (c) We fixed all positive rewards to be 1 and all negative rewards to be $-$1, leaving 0 rewards unchanged. (d) We clipped the error term for the update through the Huber loss. 

We also ran the same network using the input from the visual preprocessor described above. 

\subsection{Supervised deep neural network}

We trained a deep neural network (DNN) in a supervised manner to play Breakout and Pong and tested generalization between games. One network was trained using random frame skipping and the other with fixed frame skipping. 

The inputs to the network were the output of the visual preprocessor described above. Specifically, the network took as input the x and y positions of the ball and player-controlled paddle, as well as the left paddle for Pong (left as zeros when playing Breakout). The input to the neural network was a vector of size 24 corresponding to the pre-processed last seen 4 frames. This was fed to three fully connected layers of size 100 each with a relu activation function. The output layer was fully connected with size 6 and a softmax activation function. 

The criteria for training was the correct action to take in order to keep the agent-controlled paddle aligned with the ball. In Breakout if the ball was to the left of the paddle the correct action las ‘LEFT’, if the ball was to the right of the paddle the correct action was ‘RIGHT’ and if the ball and the paddle were at the same level on the x-axis the correct action was ‘NOOP’. In Pong if the ball was higher than paddle the correct action was ‘RIGHT’, if the ball was lower than paddle the correct action was ‘LEFT’ and if the ball and the paddle were at the same level on the y-axis the correct action was ‘NOOP’. This action was encoded as a one-hot vector (i.e., activation of 1 for the correct action and cero for all other actions). 

\subsection{Graph network}

Graph networks \parencite[see][, for a review]{Battaglia2018RelationalIB} are neural network models designed to approximate functions on graphs. A graph is a set of nodes, edges, and a global feature. The representation of the nodes, edges, and the global attribute encode feature information. A graph network takes as input a graph and returns a graph of the same size and shape, but with updated attributes. 

Our graph net agent used a encode-process-decode architecture  \parencite{Battaglia2018RelationalIB} where three different graph networks are arranged in series. The first graph net encodes the nodes, edges and global attributes independently, the second graph net performs three recurrent steps of "message passing" and the third graph net decodes the nodes, edges and global attributes independently. 

The graph agent takes in a graph-structured representation of the screen where each object corresponds to a node in the graph. In our simulations, the node representation corresponds to the position, area, color and velocity of the objects in the screen. In order to use the graph network as a reinforcement learning agent we set the number of edge attributes to the number of possible actions. In this way, our agent produces a vector of Q-values for each edge, corresponding to the valid actions in each game. To choose actions, the agent takes an argmax across all edges’ Q-values. 

To train our agent we used a replay memory of size 50000. Before training we feed the replay memory with 1600 memories (i.e., tuples containing a state graph, action, edge, reward, next state graph, and a "done" variable). At each time step, we saved the current memory and sample a batch of 32 memories from the replay memory to train the agent. We used the Adam learning algorithm with a learning rate of 0.01 and default learning parameters. 

\section{Supplemental results}

\subsection{Supplemental simulation 1: Pong to breakout generalization results}

\begin{figure*}[!htbp]
\centering
\includegraphics[width=13.5cm]{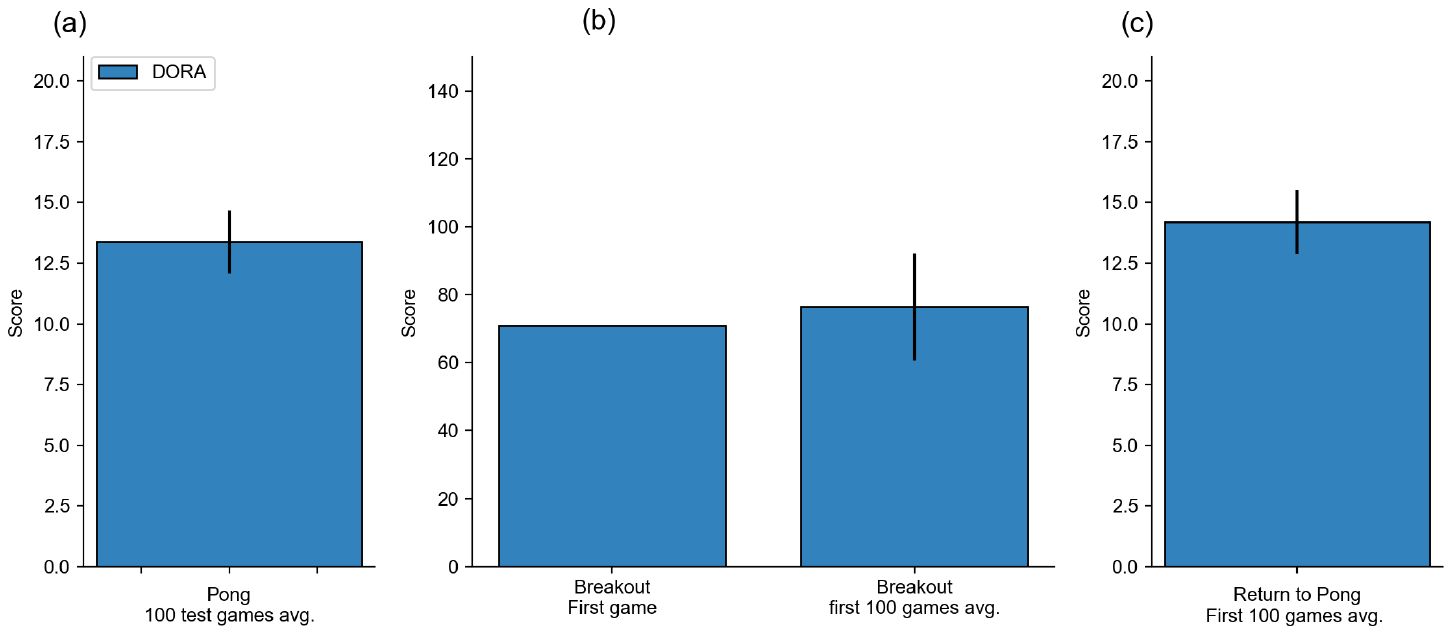}
\caption{Results of simulations with DORA trained on Pong and generalizing to Breakout, with DORA learning representations from Pong. Error bars represent 2 stderrors. (A) Performance of DORA on Breakout as an average of 100 test games. (B) Results of DORA playing Breakout after trainig on Pong as the score of the first game played and an average score of the first 100 games played. (C) Results of DORA when returning to play Pong, as an average score for the first 100 games played. }
\label{fig:S1}
\end{figure*}

As described in the main text, DORA learned representations from Pong games. The model learned to play Pong first, and then generalized to Breakout. Results in Fig. \ref{fig:S1}. 

\subsection{Supplemental data 1}

\begin{figure*}[!htbp]
\centering
\includegraphics[width=13.5cm]{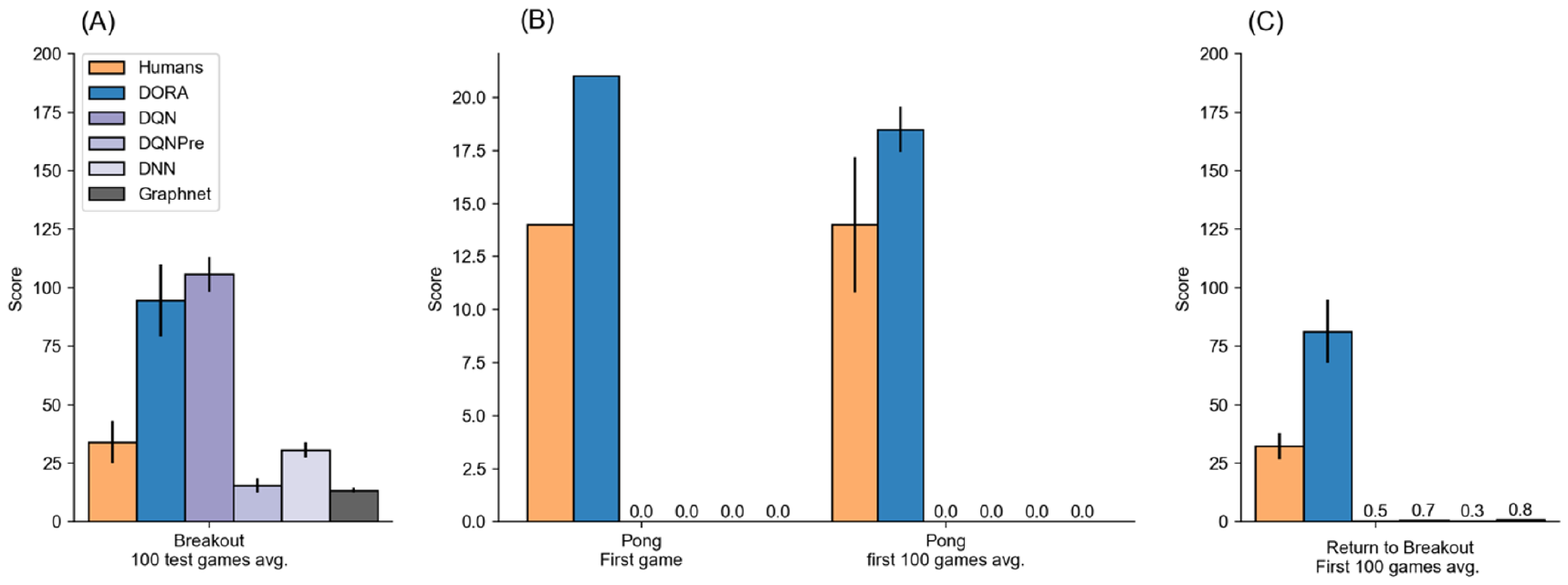}
\caption{Figure S2. Results of game play simulations with Humans, DORA, the DQNs and the DNNs. Error bars represent 2 stderrors. (A) Performance humans and networks on Breakout as an average of 100 test games. (B) Results of humans and networks playing Pong after training on Breakout as score on the first game played and mean score over the first 100 games played. (C) Results of humans and networks when returning to play Breakout after playing or learning to play Pong as an average of the first 100 games played.} 
\label{fig:S2}
\end{figure*}

Two human novices were trained on Breakout for 300 games, then transferred to playing Pong for 100 games, followed by moving back to Breakout for 100 games (these games were played in 2 hours session spread across 6 days; the last 50 games of Breakout and first 20 games of Pong were completed in the same session). Human players, of course, come into playing these games with a life of experience with the world, spatial relations, and other video games, and bring this experience to bear on playing both games. As humans regularly engage in cross-domain generalization, we expect the participants to generalize between games. A comparison of these highly trained humans and DORA and the various DNNs tested in the main text appears in Figure \ref{fig:S2}. 

In addition to the two participants who played several hundred games of Breakout and Pong, we ran 8 additional participants in a simple transfer task. Participants either played Breakout for 50 minutes followed by playing Pong for 10 minutes (4 participants) or played Pong for 50 minutes followed by playing Breakout for 10 minutes (4 participants). We had players play to a time limit rather than a number of games, as a game of Pong takes roughly 4 times as long as a game of Breakout. The average score on the first game of Pong when played first was 6.25 vs. 14.0 when played after Breakout. The average score on the first game of Breakout when played first was 9.5 vs. when played 19.0 when played after Pong. We analyzed the effects of order (whether a game was played first or after another game) on performance use a simple linear mixed effects model with Score predicted by order with participants as a random variable. Because the scores in Breakout and Pong are on different scales (Pong goes to 21, Breakout is (theoretically) unbounded) we normalized all scores by subtracting each score from the grand mean of scores on that game (e.g., each Pong score had the mean of score of all Pong games subtracted from it). We then compared the model using order to predict score (with participants as a random variable) to the null model. The full model explained significantly more variance than the null model (chi-squre(1) = 4.07, p < 0.05), with scores in the transfer condition significantly higher than scores in the initial game condition. 

Participants were run using the online javatari system (https://javatari.org/). 

One noteworthy limitation of DORA’s gameplay is that it was slower to learn Breakout than the human players we tested. We suspect the reason for this limitation is that in most simulations, DORA, like the DNNs we ran for comparison, began as a \emph{tabula rasa} with no understanding of anything at the beginning of learning. As a result, DORA spent much of its early experience in these simulations simply acquiring basic relational concepts such as \textit{left-of} (). By contrast, most people start playing video games long after they have acquired such basic concepts. In other words, our ability to play a game such as Breakout, even for the first time, is already facilitated by an enormous amount of cross-domain transfer: People know what to look for (e.g., “where is the paddle relative to the ball?”) and how to represent the answer (“to the left of the ball”) even before starting to learn how to play the game. DORA, by contrast, had to learn these basic concepts while learning to play the game. 

We argue that the reason DORA learned so much faster than the DNNs is that DORA was biased from the beginning to look for the right thing. Whereas DNNs search for representations that minimize the error in the input-output mapping of the task at hand, DORA looks for systematic relations that allow it to build a model of the task it is learning. Once it has learned this model, DORA is off and running, prepared to transfer its learning to new tasks, such as Pong. By contrast, the DNN is trying to be the best it can at exactly this one task; it is trying to memorize exactly what to do in response to every possible situation. In the end, the DNN will be a better Breakout player than DORA. But DORA, unlike the DNN, will be able to transfer its learning to other tasks, including but not limited to Pong. 

We argue that people are more like DORA than a DNN. You and I will never beat a well-trained DNN at chess, or Go, or probably any other task on which a DNN has been adequately trained. But at the end of the day, we will be able to drive home, make dinner, put our children to bed, and have a glass of wine. All the DNN will know how to do is beat the next competitor. And more importantly, the DNN will likely be unable to learn how to perform these other tasks without forgetting how to play chess. A human is a general-purpose learning machine that exceeds at using what it already knows to bootstrap its learning of things it doesn’t already know. A deep net is more a one-trick pony (though possibly the best in the world at its one over-trained task). 

\subsection{Supplemental simulation 2: Inverse Breakout} 

We ran a simple simulation of this capacity using a modified version of Breakout. In this version, the rules were adjusted such that missing the ball was rewarded and hitting the ball was punished (i.e., points were scored when the ball went past the paddle, and a life was lost when the ball struck the paddle; essentially the reverse of the regular Breakout rules). We ran tested a version of the DORA model that had previously learned to play Breakout successfully (see simulation 2, main text). Unsurprisingly, initially the model followed the previously successful strategy of following the ball to contact it and send it towards the point-scoring bricks. However, upon contact with the paddle, the model was punished with a lost life. As noted in the main text, the model had previously learned that following the ball predicted reward (points), and that moving away from the ball predicted punishment (lost life). After three lost lives, DORA attempted to compare the representation of the current game to the representation it had previously learned from Breakout. 

The current representation was that moving the paddle toward the ball resulted in punishment, or \textit{left-of} (ball, paddle1) then \textit{left-of} (paddle2, paddle1) $\rightarrow$ punishment signal. The previous representation of the game was that moving toward the ball resulted in reward and away from the ball resulted in punishment, or \textit{left-of} (ball, paddle1) then \textit{left-of} (paddle2, paddle1) $\rightarrow$ reward signal, and left-of (ball, paddle1) then right-of (paddle2, paddle1) $\rightarrow$ punishment signal. As described in the main text, DORA performed mapping and relational inference with these two representations. With P1:  \textit{left-of}(paddle2, paddle1) $\rightarrow$ punishment signal in the driver, and P2: \textit{left-of} (paddle2, paddle1) $\rightarrow$ reward signal and P3: \textit{right-of} (paddle2, paddle1) $\rightarrow$ punishment signal in the recipient, DORA mapped left-of (paddle2, paddle1) in P1 to \textit{left-of} (paddle2, paddle1) in P2, and punishment signal in P1 to punishment signal in P3. DORA then flipped the driver and recipient (P2 and P3 now in the driver and P1 in the recipient) and performed relational inference. During relational inference, it copied the unmapped reward-signal from P2 and right-of (paddle2, paddle1) from P3 into the recipient, thus inferring that right-of (paddle2, paddle1) predicts a reward signal. When adopting this strategy, moving away from the ball, DORA started scoring points (because the new task was so easy, we had to decide a point total to stop the game). 

\end{appendix}

\end{document}